\documentclass[]{novel2film}
\definecolor{w_blue}{RGB}{52,204,204}
\definecolor{w_yellow}{RGB}{255,192,0}

\usepackage{amsfonts}
\usepackage{amsmath}
\usepackage{amssymb}
\usepackage{enumitem}
\usepackage{colortbl}

\usepackage{makecell}
\usepackage{multicol}
\usepackage{multirow}
\usepackage{pifont}

\definecolor{red}{rgb}{0.8,0,0}  
\definecolor{green}{RGB}{0, 133, 21}  
\definecolor{grey}{rgb}{0.5,0.5,0.5}

\definecolor{w_1}{RGB}{52,204,204}
\definecolor{w_2}{RGB}{70,203,187}
\definecolor{w_3}{RGB}{95,202,161}
\definecolor{w_4}{RGB}{119,200,136}
\definecolor{w_5}{RGB}{155,199,101}
\definecolor{w_6}{RGB}{185,197,70}
\definecolor{w_7}{RGB}{227,194,28}
\definecolor{w_8}{RGB}{243,193,12}
\definecolor{w_9}{RGB}{255,192,0}

\makeatletter
\def\blfootnote{\xdef\@thefnmark{}\@footnotetext}
\DeclareRobustCommand\onedot{\futurelet\@let@token\@onedot}
\def\@onedot{\ifx\@let@token.\else.\null\fi\xspace}

\makeatother

\def\eqref#1{Equation~\ref{#1}}

\usepackage{soul}

\usepackage{titletoc}
\usepackage{caption}

\usepackage{float}
\usepackage[utf8]{inputenc}
\usepackage{hyperref}
\usepackage{amsmath, amssymb}
\usepackage{bm}
\usepackage{xcolor}
\usepackage{tikz}
\usepackage{arydshln}
\usetikzlibrary{arrows.meta, positioning, calc, fit, backgrounds, decorations.pathreplacing}
\usepackage{tabularx}
\usepackage{CJKutf8}
\usepackage{booktabs}
\usepackage{multirow}
\usepackage{graphicx}
\usepackage{listings}

\definecolor{headercolor}{RGB}{230,230,230}
\definecolor{easycolor}{RGB}{223,240,216}   
\definecolor{medcolor}{RGB}{255,242,204}    
\definecolor{hardcolor}{RGB}{248,215,218}   

\usepackage[most]{tcolorbox}
\tcbuselibrary{skins, breakable}

\definecolor{storybg}{RGB}{250, 248, 242}  

\newtcolorbox{storyquote}{
    colback=storybg,
    colframe=storybg,
    boxrule=0pt,
    sharp corners,
    breakable,
    left=10pt, right=10pt, top=8pt, bottom=8pt,
    fontupper=\itshape
}

\definecolor{tieravg}{HTML}{EDF1F7}   
\definecolor{grandavg}{HTML}{D9E2EF}   

\newcommand{\pernovelhead}{%
\toprule
 & \multicolumn{3}{c}{\textbf{Cinematic Presentation}}
 & \multicolumn{3}{c}{\textbf{Film Consistency}}
 & \multicolumn{3}{c}{\textbf{Novel Fidelity}} & \\
\cmidrule(lr){2-4}\cmidrule(lr){5-7}\cmidrule(lr){8-10}
{ID} & VP & NEP & AVP & CC & SC & OC & NHR & LR & SR & {Overall} \\
\midrule
}

\usepackage{titletoc}
\usepackage{hyperref}
\definecolor{niceblue}{RGB}{0, 82, 155}

\hypersetup{
    colorlinks=true,
    linkcolor=niceblue,      
    citecolor=niceblue
}

\title{FilmWorld: Agentic Novel-to-Film Generation through Dynamic Cinematic World Modeling}

\author[1,2]{Jialong Zuo}
\author[1]{Haotong Zuo}
\author[2]{Shiwei Zhang}
\author[2]{Xiang Wang}
\author[1]{Chen Li}
\author[1]{Nong Sang}
\author[1]{\\Changxin Gao}
\author[1]{Xiang Bai}
\affiliation[1]{Huazhong University of Science and Technology}
\affiliation[2]{Wan Team, Alibaba Group}

\abstract{
Translating novels into films is a grand challenge for generative artificial intelligence, requiring the conversion of abstract literary prose into long-form, multi-scene visual narratives. While current video generation models excel at producing short, single-scene clips within narrowly bounded temporal and spatial contexts, novel-to-film generation operates in a far more complex regime, demanding long-duration content across diverse scenes with dynamically evolving entity states. To address this, we formalize novel-to-film generation as \textbf{dynamic cinematic world modeling}, which we decompose into two phases: construction, which grounds abstract, underspecified literary narratives into concrete, stateful, and persistent world entities; and evolution, which governs how these entities dynamically update under plot progression to maintain causal consistency across scenes. To realize this formulation, we propose \textbf{FilmWorld}, an end-to-end agentic system in which two groups of specialized agents collaborate to instantiate these two phases. The construction-side agents perform narrative structured translation, world entity state modeling with visual anchoring, and state-driven shot planning, progressively projecting literary language into a cinematic blueprint. The evolution-side agents then perform state-anchored visual generation, cross-shot dynamic state propagation, and closed-loop state verification to maintain causal consistency and visual coherence. Furthermore, to address the evaluation gap in long-form generation, we introduce \textbf{FilmEval}, a systematic evaluation framework that couples a difficulty-graded benchmark of 15 representative novels with an automated protocol of nine objective metrics spanning three dimensions: cinematic presentation, film consistency, and novel fidelity. Extensive experiments demonstrate that FilmWorld consistently outperforms state-of-the-art video generation agent systems, with particularly pronounced improvements in narrative fidelity and cross-scene consistency. 
}

\metadata[
\raisebox{-0.2em}{\includegraphics[width=0.025\linewidth]{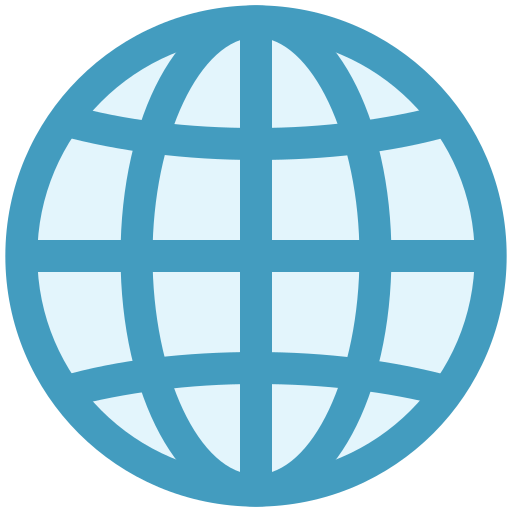}}~~Project Page]{\href{https://filmworld-ai.github.io/}{\texttt{https://filmworld-ai.github.io}}}

\begin{document}
\maketitle

\section{Introduction}
\label{sec:intro}
Novels offer rich textual narratives, whereas films provide highly expressive visual storytelling. Traditionally, adapting a novel into a film is a highly demanding, resource-intensive craft, requiring massive collaborative effort to manually translate abstract prose into concrete audiovisuals. Automating this process by generating films directly from novels holds immense value, promising to revolutionize content creation in creative industries while serving as a frontier challenge that drives AI advancements in language comprehension, visual synthesis, and long-form narrative coherence~\cite{elmoghany2025survey}.

Existing video generation research primarily focuses on short, single-scene clips over narrowly bounded temporal horizons, treating generation as a local visual synthesis problem~\cite{brooks2024sora,yang2024cogvideox,kong2024hunyuanvideo,wan2025wan}. In contrast, novel-to-film generation operates in a vastly more complex regime. It demands long-duration content across dozens of diverse scenes within a dynamically evolving world where all key entities, from characters to environments, continuously undergo state transitions. Moreover, the input consists of abstract literary prose rich in psychological depth, non-linear narration, and metaphor. Simply extrapolating current models to this task inevitably fails to maintain narrative and visual coherence over time~\cite{villegas2023phenaki,chen2025skyreelsv2,cui2025selfforcingpp}. Consequently, this narrative-level generation is not a mere extension of existing tasks, but a fundamentally new research challenge.

\begin{figure}[t]  
  \centering  
  \includegraphics[width=\textwidth]{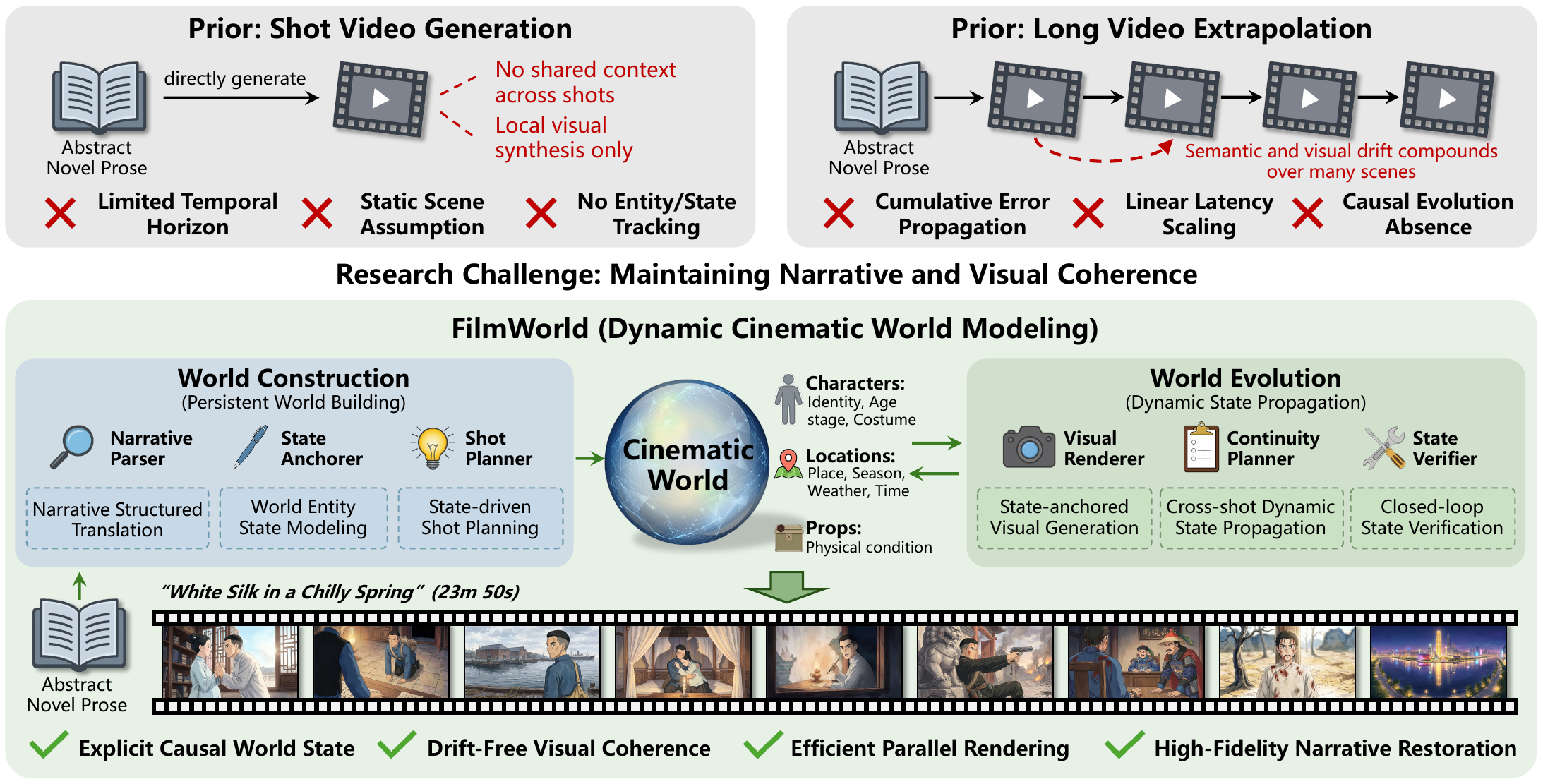}  
  \vspace{-6mm}
  \caption{\textbf{Conceptual comparison of video generation paradigms for novel-to-film generation.} Existing methods either synthesize shots independently without cross-shot state tracking (top-left) or extrapolate sequentially where semantic drift compounds over long narratives (top-right). FilmWorld (bottom) instead constructs a persistent cinematic world grounded in the source novel and evolves it through plot-driven state transitions, with construction-side and evolution-side agent groups jointly enforcing causal consistency and visual coherence across scenes.} 
  \label{fig1}  
\end{figure}

We argue that the essence of novel-to-film generation lies in dynamic cinematic world modeling, which operates through two core phases. The construction phase grounds the abstract, underspecified literary narrative into concrete, stateful, and persistent world entities. The evolution phase governs how this world dynamically updates under plot progression while maintaining causal consistency across multiple scenes. Consequently, every shot in the final film is simply a specific view rendered from this underlying world at a given moment. This conceptual reframing shifts the focus of long-form generation from optimizing isolated shots to building an evolving world, distinguishing our approach from prior per-shot synthesis paradigms.

Modeling a dynamic cinematic world is an inherently multi-faceted task spanning heterogeneous capabilities, which is more effectively addressed through specialized agent collaboration than through a single monolithic model~\cite{wu2025movieagent,hu2024storyagent,xie2025largemultimodal}. We therefore propose FilmWorld, an end-to-end agentic system in which multiple specialized agents collaborate to accomplish the novel-to-film pipeline. On the construction side, three agent groups work in concert to progressively project abstract literary language into an addressable and reusable cinematic world blueprint: (i) Narrative Structured Translation, (ii) World Entity State Modeling with Visual Anchoring, and (iii) State-driven Shot Planning. On the evolution side, three agent groups jointly enforce causal consistency and visual coherence across dozens of scenes: (i) State-anchored Visual Generation, (ii) Cross-shot Dynamic State Propagation, and (iii) Closed-loop State Verification with Correction. Together, the two sides form a unified construction-evolution mechanism that instantiates the full novel-to-film pipeline.

From an evaluation perspective, existing benchmarks primarily focus on short-form video quality and are ill-equipped for long-form cinematic narratives~\cite{wu2025moviebench,zhuang2026vistorybench,hua2025vabench,yang2026evalverse}. To address this, we introduce FilmEval, a systematic evaluation framework consisting of a curated benchmark of 15 representative source novels and an automated evaluation protocol. We stratify this benchmark into three difficulty levels (easy, medium, and hard) based on narrative richness, structural complexity, and temporal span. This design enables a fine-grained diagnosis of method robustness across varying narrative complexities. The evaluation protocol comprises nine objective metrics organized along three dimensions: cinematic presentation, film consistency, and novel fidelity. Crucially, these metrics systematically align with the core requirements of dynamic cinematic world modeling, collectively providing a principled and multi-dimensional characterization of the generated film quality.

Extensive experiments on diverse and challenging novel adaptation scenarios demonstrate that our FilmWorld consistently outperforms existing state-of-the-art video generation agent systems~\cite{xu2025mmstoryagent,zheng2025vgot,wu2025movieagent,huang2026vimax,videoclaw2026} across all evaluation dimensions, achieving particularly pronounced improvements in narrative fidelity and cross-scene consistency. Ablation studies further validate the effectiveness of each component in our system, confirming the necessity of the proposed construction and evolution mechanisms.

We summarize our contributions as follows:

\begin{itemize}
    \item At the problem level, we formalize novel-to-film generation as dynamic cinematic world modeling, decomposed into a construction phase that grounds abstract literary narratives into stateful world entities, and an evolution phase that governs how these entities update causally across scenes.

    \item At the method level, we propose FilmWorld, an end-to-end agentic system in which construction-side and evolution-side agent groups jointly instantiate and preserve the dynamic cinematic world, enabling coherent and high-quality long-form film generation from novels.

    \item At the evaluation level, we propose FilmEval, a systematic evaluation framework for long-form generated film assessment, comprising a difficulty-graded benchmark of 15 representative novels and a protocol of nine automatic metrics across three dimensions, providing a principled and comprehensive characterization of dynamic cinematic world quality.

\end{itemize}

\vspace{-3mm}
\section{Related Work}
\label{sec:related}
\subsection{Long Video Generation}
Foundation models such as Sora~\cite{brooks2024sora}, CogVideoX~\cite{yang2024cogvideox}, HunyuanVideo~\cite{kong2024hunyuanvideo}, Wan~\cite{wan2025wan}, and Kling~\cite{kling2024} generate high-quality single-shot clips but remain constrained to short temporal horizons. To reach minute-scale sequences, Phenaki~\cite{villegas2023phenaki} employs autoregressive token prediction, while Self-Forcing++~\cite{cui2025selfforcingpp} and Rolling Sink~\cite{li2026rollingsink} stabilize infinite-horizon inference. SkyReels-V2~\cite{chen2025skyreelsv2} integrates language models with structured pipelines for cinematic-length videos, and FilmWeaver~\cite{luo2026filmweaver} further stabilizes multi-shot generation through cache-guided autoregressive diffusion. For multi-scene storytelling, OneStory~\cite{an2026onestory}, InfinityStory~\cite{elmoghany2026infinitystory}, StoryMem~\cite{zhang2025storymem}, and EM-Vid~\cite{vandersanden2026emvid} maintain cross-scene consistency via memory-guided prediction, explicit state tracking, and entity-centric memory. Holocine~\cite{meng2026holocine} optimizes shot composition for holistic narratives, and ReCA~\cite{liu2026reca} recursively allocates context to preserve anchor states during multi-shot extrapolation. Despite this progress, these methods propagate context primarily through implicit hidden states or sliding-window attention, and thus suffer from compounding semantic drift when entities undergo dramatic, plot-driven transitions, making them ill-suited for the long-range narrative and visual continuity required in the novel-to-film generation task.

\subsection{Story Visualization}
Story visualization generates sequential images or videos from textual narratives. Early frameworks like StoryGAN~\cite{li2019storygan} and StoryDALL-E~\cite{maharana2022storydalle} pioneered this task by employing sequential generative networks and text-to-image retrofitting to preserve context across frames. With the rise of diffusion models, StoryDiffusion~\cite{zhou2024storydiffusion} enforces character consistency via cross-frame self-attention, while TaleCrafter~\cite{gong2023talecrafter} combines interactive layout planning with retrieval-augmented cross-attention to handle multi-character scenes. Beyond static storyboards, narrative video models capture complex temporal dynamics: VideoDirectorGPT~\cite{lin2024videodirectorgpt}, VideoGen-of-Thought~\cite{zheng2025vgot}, and StoryGPT-V~\cite{shen2025storygptv} employ large language models to decompose text into localized scripts and spatial layouts, VideoStudio~\cite{long2024videostudio} and ShotAdapter~\cite{kara2025shotadapter} incorporate entity tracking and temporal adapters for smooth transitions, and more recently STAGE~\cite{zhang2025stage} anchors multi-shot narratives on explicit storyboards while Narrative Weaver~\cite{yao2026narrativeweaver} pursues controllable long-range visual consistency via multi-modal conditioning. To evaluate these systems, comprehensive suites such as ViStoryBench~\cite{zhuang2026vistorybench}, VABench~\cite{hua2025vabench}, and EvalVerse~\cite{yang2026evalverse} have been established to measure character and scene consistency across sequences. However, existing methods face two key bottlenecks. First, they cannot bridge the semantic gap between raw, metaphorical literary prose and the explicit visual layouts they require as input. Second, they lack causal world modeling, treating characters as static visual templates rather than dynamic entities whose physical states and relationships evolve organically with the plot.

\subsection{Agentic Video Generation}
Multi-agent architectures decompose complex video generation into coordinated sub-tasks. Early systems like MovieAgent~\cite{wu2025movieagent}, StoryAgent~\cite{hu2024storyagent}, and ScriptAgent~\cite{mu2026scriptagent} employ hierarchical planning agents for scriptwriting, character design, and camera placement. For longer sequences, MAViS~\cite{wang2026mavis} and MUSE~\cite{sun2026muse} orchestrate closed-loop cognitive planning, while AniME~\cite{zhang2025anime} uses director-oriented decomposition for structured animations. For better continuity, VideoMemory~\cite{zhou2026videomemory} and MM-StoryAgent~\cite{xu2025mmstoryagent} leverage structured memories to track entities across shots, whereas VideoClaw~\cite{videoclaw2026} and ViMax~\cite{huang2026vimax} offer user-editable pipelines with dependency-aware planning. Pushing toward higher production quality, FilMaster~\cite{huang2025filmaster} and Co-Director~\cite{song2026codirector} embed cinematic principles into storytelling, and VISTA~\cite{long2025vista} refines generation through test-time self-improvement. Concurrent efforts like Shi et al.~\cite{shi2026onesentence} and Toonflow~\cite{toonflow2026} further explore specialized multi-agent workflows for drama generation. While effective, these systems operate primarily on scene-by-scene script snippets and lack a shared ``single source of truth'' to track state transitions, identity, and spatial continuity across the full film. FilmWorld addresses this gap by centralizing the construction and evolution of a dynamic world state as the persistent anchor that aligns all collaborative agents.

\vspace{-2mm}
\section{Problem Definition}
\label{sec:problem}

\vspace{-1.5mm}
\subsection{Task Formulation}

We define the \textit{Novel-to-Film Generation} task as follows. Given a source novel $\mathcal{N}$ written in literary prose, the goal is to produce a film $\mathcal{F}$ that faithfully renders the narrative content of $\mathcal{N}$ into a coherent, long-form audiovisual experience. Structurally, $\mathcal{F}$ is organized as a hierarchical, ordered sequence of scenes $\{S_1, S_2, \ldots, S_M\}$, where each scene $S_i$ consists of an ordered sequence of shots $\{v_i^1, v_i^2, \ldots, v_i^{K_i}\}$. Each shot $v_i^k$ is a video segment of several seconds, optionally paired with synchronized dialogue audio.

To balance narrative complexity against computational budget, we deliberately calibrate the research scope and execution boundary of this task. Although $\mathcal{N}$ can in principle be of arbitrary length and written in different languages, we focus on texts of moderate length, spanning roughly 1{,}000--5{,}000 Chinese characters or, equivalently, around 1000--3{,}000 English words. At this input scale, the idealized cinematic output $\mathcal{F}$ typically lasts 5--30 minutes and comprises 20--50 scenes and 50--300 shots.

This hierarchical structure (novel $\rightarrow$ scenes $\rightarrow$ shots) determines the granularity at which content is generated and coordinated. The central challenge lies not in the quality of any individual shot in isolation, but in maintaining narrative coherence and visual consistency across the \textit{entire} sequence of $N$ shots.

\vspace{-1.5mm}
\subsection{Why a Dynamic World Representation is Necessary}

A naive approach to novel-to-film generation would be to generate shots independently~\cite{zhou2024storydiffusion,lin2024videodirectorgpt,zheng2025vgot} or sequentially with limited context~\cite{villegas2023phenaki,cui2025selfforcingpp,chen2025skyreelsv2}. However, this inevitably fails at the scale of dozens of scenes due to a fundamental \textit{representation gap}: there exists no explicit mechanism to enforce that the visual state of an entity at shot $t$ is causally consistent with what happened to that entity in all preceding shots.

Consider a character who is injured in an earlier scene. Without an explicit state, subsequent scenes have no reliable way to ``know'' that this character should still appear injured. This information must either survive implicitly in an ever-growing context window~\cite{an2026onestory,zhang2025storymem,liu2026reca}, which degrades as the sequence grows longer, or be re-inferred from the source text for every shot, which is error-prone and computationally prohibitive.

This problem is compounded by two structural properties of the novel-to-film generation task:

\textbf{1) The world must be constructed, not given.} Unlike tasks where visual descriptions are provided as inpu, the entities of the cinematic world, including their identities, visual appearances, and spatial configurations, must be \textit{inferred and instantiated} from abstract literary language that often leaves them implicit or underspecified.

\textbf{2) The world evolves causally.} Entities do not remain static. Characters change costumes, age, get injured, or alter emotional states; locations shift in season, weather, and time of day; props are created, damaged, or destroyed. These transitions are driven by plot events and must be tracked explicitly to prevent state drift.

These two properties jointly necessitate an explicit, addressable, and causally evolving intermediate representation, which we term the \textit{Dynamic Cinematic World}.

\subsection{Formal Definition of the Dynamic Cinematic World}

To mathematically operationalize our framework, we formally define the \textit{Dynamic Cinematic World} as a system governed by the following five-tuple:
\begin{equation}
\mathcal{W} = \langle \mathcal{E}, \mathcal{X}, \Phi, \mathcal{T}, \mathcal{R} \rangle
\end{equation}
where each component and the state evolution at shot step $t \in \mathbb{N}$ are defined as follows:

\textbf{Entity Set ($\mathcal{E}$).} 
The world entities are defined as a disjoint union $\mathcal{E} = \mathcal{E}_c \cup \mathcal{E}_l \cup \mathcal{E}_p$, representing the set of characters $\mathcal{E}_c$, locations $\mathcal{E}_l$, and props $\mathcal{E}_p$, respectively. 

\textbf{State Space ($\mathcal{X}$).} 
The state space is the union of specialized attribute spaces $\mathcal{X} = \mathcal{X}_c \cup \mathcal{X}_l \cup \mathcal{X}_p$. 
At any shot step $t$, the global world state $W_t$ is defined as the set of state assignments for all entities:
\begin{equation}
W_t = \{ (e, s_{e, t}) \mid e \in \mathcal{E} \}
\end{equation}
where $s_{e, t} \in \mathcal{X}$ represents the concrete state of entity $e$ at step $t$. For notation convenience, we denote $W_t[e] = s_{e, t}$ as the state lookup for entity $e$. Specifically, for a character $e \in \mathcal{E}_c$, its state $W_t[e] \in \mathcal{X}_c$ encodes attributes such as age stage, costume, emotion, and physical condition; for locations, it tracks environmental factors like weather and lighting; for props, it tracks physical status and spatial placement. The configuration space of all possible world states is denoted as $\mathbf{\Omega} \subseteq \mathcal{X}^{|\mathcal{E}|}$.

\textbf{State Identifier ($\Phi$).} 
The state mapping function $\Phi: \mathcal{X} \rightarrow \mathcal{I}$ assigns each concrete state in $\mathcal{X}$ a unique, addressable identifier $id \in \mathcal{I}$. For an entity $e$ at step $t$, its active visual identity is indexed by:
\begin{equation}
\phi_{e, t} = \Phi(W_t[e])
\end{equation}
This identifier serves as the atomic indexing unit across our entire architecture: all visual assets, reference images, and rendering constraints are anchored to these state identifiers rather than to raw entity names.

\textbf{Transition Function ($\mathcal{T}$).} 
The transition function $\mathcal{T}: \mathbf{\Omega} \times \mathcal{P} \rightarrow \mathbf{\Omega}$ governs how the world state evolves, where $\mathcal{P}$ is the space of plot events. Given the current world state $W_t$ and a plot event $p_t \in \mathcal{P}$ extracted from the narrative prose, the transition function yields the updated world state:
\begin{equation}
W_{t+1} = \mathcal{T}(W_t, p_t)
\end{equation}
This formalizes causal plot-driven changes, such as a character sustaining an injury, changing costume, or a location transitioning from day to night.

\textbf{Rendering Function ($\mathcal{R}$).} 
The rendering function $\mathcal{R}: \mathbf{\Omega} \times \mathcal{D} \rightarrow \mathcal{V}$ synthesizes a video shot $v_t \in \mathcal{V}$ given the world state $W_t$ and a shot directive $d_t \in \mathcal{D}$, where $\mathcal{D}$ represents shot parameters (such as framing, camera intent, participating entities, and temporal dynamics). This process is formally expressed as:
\begin{equation}
v_t = \mathcal{R}(W_t, d_t)
\end{equation}
During synthesis, $\mathcal{R}$ resolves the visual appearance of each participating entity $e$ by querying its corresponding state identifier $\Phi(W_t[e])$, structurally guaranteeing cross-shot visual consistency.

\subsection{Reformulation of Novel-to-Film Generation}
\label{subsec:reformulation}

Under this formalization, the novel-to-film generation task naturally decomposes into two sub-problems:

\textbf{World Construction.} 
Given the source novel text $\mathcal{N}$, the objective is to construct the initial world state $W_1$ (which instantiates the entity set $\mathcal{E}$ and their initial states $W_1[e]$), generate their corresponding state identifiers with anchored visual reference assets, and plan the sequential execution narrative. Formally, this construction step translates the novel into the initial state and two aligned operational sequences:
\begin{equation}
\mathcal{N} \xrightarrow{\text{Construction}} W_1, \{p_1, p_2, \ldots, p_N\}, \{d_1, d_2, \dots, d_N\}
\end{equation}
where $\{p_t\}$ represents the sequence of plot events and $\{d_t\}$ represents the sequence of shot directives.

\begin{figure}[t]
  \centering
  \includegraphics[width=0.76\textwidth]{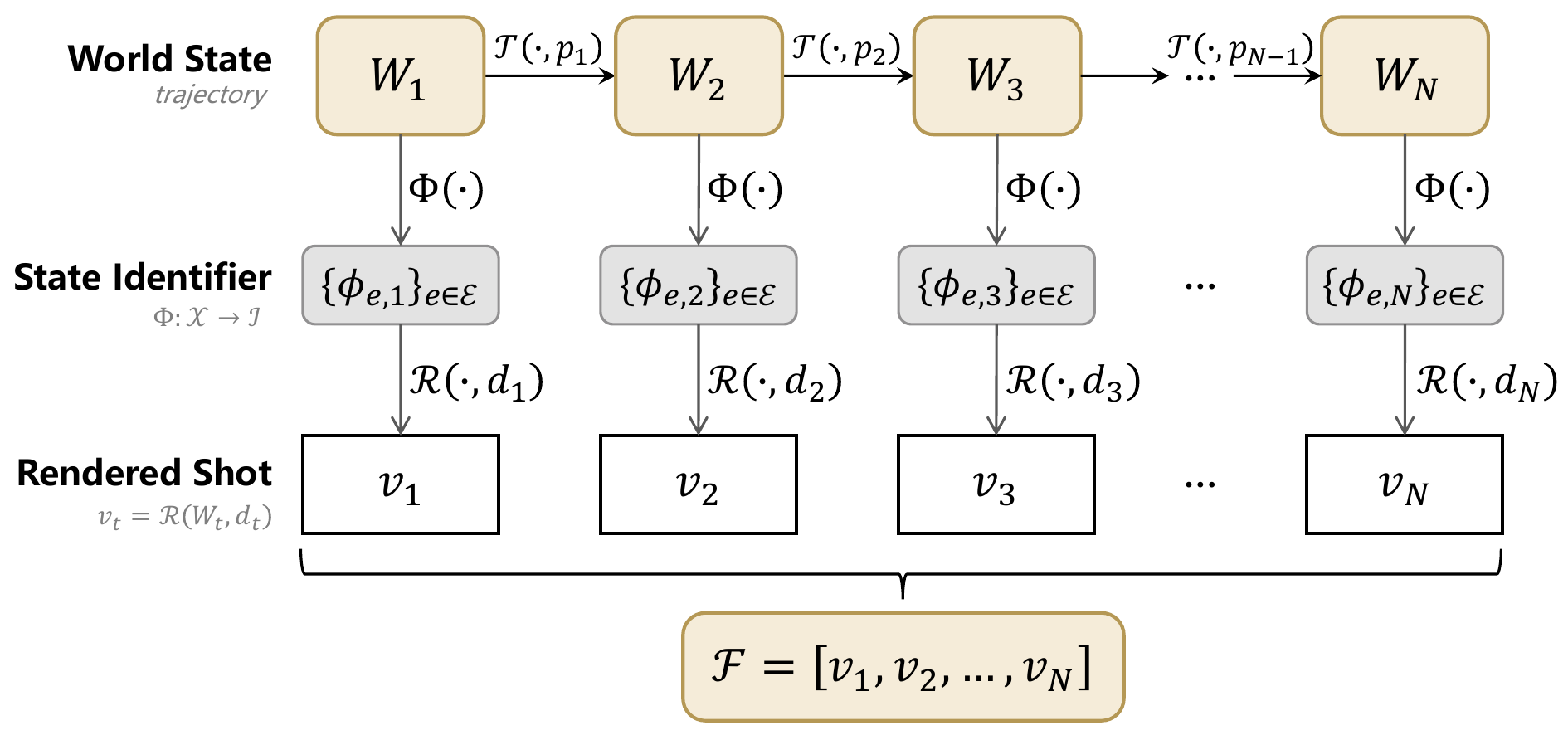}
  \caption{\textbf{Schematic of the Dynamic Cinematic World evolution.}
  At each shot step~$t$, the world state~$W_{t}$ is advanced to~$W_{t+1}$
  via the transition function~$\mathcal{T}(\cdot,\,p_t)$ conditioned on
  the plot event~$p_t$.
  The state mapping function~$\Phi(\cdot)$ maps each world state to a set
  of entity-level state identifiers~$\{\phi_{e,t}\}_{e\in\mathcal{E}}$,
  which are then passed to the rendering function~$\mathcal{R}(\cdot,\,d_t)$
  together with the shot directive~$d_t$ to produce the visual shot~$v_t$.}
  \label{fig:evo}
\end{figure}

\textbf{World Evolution.} 
This problem sequentially generates each shot while maintaining dynamic world consistency. As shown in Figure~\ref{fig:evo}, at each shot step $t \in \{1, 2, \ldots, N\}$, the system renders the current shot using the active world state, and then applies the transition function to advance the state to guide the next step:
\begin{equation}
v_t = \mathcal{R}(W_t, d_t), \quad W_{t+1} = \mathcal{T}(W_t, p_t)
\end{equation}

The final film $\mathcal{F}$ is produced by the ordered temporal concatenation of all video shots: $\mathcal{F} = [v_1, v_2, \ldots, v_N]$.

This formulation makes explicit what prior work leaves implicit~\cite{wu2025movieagent,huang2026vimax,videoclaw2026}: the existence of a persistent, evolving world state that mediates between the abstract source narrative and the generated visual output. The Method section that follows describes how our proposed framework, FilmWorld, instantiates each component of this theoretical formulation through coordinated multi-agent collaboration.

\section{Methodology}
\label{sec:method}

\begin{figure}[t]
  \centering
  \includegraphics[width=\textwidth]{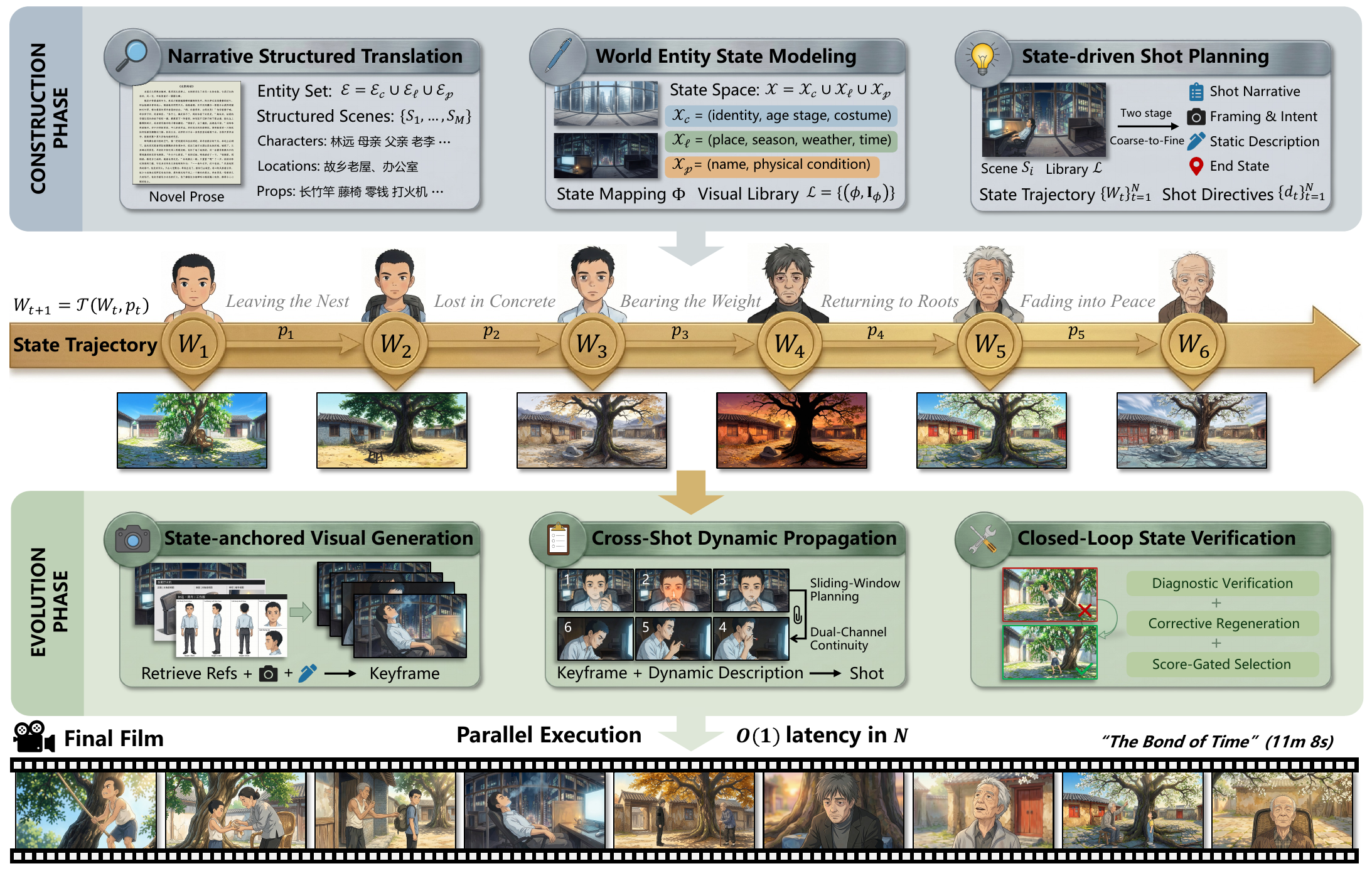}
  \caption{\textbf{Overview of the FilmWorld framework.} Using a simplified world-state trajectory from the novel \textit{``The Bond of Time''} for illustrative clarity, FilmWorld translates raw prose into films via a decoupled paradigm. In the \textit{Construction Phase}, we parse narrative structures, anchor entity states to visual references and plan the global trajectory of states and shots. This trajectory explicitly tracks how consecutive plot events drive entity state updates. In the \textit{Evolution Phase}, we render state-guided keyframes, propagate dynamic motion via sliding windows, and resolve discrepancies via closed-loop feedback. This decoupling enables parallel rendering with $O(1)$ theoretical latency.}
  \label{fig:overview}
\end{figure}

\subsection{Overview}
\label{subsec:overview}

We propose \textit{FilmWorld}, an end-to-end multi-agent framework that instantiates the Dynamic Cinematic World formalized in Section~\ref{sec:problem}. As illustrated in Figure~\ref{fig:overview}, FilmWorld decomposes novel-to-film generation into two phases mirroring the construction--evolution paradigm (Section~\ref{subsec:reformulation}). The \textit{Construction Phase} translates the novel $\mathcal{N}$ into a trajectory of world states $\{W_t\}_{t=1}^{N}$, aligned plot events $\{p_t\}_{t=1}^{N}$, and shot directives $\{d_t\}_{t=1}^{N}$. The \textit{Evolution Phase} then renders each shot $v_t = \mathcal{R}(W_t, d_t)$ under explicit consistency constraints. Each phase is orchestrated by three specialized agent groups, coordinating six distinct roles in total.

A key architectural feature of FilmWorld is the decoupling of state evolution from visual rendering. Although state evolution is formulated recursively as $W_{t+1} = \mathcal{T}(W_t, p_t)$, FilmWorld resolves this recurrence \textit{entirely within the symbolic Construction Phase}. By fully materializing the state trajectory $\{W_t\}_{t=1}^{N}$ prior to pixel-level generation, rendering operations in the Evolution Phase are freed from sequential temporal dependencies. This design bypasses cumulative drift and enables fully parallelized shot synthesis. We elaborate on the efficiency and scheduling implications of this decoupling in Section~\ref{subsec:discussion}.

\subsection{Construction Phase}
\label{subsec:construction}

The Construction Phase progressively projects abstract literary language into a structured, addressable, and visually anchored cinematic world. It comprises three agent groups operating in cascade: the first instantiates the entity set $\mathcal{E}$; the second defines the state space $\mathcal{X}$, mapping function $\Phi$, and reference library $\mathcal{L}$; the third applies $\mathcal{T}$ to materialize the full state trajectory $\{W_t\}_{t=1}^{N}$ and shot directives $\{d_t\}_{t=1}^{N}$, resolving all temporal dependencies prior to rendering.

\subsubsection{Narrative Structured Translation}
\label{subsubsec:nst}

This agent group is responsible for defining the global entity set $\mathcal{E}$ and recovering the concrete physical details that literary prose typically leaves implicit. It operates in two coordinated stages:

\textbf{Chapter Segmentation and Entity Resolution.}
First, the input novel $\mathcal{N}$ is segmented into an ordered list of chapters. This partitioning bounds the context window for downstream agents and establishes a manageable scope for tracking story consistency. Over this chapter sequence, an entity resolution agent identifies and clusters referential variations of characters, locations, and props. Because novels routinely refer to the same entity using aliases, honorifics, pronouns, or descriptors (e.g., referring to the same character as ``the detective'' or ``Sherlock''), failing to resolve these would cause downstream generators to treat them as distinct characters, fragmenting the visual identity. This stage assigns a canonical identifier to each unique entity, populating the global set $\mathcal{E} = \mathcal{E}_c \cup \mathcal{E}_l \cup \mathcal{E}_p$ to ensure identity consistency across the entire narrative.

\textbf{Scene Structuring and Cinematic Detail Recovery.}
Next, each chapter is further segmented into an ordered sequence of scenes $\{S_1, \dots, S_M\}$, where each scene represents a self-contained narrative unit in a specific time and place. Beyond simple segmentation, this stage bridges the gap between abstract language and concrete imagery by recovering essential cinematic details omitted in prose. For instance, while a novel might simply write ``she returned home at dusk,'' a film must commit to explicit visual choices: the specific warm-toned lighting of twilight, a particular season, weather conditions, and an atmospheric color palette. The structuring agent infers these parameters and attaches them as structured attributes to each scene, establishing the physical and aesthetic foundation for subsequent visual generation.

\subsubsection{World Entity State Modeling with Visual Anchoring}
\label{subsubsec:wesm}

This agent group instantiates the state space $\mathcal{X}$, the state mapping function $\Phi$, and the visual asset reference library $\mathcal{L}$ that anchors identifiers to concrete generative resources. 

\textbf{State Space Discretization.}
We discretize each entity class into key attributes that capture essential narrative variations. Specifically, characters are parameterized along three dimensions: identity, age stage, and costume, which represent the main sources of visual changes across scenes. Locations are defined by four dimensions: place, season, weather, and time of day. Props are parameterized by their physical condition. For each entity $e$ at shot step $t$, the agent assigns a concrete state $W_t[e] \in \mathcal{X}$ within this structured space.

\textbf{State Identifier Realization.}
The state mapping function $\Phi$ is implemented as a deterministic structured hash over the discretized attribute tuple. For a character $e \in \mathcal{E}_c$ with state $(\text{identity}, \text{age stage}, \text{costume})$, the identifier $\phi_{e,t} = \Phi(W_t[e])$ is uniquely determined by this attribute tuple; similar hashes are applied to locations and props. Under this formulation, two states are visually identical if and only if they share the exact same identifier. This unique mapping transforms the open-ended consistency challenge into a closed-form retrieval and matching problem.

\textbf{First-Appearance Visual Anchoring.}
For each unique state identifier $\phi$ encountered along the planned narrative trajectory, a reference visual asset $\mathbf{I}_\phi$ is generated at its first appearance and stored in a central library $\mathcal{L} = \{(\phi, \mathbf{I}_\phi)\}$. Any subsequent shot featuring this identical state retrieves $\mathbf{I}_\phi$ as a generation guide rather than creating a new asset from scratch, following the reference-guided consistency principle explored in prior story generation work~\cite{zhou2024storydiffusion,kara2025shotadapter,vandersanden2026emvid}. For audio consistency, we apply a similar first-appearance principle but isolate the auditory modality from visual-only attributes like costume. Specifically, a persistent voice profile is anchored strictly to the combination of the character's identity and active age stage, ensuring that their vocal identity remains stable across scenes while naturally reflecting physical aging. This strategy ensures that the computational cost of maintaining multi-modal consistency remains constant relative to the number of shots, decoupling long-range coherence from the film's total length.

\subsubsection{State-driven Shot Planning}
\label{subsubsec:ssp}

This agent group applies $\mathcal{T}$ to materialize the full state trajectory $\{W_t\}_{t=1}^{N}$ and produces the corresponding shot directives $\{d_t\}_{t=1}^{N}$ that guide the Evolution Phase. Each directive consists of a per-shot static description and end-state metadata defining the visual layout at the shot's terminal keyframe, with the opening configuration of shot $t$ derived from the terminal state of shot $t{-}1$, resolving all temporal dependencies prior to rendering. To manage the cross-modal complexity of translating narrative scenes into well-composed shot sequences, we employ a two-stage, coarse-to-fine planning workflow.

\textbf{Scene-level Narrative Planning.}
For each scene $S_i$, a planning agent decomposes the narrative into an ordered sequence of shot proposals based on plot functionality, in line with prior LLM-guided shot planning approaches~\cite{lin2024videodirectorgpt,zheng2025vgot,wu2025movieagent}. Each proposal contains a \textit{shot narrative}, a working title, the set of participating characters and props, and a rationale justifying its cinematic necessity. Operating in a text-only, unimodal manner, this stage achieves high concurrency by intentionally avoiding low-level visual details, producing a robust narrative skeleton for the next stage.

\textbf{Scene-conditioned Visual Realization.} A multimodal refinement agent then processes these proposals alongside a visual conditioning bundle compiled from $\mathcal{L}$. For each participating entity $e$, including the location, active characters, and key props, its reference asset is retrieved using the active state identifier $\phi_{e,t} = \Phi(W_t[e])$. Conditioned jointly on these references and the scene-level narrative context, the agent outputs the final camera framing, camera intent, \textit{static description}, and \textit{end-state metadata} for each shot. 

The end-state metadata encodes the terminal world state of each shot, from which the subsequent shot's opening configuration is derived, thereby realizing $\mathcal{T}$ and materializing the full state trajectory $\{W_t\}_{t=1}^{N}$ entirely at the symbolic level before rendering begins. Critically, all shots within a scene are planned simultaneously rather than in isolation, ensuring cross-shot contextual coherence. Since this stage focuses on static, keyframe-level layouts, temporal motion synthesis is deferred to the Evolution Phase (Section~\ref{subsubsec:csdp}).

\vspace{-1.5mm}
\subsection{Evolution Phase}
\label{subsec:evolution}

Given the materialized state trajectory $\{W_t\}_{t=1}^N$ and shot directives $\{d_t\}_{t=1}^N$, the Evolution Phase realizes the rendering function $\mathcal{R}$ under explicit cross-shot consistency constraints. It comprises three agent groups operating in coordinated parallelism over the shot sequence.

\subsubsection{State-anchored Visual Generation}
\label{subsubsec:savg}

This agent group renders each shot's keyframe image, realizing the static portion of the rendering function $\mathcal{R}$. For each shot $t$, let $\mathcal{E}_t \subset \mathcal{E}$ denote the set of participating entities specified in $d_t$. The agent retrieves the corresponding reference assets $\{\mathbf{I}_{\phi_{e,t}}\}_{e \in \mathcal{E}_t}$ from the library $\mathcal{L}$ using active state identifiers $\phi_{e,t} = \Phi(W_t[e])$. It then compiles these assets alongside the shot's static description and camera directives into a structured conditioning bundle, synthesizing the keyframe image through a multimodal generator. This keyframe serves as a visual contract for the shot, committing to entity identities, costumes, and spatial composition, and acting as the spatial anchor against which subsequent video synthesis (Section~\ref{subsubsec:csdp}) is conditioned. Additionally, we attach explicit entity-specific attribution tags to each reference image, ensuring the generator binds visual attributes correctly to their respective subjects.

\begin{figure}[t]
  \centering
  \includegraphics[width=\textwidth]{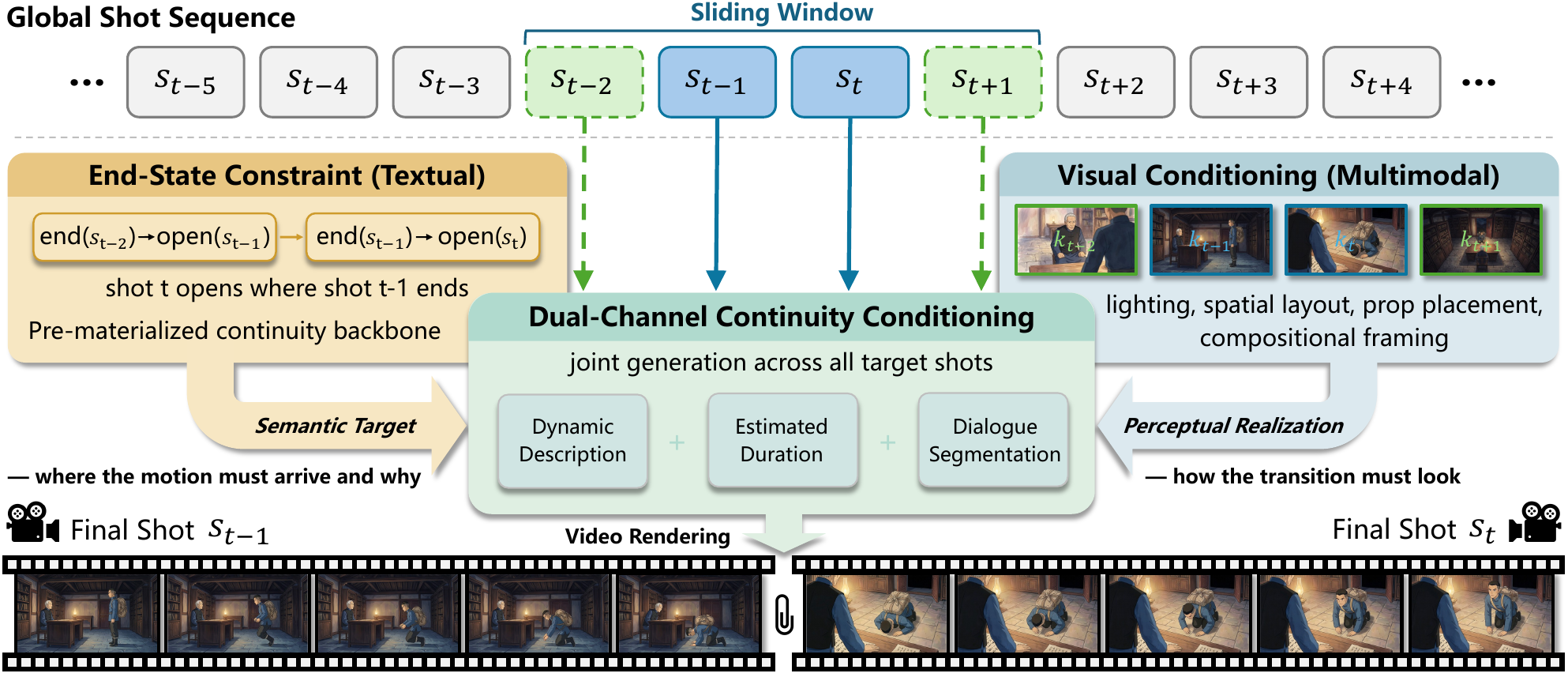}
  \caption{\textbf{Illustration of the Cross-Shot Dynamic State Propagation mechanism.} A scene-local sliding-window planner jointly synthesizes motion specifications for all target shots, enforcing cross-shot coherence via: 1) a pre-materialized textual end-state constraint, forming a continuity backbone where each shot's opening derives from its predecessor's terminal state; and 2) a visual conditioning channel supplying rendered keyframes as ordered multimodal inputs, constraining the perceptual realization of each transition in lighting, composition, spatial layout and so on.}
  \label{fig:window}
\end{figure}

\subsubsection{Cross-Shot Dynamic State Propagation}
\label{subsubsec:csdp}

This agent group synthesizes the temporal motion specification, consisting of the dynamic description and dialogue segmentation, for each shot and renders the final video segment $v_t$. Because motion cannot be planned in isolation, physical properties like pose continuity and causal plot consequences must persist seamlessly across shot boundaries. For instance, the opening pose of shot $t$ must coincide with the terminal configuration of shot $t{-}1$, and physical states, such as a character who has just fallen remaining on the ground, must remain consistent. We resolve these temporal dependencies using a scene-local sliding-window planner that operates on two pre-materialized continuity signals: textual end-state constraints and visual keyframe conditioning. This design enables fully parallel window execution, bypassing sequential inter-window dependencies.

\textbf{Scene-Local Sliding-Window Planning.}
Within each scene, the shot sequence is partitioned into overlapping windows. Each window comprises $g$ contiguous \textit{target} shots, for which dynamic descriptions are generated, padded by up to $c$ \textit{context} shots on each side. Drawing context strictly from the same scene prevents cross-scene conditioning leakage. A multimodal planning agent receives all shots in the window, including their rendered keyframe images labeled with explicit role annotations (target vs.\ context) and per-shot metadata. The agent then jointly produces for every target shot a timestamped dynamic description detailing segment-by-segment actions, an estimated total duration, and a structured dialogue segmentation. This joint generation allows the agent to coordinate rhythm and kinematic transitions across consecutive targets, while the surrounding context anchors these transitions in visually observed boundary states.

\textbf{Dual-Channel Continuity Conditioning.}
Cross-shot consistency is enforced through two channels that together constrain both the target state the motion must reach and the visual realization of the boundary frames:
\begin{enumerate}[leftmargin=*,itemsep=2pt]
\item[1)] \textit{End-state constraint (textual):} Each shot directive $d_t$ carries a pre-computed end-state specification produced during the upstream shot-planning stage, prescribing the terminal visual and spatial configuration of the shot. These end-states are planned with explicit sequential chaining, where shot $t$'s opening frame is derived from shot $t{-}1$'s end-state, thereby forming a pre-materialized continuity backbone across the narrative scene. The planning agent is mandated to terminate its dynamic description in a configuration consistent with this prescribed end-state, transforming cross-shot coherence from an emergent property of generation into an enforced constraint.
\item[2)] \textit{Visual conditioning (multimodal):} The keyframe images of all shots in the window, targets and context alike, are provided as ordered multimodal inputs with positional annotations. These images carry spatial-layout and appearance details, such as lighting, prop placement, character-to-camera distance, and compositional framing, that textual end-state descriptions alone cannot fully specify. The agent can directly observe the realized visual state at the window boundaries and plan motion that is spatially and photographically continuous with the surrounding shots.
\end{enumerate}
These two channels are complementary: the textual channel constrains the \textit{semantic target} of each transition, specifying where the motion must arrive and why, while the visual channel constrains its \textit{perceptual realization}, defining how the transition must look. Their joint conditioning yields transitions that are simultaneously narratively faithful and visually seamless, without requiring any sequential coupling between windows.

\textbf{Video Rendering.}
Finally, given the keyframe image and the planned dynamic description, a video generator synthesizes the final video segment $v_t$ for each shot. Because the dynamic planning stage resolves all temporal and spatial dependencies offline, individual segments are rendered fully in parallel once the planning completes.

\subsubsection{Closed-Loop State Verification with Correction}
\label{subsubsec:cosv}

This agent group enforces consistency between the rendered output and the prescribed world state through a feedback-guided verification loop. Operating on both keyframe images and video segments, this loop executes a structured diagnose-correct-select cycle to systematically resolve generation discrepancies.

\textbf{Diagnostic Verification.}
For each generated keyframe, a vision-language verifier evaluates visual conformance against its static specifications and reference assets, assessing core dimensions such as identity consistency, spatial composition, and semantic alignment. Similarly, for each video segment $v_t$, a temporal verifier assesses conformance against the dynamic description and end-state constraints in $d_t$, checking for core dimensions such as temporal continuity, dialogue alignment, and visual defects. Both verifiers output structured diagnostic reports containing multi-dimensional quality metrics and localized error descriptions, rather than simple binary judgments.

\textbf{Corrective Regeneration.}
When a verifier signals a regeneration command, it emits targeted corrective guidance to condition the subsequent attempt. For keyframe images, the verifier produces a set of \textit{correction points}, which are actionable, imperative-form instructions injected directly into the prompt alongside the reference assets. For video segments, it generates a complete \textit{repaired dynamic description} with adjusted timing intervals and corrected duration hints. This targeted approach ensures that regeneration is driven by a precise diagnostic loop rather than stochastic trial-and-error.

\textbf{Score-Gated Selection.}
Upon regeneration, the new candidate is re-evaluated. The system retains the superior version by comparing evaluation scores, prioritizing content completion rate for video segments. To prevent quality regression, the system rolls back to the previous best candidate if the regenerated output shows no improvement. This cycle repeats for up to $K$ rounds, with each iteration's diagnostic report informing the next correction attempt.

This verification process is executed independently per shot, preserving the parallel efficiency of the Evolution Phase. Structurally, this closed-loop design bounds individual deviations within the verifier's tolerance, preventing the cumulative error propagation that plagues open-loop pipelines over long cinematic sequences.

\subsection{Parallelism through Explicit State Externalization}
\label{subsec:discussion}

Long-form video generation is conventionally treated as an inherently sequential process. We demonstrate that this sequential bottleneck is merely an artifact of implicit state representations rather than an intrinsic property of the task. By externalizing the world state into a pre-materialized trajectory, we reduce the theoretical latency of the runtime-dominant Evolution Phase from $O(N)$ to $O(1)$.  

\noindent\textbf{The Sequential Bottleneck of Implicit States.}
Autoregressive world models and sequential diffusion pipelines~\cite{villegas2023phenaki,cui2025selfforcingpp,chen2025skyreelsv2} embed state within the generation trajectory: rendering shot $t$ requires the realized output of shot $t{-}1$ as conditioning input, forming a global causal chain. With per-shot rendering cost $\tau$ and $N$ total shots:
\begin{equation}
  L_{\text{seq}} = N \cdot \tau = O(N).
  \label{eq:seq_latency}
\end{equation}

\noindent\textbf{Parallel Latency under Explicit State Externalization.}
In contrast, our formulation pre-materializes the world-state trajectory $\{W_t\}_{t=1}^{N}$ offline by applying the transition function $\mathcal{T}$ to the source text, requiring no rendered outputs. The Evolution Phase then executes three stages in sequence, with each internally parallelized across all $N$ shots:

1) \textit{Keyframe rendering:} Each shot independently queries the reference library $\mathcal{L}$ using its pre-assigned state identifiers. All $N$ keyframe images are rendered in parallel, yielding a constant latency $\tau_{\text{img}}$.

2) \textit{Dynamic planning:} Scene-local sliding windows of bounded size $w = g + 2c \ll N$ consume only pre-materialized signals. All $\lceil N/g \rceil$ windows execute in parallel, yielding a planning latency $\tau_{\text{plan}}(w)$.

3) \textit{Video rendering:} Each video segment $v_t$ is synthesized from its keyframe image and its dynamic description independently. All $N$ segments are rendered in parallel, yielding a constant latency $\tau_{\text{vid}}$.

The total latency of the Evolution Phase is bounded by:
\begin{equation}
  L_{\text{evo}} = \tau_{\text{img}} + \tau_{\text{plan}}(w) + \tau_{\text{vid}} = O(1) \text{ in } N,
  \label{eq:evo_latency}
\end{equation}
since none of the individual terms depend on the sequence length $N$. Compared to Eq.~\ref{eq:seq_latency}, this yields a theoretical speedup of $\Theta(N)$. In practice, available compute is finite. Given $P$ parallel workers, the batched latency scales as $O(N/P)$, which remains substantially sub-linear in $N$ for any realistic $P$.

\noindent\textbf{The State Decoupling Condition.}
This reduction in complexity holds if and only if the information set $\mathcal{I}_t$ consumed by the renderer at shot $t$ contains no rendered outputs from other shots. Sequential methods violate this by requiring $\text{output}(t{-}1) \in \mathcal{I}_t$. In contrast, our pipeline guarantees that:
\begin{equation}
  \mathcal{I}_t \subseteq \{W_t,\, \mathcal{L},\, d_t,\, \{k_j\}_{j \in \mathcal{J}_t}\},
\end{equation}
where $k_j$ represents the keyframe image of shot $j$, $\mathcal{J}_t$ denotes the sliding window context around shot $t$. This decoupling stems from explicit state externalization; any long-form generation system satisfying this condition naturally inherits the same parallel speedup, regardless of its underlying generative model backbones. 

\noindent\textbf{Scope of the Analysis.} We deliberately confine our latency analysis to the Evolution Phase, since it dominates the end-to-end runtime of novel-to-film generation. Each shot in this phase invokes heavy image and video generators, whose per-shot wall-clock is orders of magnitude larger than the LLM-driven symbolic operations of the Construction Phase, which scale with source text length rather than with the far larger shot count $N$. Therefore, parallelizing the Evolution Phase captures the vast majority of achievable end-to-end speedup, and the residual Construction latency constitutes a negligible fraction of total runtime in practice.

\section{Evaluation}
\label{sec:filmeval}
\subsection{Overview}
Existing video generation benchmarks such as MovieBench~\cite{wu2025moviebench}, ViStoryBench~\cite{zhuang2026vistorybench}, and MSAVBench~\cite{wei2026msavbench} target short-form clip quality or multi-shot audio-video consistency, but do not assess capabilities critical to novel-to-film generation: narrative fidelity to a source text, long-range entity state coherence under plot-driven evolution, or robustness across narrative complexities. To address these gaps, we introduce FilmEval, comprising two components: (i) a difficulty-graded benchmark of 15 source novels stratified into easy, medium, and hard tiers by character count, plot complexity, and world-building richness; and (ii) nine automatic metrics across three dimensions, namely Cinematic Presentation, Film Consistency, and Novel Fidelity, providing multi-perspective evaluation of generated film quality that prior benchmarks leave unassessed.

\subsection{Benchmark Dataset}
FilmEval contains a difficulty-graded dataset of 15 source novels spanning diverse genres, including everyday life, historical fiction, revolutionary romance, and comedic fantasy. The dataset comprises 6 English novels and 9 Chinese novels, drawn from both adapted classical literature (e.g., \textit{The Necklace}, \textit{The Man in a Case}, \textit{The Gift of the Magi}) and originally created contemporary fiction (e.g., \textit{The Bond of Time}, \textit{White Silk in a Chilly Spring}, \textit{Huaqiang Buying Watermelons}), ensuring broad coverage of narrative styles, cultural settings, and authorship origins. To enable systematic diagnosis of method robustness across narrative complexities, the dataset is stratified into three tiers. Easy-tier novels feature few characters, linear plots, and concrete scenes. Medium-tier novels introduce more character interactions, scene transitions, and moderate causal dependencies. Hard-tier novels present complex world-building, multiple characters, long-range plot arcs, and dense event structures. This stratification allows fine-grained analysis of how each method degrades as narrative complexity scales, rather than relying on a single aggregate score.

\begin{figure}[htb]
  \centering
  \includegraphics[width=0.98\linewidth]{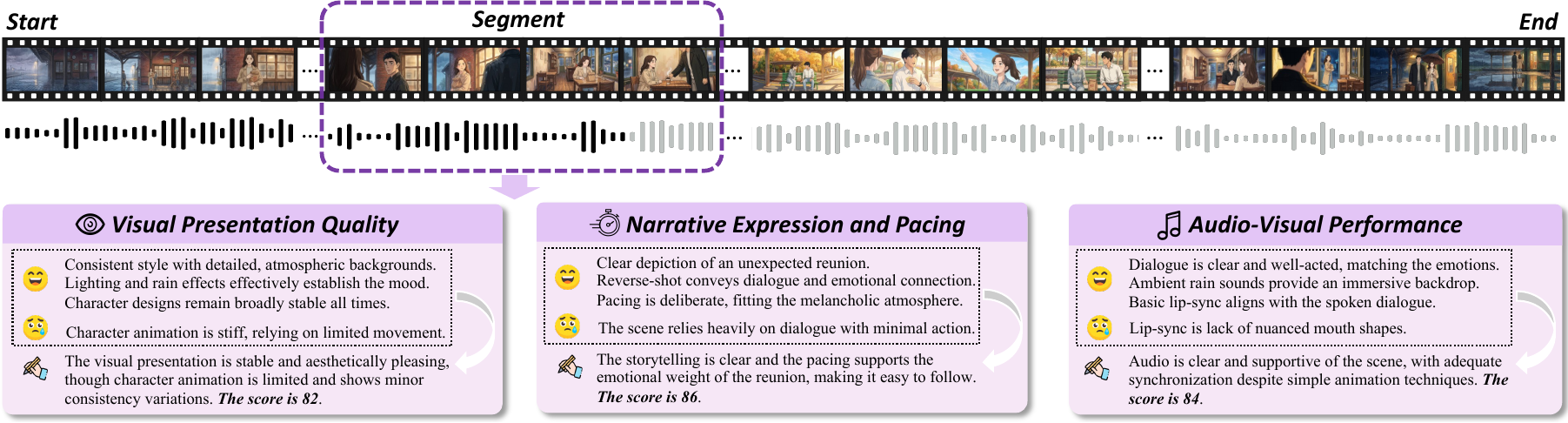}
  \caption{\textbf{Illustration of the cinematic presentation evaluation.} Given a generated film, FilmEval partitions it into temporal segments and assesses each segment along three dimensions: Visual Presentation Quality (VP), Narrative Expression and Pacing (NEP), and Audio-Visual Performance Quality (AVP). For every dimension, the evaluator identifies both positive evidence and visible defects, and then assigns a fine-grained score.}
  \label{fig:artisty}
\end{figure}

\vspace{-3mm}
\subsection{Evaluation Metrics}
\noindent\textbf{Overview.} FilmEval assesses generated films along three complementary dimensions::
\begin{description}[
    leftmargin=1.5em,
    labelwidth=1.0em,
    labelsep=0.5em,
    itemsep=0pt,
    topsep=0pt,
    parsep=0pt,
    partopsep=0pt,
    nosep,
    before=\vspace{-0.5em},
    after=\vspace{-0.3em},
    font=\normalfont
]
\item [1)] \textit{Cinematic Presentation} quantifies the film-level production quality of the generated output, encompassing visual aesthetics, shot composition, atmospheric rendering, and overall perceptual fidelity.

\item [2)] \textit{Film Consistency} evaluates the coherence of characters, scenes, motions, and fine-grained visual details across shots and along the temporal axis.

\item [3)] \textit{Novel Fidelity} measures the faithfulness of the generated film to the source novel in terms of characters, events, actions, and narrative structure.
\end{description}

All the metrics are implemented as fully automated evaluators by prompting a multimodal large language model, namely \textit{Gemini 3.1 Pro}~\cite{team2023gemini}, with carefully engineered task-specific instructions that jointly consume the source novel and the generated film segments. More details are provided in the Appendix.

\noindent\textbf{Cinematic Presentation.} This dimension assesses whether the generated film functions as a coherent cinematic artifact, focusing on perceptual quality, narrative pacing, and audio-visual expression.

\begin{enumerate}[leftmargin=*,itemsep=1pt]
\item[1)] \textit{Visual Presentation Quality $({VP})$  :}  ${VP}$ measures the visual stability, plausibility, and aesthetic appeal of generated frames. It penalizes prominent artifacts such as blur, anatomical distortions, warped faces and hands, geometric collapse, object popping, lighting inconsistency, ghosting, and physically implausible motion, and further verifies adherence to the specified film style.

\item[2)] \textit{Narrative Expression and Pacing $({NEP})$ :}  ${NEP}$ evaluates whether the film communicates a comprehensible story to viewers unfamiliar with the source novel. It examines the intrinsic storytelling quality of the output, including the clarity of character relations, the coherence of scene transitions, the traceability of major plot developments, and the appropriateness of pacing (i.e., neither rushed, hollow, nor fragmented).

\item[3)] \textit{Audio-Visual Performance Quality $({AVP})$ :} ${AVP}$ evaluates the clarity, stability, and expressive contribution of the audio track, including dialogue intelligibility, volume stability, sound effects, ambient atmosphere, and synchronization with the visuals. As many pipelines still employ lightweight or optional audio modules, ${AVP}$ is scored more leniently than visual quality; nevertheless, recurrent synchronization failures, harsh clipping, unintelligible dialogue, or audio that undermines narrative comprehension are explicitly penalized.
\end{enumerate}

\begin{figure}[htb]
  \centering
  \includegraphics[width=\textwidth]{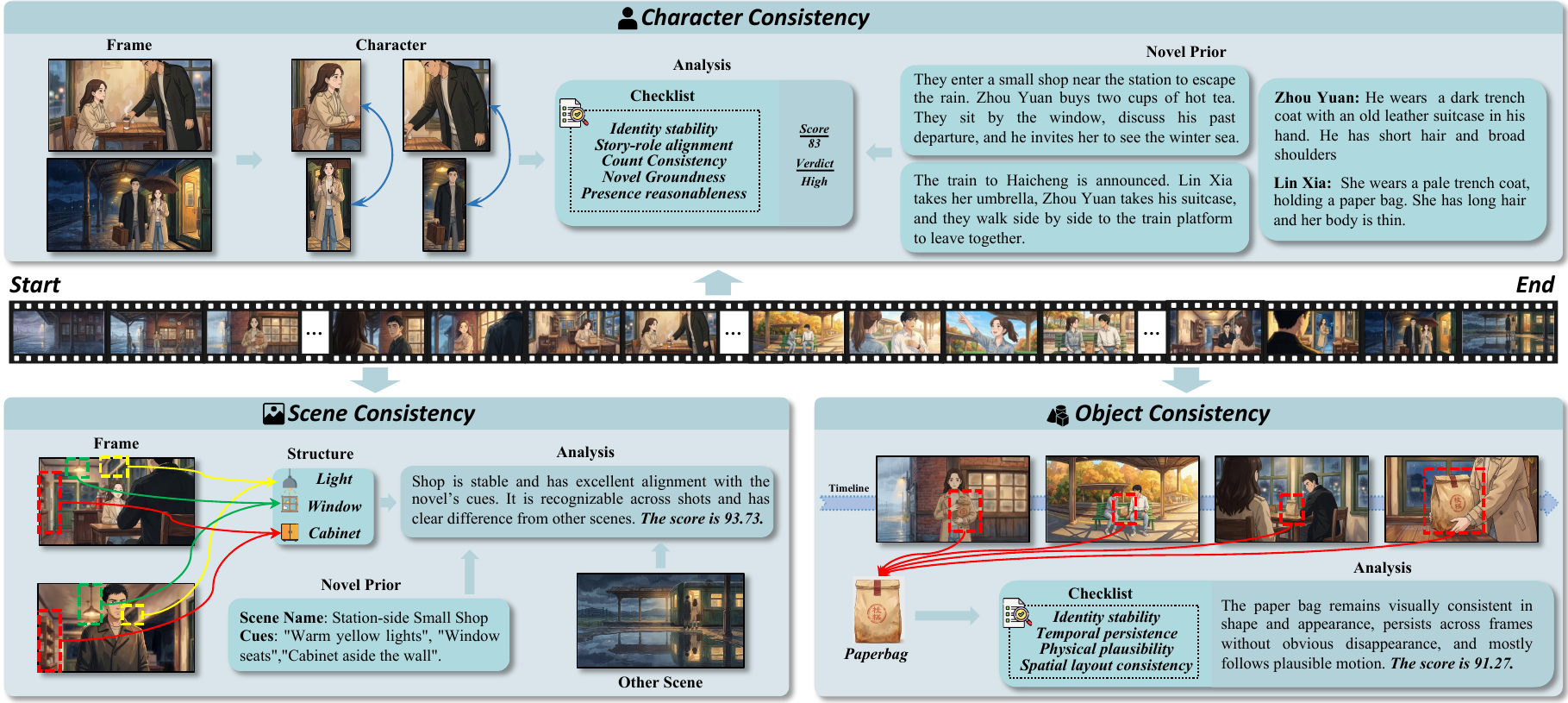}
  \caption{\textbf{Illustration of the film consistency evaluation.} The Character Consistency (CC) module compares on-screen character appearances against a novel-derived character prior to rate identity stability and story alignment. The Scene Consistency (SC) module verifies whether recurring locations preserve stable structural anchors and remain distinguishable from other scenes. The Object Consistency (OC) module tracks salient objects across frames to rate their identity, temporal persistence, and spatial coherence.}
  \label{fig:characterconsistency}
\end{figure}

\noindent\textbf{Film Consistency.} This dimension evaluates whether the generated film maintains stable visual entities and environments over time, examining the persistence of character identities, the structural stability of scenes, and the temporal and spatial coherence of key objects across shots.

\begin{enumerate}[leftmargin=*,itemsep=1pt]
\item[1)] \textit{Character Consistency $({CC})$:}  ${CC}$ evaluates whether characters remain identifiable and story-faithful throughout the film. FilmEval extracts a structured character prior from the source novel, comprising roles, visual cues, chronological events, expected on-screen appearances, prohibited co-occurrences, and count constraints. Each film segment is then rated along five sub-criteria: identity stability, story-role alignment, presence reasonableness, count and composition consistency and novel groundedness. Flashbacks, memories, time jumps, and age progressions are not treated as identity drift unless a character visibly changes into a different individual within the same continuous timeline.

\item[2)] \textit{Scene Consistency $({SC})$:} ${SC}$ evaluates whether the same scene remains recognizable and structurally stable across neighboring shots. FilmEval extracts a compact scene prior from the novel, including recurring or narratively salient settings, scene types, visual cues, and story positions. Each film segment is then rated along five sub-criteria: scene stability, structural layout continuity, cross-shot scene recognizability, inter-scene separability, and novel scene alignment. The metric anchors on stable background elements, tolerating routine camera motion, zooming, partial occlusion, and minor set-dressing variation, while penalizing substantive changes that make a supposedly continuous location appear as a different place.

\item[3)] \textit{Object Consistency $({OC})$:}  ${OC}$ evaluates the stability of narratively relevant objects, including props, furniture, vehicles, and other salient items. Each segment is rated along four sub-criteria: object identity stability, temporal persistence, physical plausibility, and spatial layout consistency. Concrete object-level failures, such as shape deformation, flickering, duplication, physical violation, and layout jumps, are further mapped to severity-graded penalties.

\end{enumerate}

\begin{figure}[h]
  \centering
   \includegraphics[width=\textwidth]{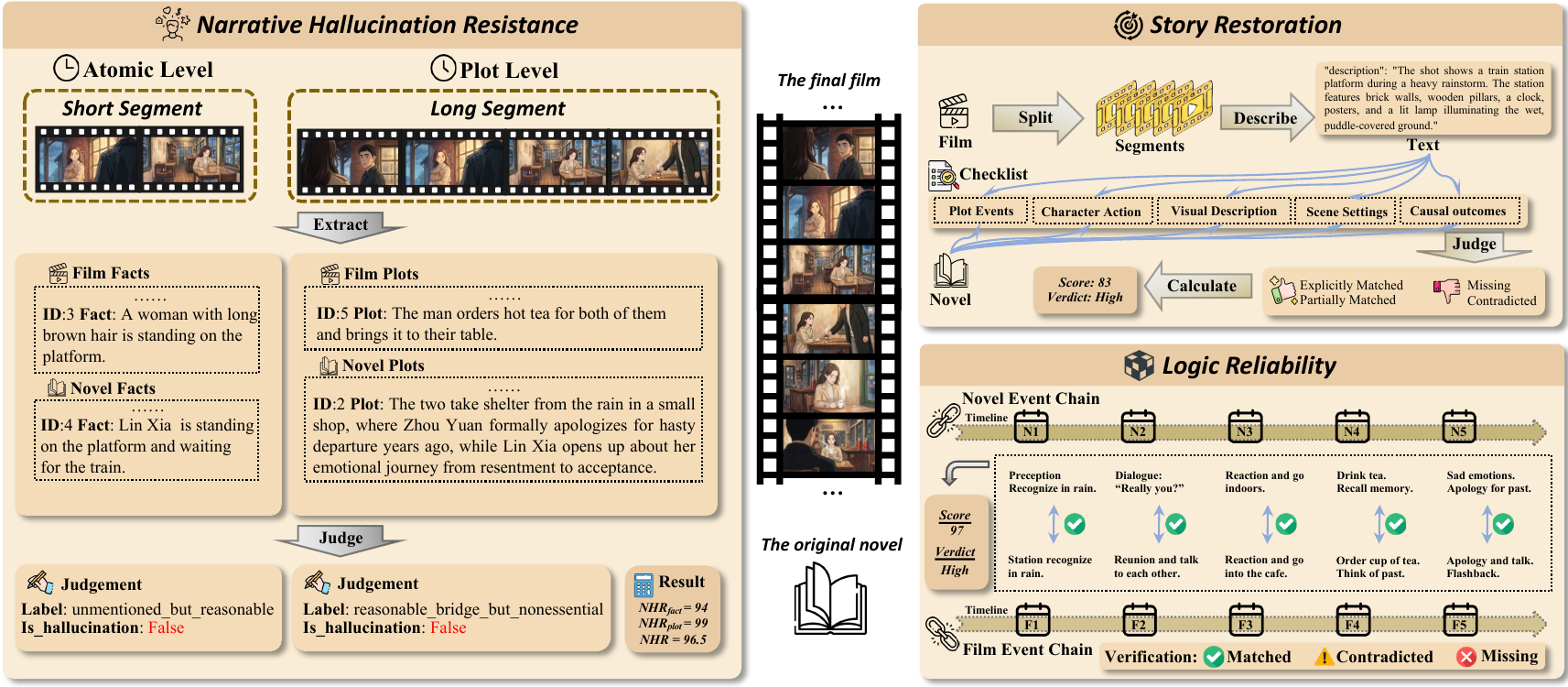}
  \caption{\textbf{Illustration of the novel fidelity evaluation.} The Narrative Hallucination Resistance (NHR) module compares atomic facts and plot events from the film against the novel, flagging unsupported or incongruent content. The Logic Reliability (LR) module aligns the novel and film event chains to verify temporal order, causality, and character-action attribution, distinguishing matched, contradicted, and missing events. The Story Restoration (SR) module examines fine-grained details from a novel-derived checklist to quantify source coverage.}
  \label{fig:storyrestoration}
\end{figure}

\noindent\textbf{Novel Fidelity.} This dimension evaluates the faithfulness of the generated film to its source novel, examining three complementary aspects: whether the film fabricates unsupported content, whether it preserves narrative logic, and whether it recovers fine-grained source details.

\begin{enumerate}[leftmargin=*,itemsep=1pt]
\item[1)] \textit{Narrative Hallucination Resistance $({NHR})$:}  ${NHR}$ measures the extent to which the generated film avoids unsupported or narratively incongruent content. FilmEval assesses hallucination at two granularities: atomic facts and plot-level events, each compared against its counterpart in the source novel. Every item is assigned one of four labels: source-supported, plausible extrapolation, incongruent fabrication, or source-conflicting rendering. The latter two are counted as hallucinations.

\item[2)] \textit{Logic Reliability $({LR})$:} ${LR}$ evaluates whether the generated film preserves the causal, temporal, and relational logic of the source narrative, focusing on contradictions rather than omissions. Typical failures include reversing cause and effect, placing events in an impossible order, attributing actions to the wrong character, and producing incompatible scene transitions.

\item[3)] \textit{Story Restoration $({SR})$:} ${SR}$ quantifies how much of the source novel is preserved in the generated film. FilmEval first distills the novel into a fine-grained adaptation checklist of atomic details spanning plot events, character actions, character relations, dialogue, and setting. The generated film is then converted into a dense chronological description, against which each checklist item is aligned and labeled as explicitly matched, partially matched, 	missing, or contradicted. A second video-grounded review is applied to initially unresolved items to reduce false negatives introduced by text-only reconstruction.

\end{enumerate}

\newpage
\section{Experiments}
\label{sec:exp}

\subsection{Experimental Settings}

\textbf{Dataset and Metrics.} We evaluate on FilmEval (Section~\ref{sec:filmeval}), reporting the three metric groups defined therein: \textit{Cinematic Presentation} ($\mathcal{CP}$), \textit{Film Consistency} ($\mathcal{FC}$), and \textit{Novel Fidelity} ($\mathcal{NF}$). All scores lie in $[0, 100]$ and are aggregated uniformly across the Easy, Medium, and Hard tiers.

\textbf{Baselines.} We compare FilmWorld against five representative agentic video generation systems: MM-StoryAgent~\cite{xu2025mmstoryagent}, VGoT~\cite{zheng2025vgot}, MovieAgent~\cite{wu2025movieagent}, ViMax~\cite{huang2026vimax}, and VideoClaw~\cite{videoclaw2026}. Long-video generation methods are excluded from this comparison, as they consume per-shot prompts rather than raw prose and are architecturally ill-suited to the novel-to-film generation. To isolate framework-level differences, all methods are deployed under a unified foundation stack: \textit{Gemini 3.1 Pro} as the multi-modal LLM, \textit{Nano Banana 2} as the image generator~\cite{team2023gemini}, and \textit{Wan 2.7} as the video generator~\cite{wan2025wan}. For baselines not natively designed to ingest raw literary prose, we apply minimal front-end adaptations to make them compatible with the novel-to-film task while strictly preserving their original orchestration logic.

\textbf{Implementation Details.} FilmWorld is implemented with fully asynchronous orchestration. Keyframes are rendered at $2$K resolution with a $16{:}9$ aspect ratio, while video segments are produced at $1080$P, with each shot duration bounded to $[1, 15]$ seconds. The dynamic state propagation uses a sliding window of $g{=}2$ target shots padded by $c{=}1$ context shot on each side. The closed-loop verifier performs up to $3$ correction rounds for keyframes and video segments, and $1$ round for reference assets. Finally, to ensure a fair comparison and without loss of generality, \textbf{\textit{we uniformly apply a Ghibli anime style across all methods}}, which is implemented straightforwardly through direct prompt injection, thereby eliminating potential evaluation bias introduced by stylistic discrepancies. More experimental details are provided in the Appendix.

\subsection{Quantitative Analysis}

\textbf{Overall Performance Comparison on FilmEval.}
Table~\ref{tab:main_comparison} compares FilmWorld against five state-of-the-art agentic video generation systems on the FilmEval benchmark.
FilmWorld achieves the highest overall score of $89.38$, outperforming the strongest baseline VideoClaw~\cite{videoclaw2026} by $5.19$ points and leading across all evaluation cells.
Notably, in \textit{Novel Fidelity}, FilmWorld surpasses the runner-up MovieAgent~\cite{wu2025movieagent} by $+6.78$, $+6.29$, and $+9.46$ points on the Easy, Medium, and Hard tiers, respectively.
This widening gap confirms that grounding generation in a symbolic state trajectory effectively prevents narrative omissions and hallucinations.
FilmWorld also yields superior \textit{Film Consistency} scores of $90.93$/$91.56$/$92.95$, verifying that state externalization and visual referencing successfully transform coherence into an enforced constraint.
Crucially, while baselines like VGoT~\cite{zheng2025vgot} and ViMax~\cite{huang2026vimax} degrade by up to $3.23$ points as complexity scales due to cumulative drift, FilmWorld's average performance remains remarkably stable within a tight $0.99$-point band ($89.10 \rightarrow 88.98 \rightarrow 89.97$).
This invariant profile empirically validates our decoupling paradigm: pre-materializing the state trajectory fully insulates downstream rendering quality from narrative depth.

\begin{table*}[!t]
\centering
\caption{\textbf{Overall performance comparison on FilmEval.} We report three metric groups, \textit{Cinematic Presentation} ($\mathcal{CP}$), \textit{Film Consistency} ($\mathcal{FC}$), and \textit{Novel Fidelity} ($\mathcal{NF}$), under the Easy, Medium, and Hard difficulty tiers. The rightmost column reports the overall mean across all nine cells. \textbf{Bold} indicates the best result and \underline{underline} the second best.}
\label{tab:main_comparison}
\renewcommand{\arraystretch}{1.18}
\setlength{\tabcolsep}{5.8pt}
\small
\begin{tabular}{l c ccc ccc ccc c}
\toprule
\multirow{2}{*}{\textbf{Method}} & \multirow{2}{*}{\textbf{Year}} & \multicolumn{3}{c}{\textbf{Easy}} & \multicolumn{3}{c}{\textbf{Medium}} & \multicolumn{3}{c}{\textbf{Hard}} & \multirow{2}{*}{\textbf{Overall}} \\
\cmidrule(lr){3-5} \cmidrule(lr){6-8} \cmidrule(lr){9-11}
 & & $\mathcal{CP}$ & $\mathcal{FC}$ & $\mathcal{NF}$ & $\mathcal{CP}$ & $\mathcal{FC}$ & $\mathcal{NF}$ & $\mathcal{CP}$ & $\mathcal{FC}$ & $\mathcal{NF}$ &  \\
\midrule
MM-StoryAgent~\cite{xu2025mmstoryagent} & 2025 & 81.53 & 84.40 & 78.49 & 77.47 & 81.22 & 74.44 & 80.47 & 87.70 & 78.88 & 80.51 \\
VGoT~\cite{zheng2025vgot}               & 2025 & \underline{83.27} & 86.93 & 76.82 & 80.33 & 87.72 & 82.37 & 80.00 & 87.99 & 72.74 & 82.02 \\
MovieAgent~\cite{wu2025movieagent}      & 2025 & 77.27 & 84.28 & \underline{85.88} & 78.93 & 87.07 & \underline{86.62} & \underline{81.73} & \underline{89.43} & \underline{84.63} & 83.98 \\
ViMax~\cite{huang2026vimax}             & 2026 & 83.13 & \underline{87.66} & 78.48 & 76.27 & 85.94 & 78.61 & 78.73 & 87.60 & 73.55 & 81.11 \\
VideoClaw~\cite{videoclaw2026}          & 2026 & 82.27 & 86.15 & 84.22 & \underline{81.67} & \underline{88.34} & 82.15 & 80.53 & 88.12 & 84.28 & \underline{84.19} \\
\midrule
\rowcolor{blue!8}
\textbf{FilmWorld (Ours)}               & 2026 & \textbf{84.00} & \textbf{90.93} & \textbf{92.66} & \textbf{82.47} & \textbf{91.56} & \textbf{92.91} & \textbf{82.87} & \textbf{92.95} & \textbf{94.09} & \textbf{89.38} \\
\bottomrule
\end{tabular}
\end{table*}

\begin{figure}[htb]  
  \centering  
  \includegraphics[width=0.58\textwidth]{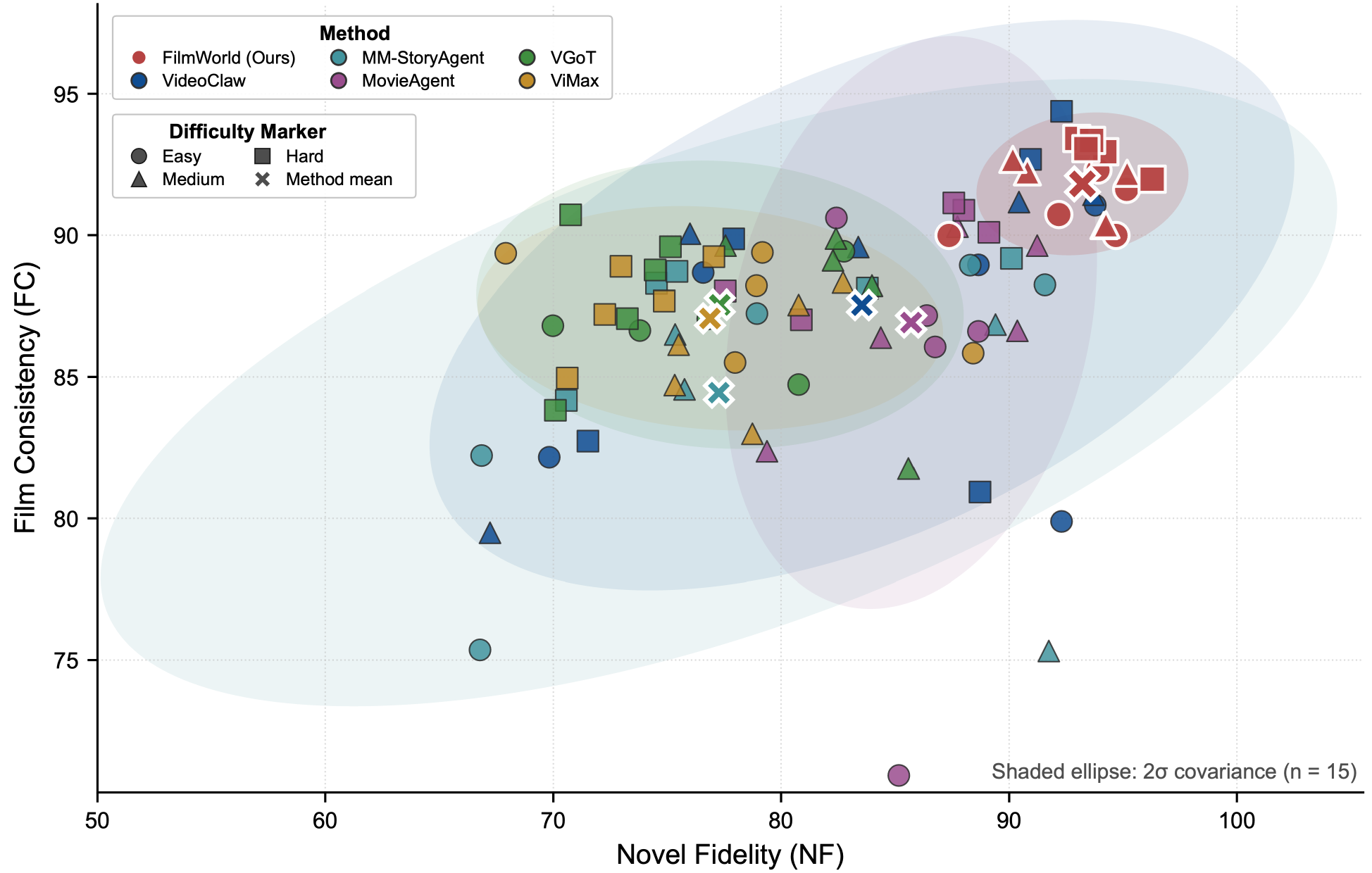}  
  \caption{\textbf{Per-novel performance distribution on FilmEval.} Each point represents a single novel evaluated under the corresponding difficulty tier, plotted in the Novel Fidelity--Film Consistency space. Shaded regions denote $2\sigma$ covariance ellipses estimated over all 15 novels. FilmWorld occupies the top-right region with the most compact ellipse, indicating both consistently superior performance and significantly lower inter-novel variance compared to all baselines.}
  \label{fig:per_novel_distribution}  
\end{figure}

\textbf{Fine-grained Performance Comparison on FilmEval.} We provide a fine-grained comparison from two complementary perspectives. Table~\ref{tab:sub_metric_comparison} dissects performance at the sub-metric level across all three difficulty tiers. FilmWorld achieves the highest overall average scores across all difficulty tiers, maintaining top-tier performance in the vast majority of the nine sub-metrics. Its advantage is most pronounced within the Novel Fidelity group (e.g., NHR, LR, and SR), where the gap over baselines widens substantially under the Hard tier, validating the effectiveness of our symbolic state trajectory in preventing narrative omissions and hallucinations. Conversely, the Cinematic Presentation group exhibits relatively uniform scores across all methods, reflecting the dominant influence of the shared foundation model backbone. To further assess robustness against narrative complexity, Figure~\ref{fig:per_novel_distribution} plots each method's per-novel results in the Novel Fidelity–Film Consistency space. FilmWorld's covariance ellipse occupies the top-right region with the highest mean coordinates, confirming that its overall advantage is consistent across all 15 novels rather than driven by favorable outliers. Crucially, FilmWorld's ellipse is substantially more compact than those of all baselines, indicating significantly lower inter-novel variance. In contrast, baseline methods exhibit large, dispersed ellipses alongside a clear downward-leftward displacement from Easy to Hard points, revealing their sensitivity to increasing narrative difficulty.

\begin{table*}[!t]
\centering
\caption{\textbf{Fine-grained sub-metric comparison on FilmEval across all three difficulty tiers.} We report all nine sub-metrics of the three metric groups introduced in Table~\ref{tab:main_comparison}: \textit{Cinematic Presentation} ($\mathcal{CP}$) contains VP, NEP, AVP; \textit{Film Consistency} ($\mathcal{FC}$) contains CC, SC, OC; and \textit{Novel Fidelity} ($\mathcal{NF}$) contains NHR, LR, SR. Each method reports its three difficulty rows separately; the rightmost Avg column is the mean over the nine sub-metrics of that row. \textbf{Bold} and \underline{underline} mark the best and the second best result \emph{within the same difficulty tier}.}
\label{tab:sub_metric_comparison}
\renewcommand{\arraystretch}{1.18}
\setlength{\tabcolsep}{5.4pt}
\footnotesize
\begin{tabular}{l c c ccc ccc ccc c}
\toprule
\multirow{2}{*}{\textbf{Method}} & \multirow{2}{*}{\textbf{Year}} & \multirow{2}{*}{\textbf{Difficulty}}
 & \multicolumn{3}{c}{\textbf{Cinematic Presentation}}
 & \multicolumn{3}{c}{\textbf{Film Consistency}}
 & \multicolumn{3}{c}{\textbf{Novel Fidelity}}
 & \multirow{2}{*}{\textbf{Avg}} \\
\cmidrule(lr){4-6} \cmidrule(lr){7-9} \cmidrule(lr){10-12}
 & & & VP & NEP & AVP & CC & SC & OC & NHR & LR & SR & \\
\midrule
\multirow{3}{*}{MM-StoryAgent~\cite{xu2025mmstoryagent}} & \multirow{3}{*}{2025}
  & Easy   & \underline{82.60} & 81.00 & 81.00 & 71.75 & 85.10 & \underline{96.35} & 83.06 & 84.40 & 68.00 & 81.47 \\
 &  & Medium & 77.20 & 76.20 & 79.00 & 63.27 & 86.46 & 93.94 & 80.71 & 83.40 & 59.20 & 77.71 \\
 &  & Hard   & 80.60 & 80.20 & 80.60 & 75.93 & 89.50 & 97.68 & \underline{94.45} & 81.80 & 60.40 & 82.35 \\
\cmidrule(l){1-13}
\multirow{3}{*}{VGoT~\cite{zheng2025vgot}} & \multirow{3}{*}{2025}
  & Easy   & \textbf{84.60} & \underline{84.00} & 81.20 & 80.30 & 86.84 & 93.67 & 80.25 & 84.80 & 65.40 & 82.34 \\
 &  & Medium & 79.80 & \underline{82.80} & 78.40 & \underline{79.09} & 88.61 & 95.45 & 91.30 & 86.80 & 69.00 & 83.47 \\
 &  & Hard   & 78.80 & 80.60 & 80.60 & 78.77 & 88.19 & 97.01 & 89.23 & 78.80 & 50.20 & 80.24 \\
\cmidrule(l){1-13}
\multirow{3}{*}{MovieAgent~\cite{wu2025movieagent}} & \multirow{3}{*}{2025}
  & Easy   & 76.00 & 78.40 & 77.40 & 72.34 & 85.56 & 94.93 & 85.83 & \underline{93.80} & \underline{78.00} & 82.47 \\
 &  & Medium & 80.00 & 80.80 & 76.00 & 76.50 & 87.54 & \underline{97.16} & 90.26 & \underline{91.60} & \underline{78.00} & \underline{84.21} \\
 &  & Hard   & \textbf{81.40} & \textbf{84.20} & 79.60 & 79.22 & 90.52 & \underline{98.56} & 90.08 & \underline{94.60} & \underline{69.20} & \underline{85.26} \\
\cmidrule(l){1-13}
\multirow{3}{*}{ViMax~\cite{huang2026vimax}} & \multirow{3}{*}{2026}
  & Easy   & 81.80 & 82.40 & \textbf{85.20} & \underline{82.02} & 87.02 & 93.95 & 87.65 & 80.80 & 67.00 & 83.09 \\
 &  & Medium & 73.80 & 78.40 & 76.60 & 78.25 & 86.08 & 93.49 & \textbf{94.83} & 77.40 & 63.60 & 80.27 \\
 &  & Hard   & 74.80 & 80.20 & 81.20 & \underline{79.91} & 88.81 & 94.08 & 90.26 & 76.60 & 53.80 & 79.96 \\
\cmidrule(l){1-13}
\multirow{3}{*}{VideoClaw~\cite{videoclaw2026}} & \multirow{3}{*}{2026}
  & Easy   & 81.00 & 83.20 & 82.60 & 68.28 & \textbf{94.78} & 95.38 & \underline{88.07} & 89.80 & 74.80 & \underline{84.21} \\
 &  & Medium & \textbf{81.40} & 81.40 & \underline{82.20} & 77.33 & \underline{92.43} & 95.27 & 90.64 & 85.00 & 70.80 & 84.05 \\
 &  & Hard   & 76.80 & 81.40 & \underline{83.40} & 74.17 & \underline{94.52} & 95.67 & 93.84 & 90.00 & 69.00 & 84.31 \\
\midrule
\rowcolor{blue!8}
 &  & Easy   & 81.80 & \textbf{85.80} & \underline{84.40} & \textbf{82.90} & \underline{92.64} & \textbf{97.22} & \textbf{96.97} & \textbf{97.00} & \textbf{84.00} & \textbf{89.19} \\
\rowcolor{blue!8}
 &  & Medium & \underline{80.20} & \textbf{84.00} & \textbf{83.20} & \textbf{82.15} & \textbf{95.25} & \textbf{97.29} & \underline{93.53} & \textbf{99.20} & \textbf{86.00} & \textbf{88.98} \\
\rowcolor{blue!8}
\multirow{-3}{*}{\textbf{FilmWorld (Ours)}} & \multirow{-3}{*}{2026}
  & Hard   & \underline{81.00} & \underline{84.00} & \textbf{83.60} & \textbf{84.11} & \textbf{96.04} & \textbf{98.70} & \textbf{97.27} & \textbf{98.80} & \textbf{86.20} & \textbf{89.97} \\
\bottomrule
\end{tabular}
\vspace{-3mm}
\end{table*}

\begin{figure}[H]  
  \centering  
  \vspace{3mm}
  \includegraphics[width=\textwidth]{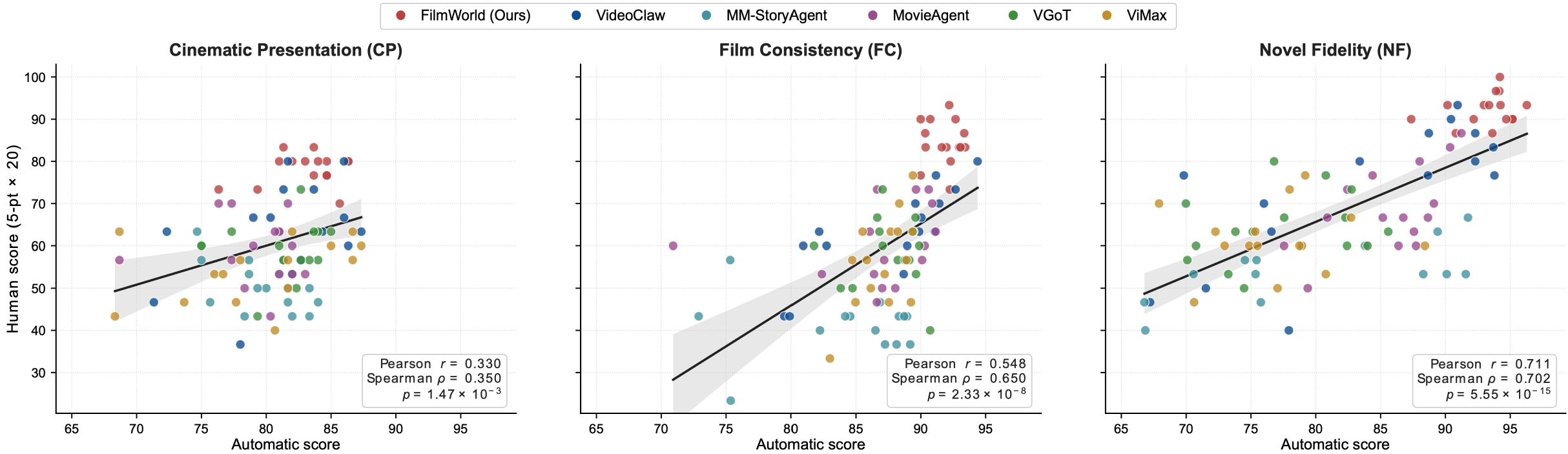}  
  \caption{\textbf{Alignment between FilmEval automatic scores and human ratings.} Each panel plots the per-film automatic score against the mean human rating ($0$--$5$ scaled to $[0,100]$) for one metric group, with $90$ paired points (15 novels $\times$ 6 systems) colored by system. The solid line is a linear fit and the shaded band its $95\%$ confidence interval; Pearson $r$, Spearman $\rho$, and the two-sided $p$-value are annotated in each panel. All three groups are significantly positively correlated, and the agreement strengthens from CP ($r=0.330$) to FC ($r=0.548$) to NF ($r=0.711$).}
  \label{fig:human_eval}  
\end{figure}

\textbf{Human Evaluation and Automatic-Human Alignment.} To validate that our automatic metrics faithfully reflect human perception, we recruit six annotators to rate all 90 generated films (15 novels × 6 systems) on a 0–5 scale across the same three metric groups, yielding 540 judgments (Table~\ref{tab:human_eval}). FilmWorld attains the highest overall rating of $4.25$, surpassing the strongest baseline VideoClaw~\cite{videoclaw2026} ($3.33$) by a decisive $+0.92$ margin and ranking first in all nine difficulty–dimension cells; the gap is most pronounced on Novel Fidelity, where it exceeds $4.5$ on every tier while all baselines stay below $3.8$. Critically, the human ranking of all six systems is identical to that induced by FilmEval (Spearman $\rho=1.0$), demonstrating that our benchmark preserves the correct system-level ordering. At the finer per-film level, as shown in Figure~\ref{fig:human_eval}, all three groups show statistically significant positive correlations (Pearson $r=0.330$/$0.548$/$0.711$ for CP/FC/NF, all $p<10^{-2}$), and the alignment strengthens monotonically from CP to FC to NF: our checklist-grounded, novel-anchored consistency and fidelity metrics reach strong agreement ($r>0.54$, $p<10^{-7}$) by capturing objectively verifiable properties, whereas the weaker CP correlation reflects the subjective nature of aesthetic presentation.

\begin{table*}[!t]
\centering
\vspace{3mm}
\caption{\textbf{Human evaluation results on FilmEval.} We recruit a pool of human annotators to rate the generated films on a 0--5 scale (higher is better) across the same three metric groups as Table~\ref{tab:main_comparison}, under the Easy, Medium, and Hard difficulty tiers. Each cell reports the mean $\pm$ standard deviation of the collected ratings within the tier. The rightmost column reports the overall mean across all nine cells. \textbf{Bold} indicates the best result and \underline{underline} the second best.}
\label{tab:human_eval}
\renewcommand{\arraystretch}{1.18}
\newcommand{\hs}[2]{#1{\scriptsize$\pm$#2}}
\setlength{\tabcolsep}{2.0pt}
\footnotesize
\begin{tabular}{l ccc ccc ccc c}
\toprule
\multirow{2}{*}{\textbf{Method}} & \multicolumn{3}{c}{\textbf{Easy}} & \multicolumn{3}{c}{\textbf{Medium}} & \multicolumn{3}{c}{\textbf{Hard}} & \multirow{2}{*}{\textbf{Overall}} \\
\cmidrule(lr){2-4} \cmidrule(lr){5-7} \cmidrule(lr){8-10}
 & $\mathcal{CP}$ & $\mathcal{FC}$ & $\mathcal{NF}$ & $\mathcal{CP}$ & $\mathcal{FC}$ & $\mathcal{NF}$ & $\mathcal{CP}$ & $\mathcal{FC}$ & $\mathcal{NF}$ &  \\
\midrule
MM-StoryAgent~\cite{xu2025mmstoryagent} & \hs{2.47}{0.78} & \hs{1.73}{0.87} & \hs{2.53}{0.73} & \hs{2.60}{0.72} & \hs{2.30}{1.12} & \hs{2.83}{0.87} & \hs{2.47}{0.78} & \hs{2.03}{1.03} & \hs{2.80}{0.61} & 2.42 \\
VGoT~\cite{zheng2025vgot} & \hs{\underline{3.00}}{0.83} & \hs{3.17}{0.70} & \hs{3.63}{0.72} & \hs{2.83}{0.75} & \hs{2.87}{0.82} & \hs{3.17}{0.70} & \hs{3.00}{0.64} & \hs{2.73}{0.78} & \hs{2.83}{0.75} & 3.03 \\
MovieAgent~\cite{wu2025movieagent} & \hs{2.80}{0.89} & \hs{3.00}{1.02} & \hs{3.33}{0.66} & \hs{3.13}{0.82} & \hs{3.13}{1.04} & \hs{3.57}{0.97} & \hs{2.93}{0.98} & \hs{2.90}{0.88} & \hs{3.40}{0.81} & 3.13 \\
ViMax~\cite{huang2026vimax} & \hs{2.93}{0.74} & \hs{\underline{3.23}}{0.90} & \hs{3.40}{0.97} & \hs{2.63}{0.93} & \hs{2.57}{0.97} & \hs{3.03}{0.89} & \hs{2.57}{0.77} & \hs{2.67}{0.80} & \hs{2.80}{0.92} & 2.87 \\
VideoClaw~\cite{videoclaw2026} & \hs{2.87}{0.78} & \hs{2.83}{0.75} & \hs{\underline{3.73}}{0.64} & \hs{\underline{3.40}}{0.89} & \hs{\underline{3.27}}{0.87} & \hs{\underline{3.70}}{0.95} & \hs{\underline{3.23}}{0.73} & \hs{\underline{3.37}}{0.89} & \hs{\underline{3.57}}{1.38} & \underline{3.33} \\
\midrule
\rowcolor{blue!8}
\textbf{FilmWorld (Ours)} & \hs{\textbf{4.00}}{0.45} & \hs{\textbf{4.20}}{0.61} & \hs{\textbf{4.57}}{0.50} & \hs{\textbf{3.83}}{0.59} & \hs{\textbf{4.27}}{0.83} & \hs{\textbf{4.60}}{0.72} & \hs{\textbf{3.90}}{0.80} & \hs{\textbf{4.20}}{0.66} & \hs{\textbf{4.67}}{0.76} & \textbf{4.25} \\
\bottomrule
\end{tabular}
\end{table*}

\subsection{Ablation and Diagnostic Studies}

\textbf{Component-Wise Ablation.} To keep the ablation cost tractable given the expense of full novel-to-film generation, we conduct the study on three representative novels, drawn from each difficulty tier. Table~\ref{tab:ablation} reports the results of ablating three core designs of FilmWorld on this subset. Removing the explicit world state causes the sharpest degradation (90.15 → 84.28), with Character Consistency collapsing from 82.12 to 54.50. Without an addressable state identifier, downstream generators re-infer entity identity per shot and drift heavily, confirming that explicit state modeling is the structural foundation of long-range consistency rather than an optional enhancement. For cross-shot dynamic propagation, removing the mechanism as a whole lowers the average to 87.96. Further ablating its sub-components shows that neither channel alone suffices: dropping the textual end-state yields 87.84 with a 5.65-point CC drop, while dropping the visual keyframe yields 88.16 with AVP and SR most affected. The two channels constrain complementary axes and must be applied jointly. Disabling the closed-loop verifier lowers the average to 88.29. Unlike the previous two, its impact is diffuse across metrics, consistent with its role as a residual defect corrector rather than a first-order constraint provider. Overall, the three designs address distinct bottlenecks, and removing any of them measurably degrades the film quality.

\begin{table*}[!t]
\centering
\caption{\textbf{Ablation study on three representative novels, one from each difficulty tier.} We ablate three core designs of FilmWorld, grouped by their architectural role: (i) explicit world state modeling in the Construction Phase, (ii) cross-shot dynamic state propagation in the Evolution Phase, with three sub-mechanisms further ablated individually, and (iii) closed-loop verification. \textbf{Bold} and \underline{underline} mark the best and the second best result within the same column.}
\label{tab:ablation}

\renewcommand{\arraystretch}{1.18}
\setlength{\tabcolsep}{5.6pt}
\small
\begin{tabular}{l ccc ccc ccc c}
\toprule
\multirow{2}{*}{\textbf{Method}}
 & \multicolumn{3}{c}{\textbf{Cinematic Presentation}}
 & \multicolumn{3}{c}{\textbf{Film Consistency}}
 & \multicolumn{3}{c}{\textbf{Novel Fidelity}}
 & \multirow{2}{*}{\textbf{Avg}} \\
\cmidrule(lr){2-4} \cmidrule(lr){5-7} \cmidrule(lr){8-10}
 & VP & NEP & AVP & CC & SC & OC & NHR & LR & SR & \\
\midrule
\multicolumn{11}{l}{\textcolor{gray!80}{\textit{World State Modeling}}} \\
w/o Explicit World State
  & \underline{81.72} & 84.10 & 81.72 & 54.50 & 90.09 & 95.51 & 90.51 & 96.67 & 83.71 & 84.28 \\
\cmidrule(l){1-11}
\multicolumn{11}{l}{\textcolor{gray!80}{\textit{Cross-Shot Dynamic State Propagation}}} \\
w/o Cross-Shot Propagation
  & 80.42 & 83.80 & 82.76 & 78.27 & 91.23 & 97.53 & 94.96 & 97.65 & 84.99 & 87.96 \\
w/o Sliding Window
  & 81.38 & 84.11 & \underline{83.75} & 79.33 & \underline{92.65} & 97.27 & 92.11 & 98.67 & \underline{85.40} & \underline{88.30} \\
w/o Textual End-state
  & 81.07 & 82.47 & 82.75 & 76.47 & 92.59 & \underline{97.91} & 93.65 & 98.33 & 85.35 & 87.84 \\
w/o Visual Keyframe
  & 81.04 & 82.78 & 82.47 & \underline{79.86} & 92.43 & 96.75 & 94.75 & 98.67 & 84.71 & 88.16 \\
\cmidrule(l){1-11}
\multicolumn{11}{l}{\textcolor{gray!80}{\textit{Closed-Loop Verification}}} \\
w/o Closed-Loop Verifier
  & 81.08 & \underline{84.45} & 82.13 & 78.67 & 91.65 & 97.57 & \underline{95.35} & \underline{99.33} & 84.37 & 88.29 \\
\midrule

\rowcolor{blue!8}
\textbf{FilmWorld (ours)}
  & \textbf{82.00} & \textbf{85.33} & \textbf{85.00} & \textbf{82.12} & \textbf{94.24} & \textbf{99.17} & \textbf{97.15} & \textbf{99.67} & \textbf{86.67} & \textbf{90.15} \\
\bottomrule
\end{tabular}
\end{table*}

\begin{figure}[htb]  
  \centering  
  \vspace{-8mm}
 \includegraphics[width=0.86\textwidth]{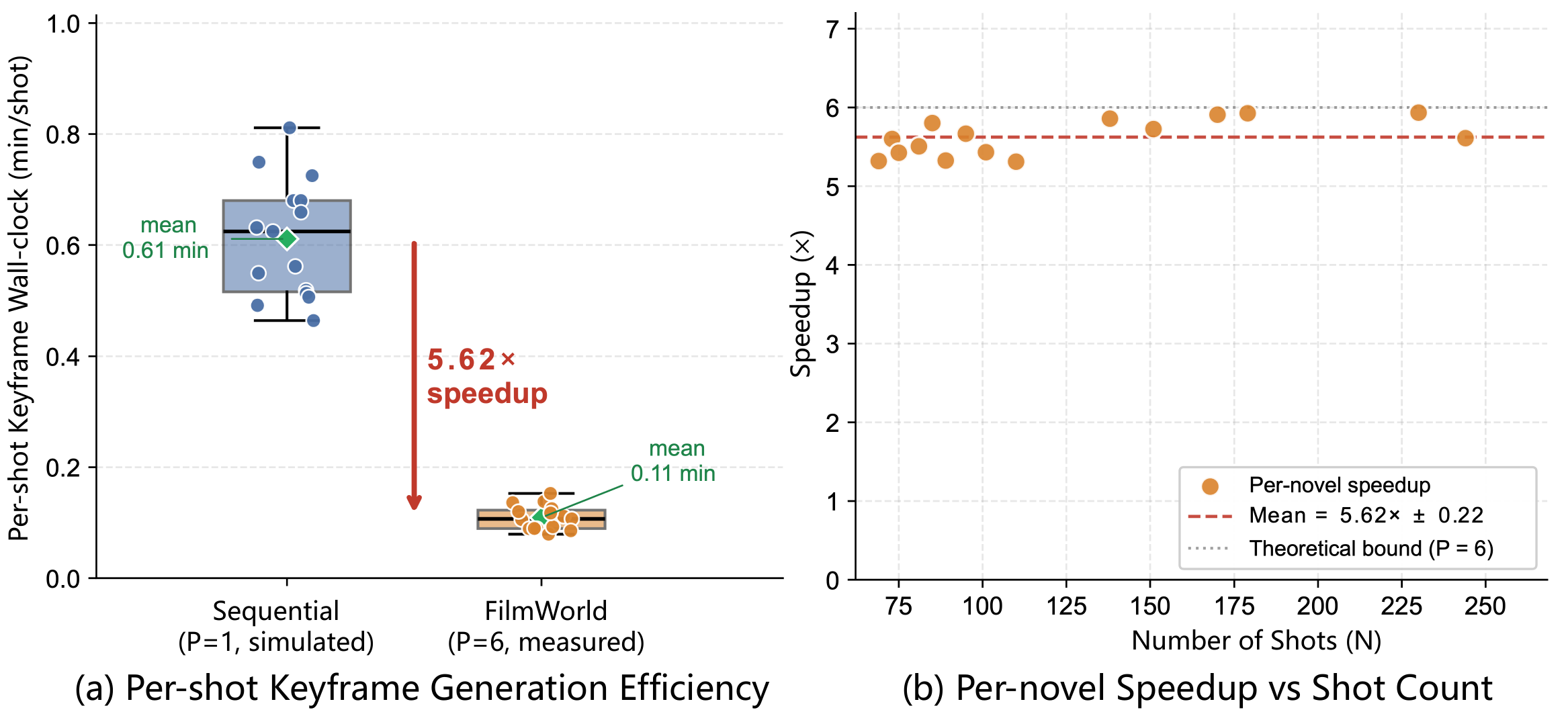}
  \caption{\textbf{Computational efficiency and scalability of FilmWorld on keyframe generation.}
(a) Efficiency. Parallel execution ($P{=}6$) achieves a $5.62\times$ mean speedup (93.7\% of theoretical bound), reducing per-shot cost from 0.61 to 0.11 min/shot. Green diamonds mark means; dots are individual novels.
(b) Scalability. Speedup stays within $5.62\times \pm 0.22$ across $N \in [69, 244]$, showing no dependence on $N$ and confirming the $O(N/P)$ scaling. The sequential baseline ($P{=}1$) is simulated by accumulating per-shot API durations from the same runs.}
  \label{fig:efficiency}  
\end{figure}

\textbf{Computational Efficiency and Scalability.} 
Section~\ref{subsec:discussion} predicts $O(N/P)$ latency for the Evolution Phase under $P$ parallel workers, versus $O(N)$ for sequential rendering. We validate this on the keyframe generation stage, the most cleanly parallelizable component with no intra-stage dependency across shots, using 15 FilmEval novels. For each novel, we measure the parallel wall-clock under our production setting $P{=}6$, and simulate the sequential baseline by summing per-shot API durations from the same run, isolating scheduling effects from methodological confounds. As shown in Figure~\ref{fig:efficiency} (a), FilmWorld reduces the mean per-shot cost from $0.61$ to $0.11$ min/shot, yielding a $5.62\times$ speedup, or $93.7\%$ of the theoretical bound $P{=}6$, with the residual gap attributable to API rate limits and scheduling overhead. Figure~\ref{fig:efficiency} (b) shows this speedup remains tightly concentrated within $5.62\times \pm 0.22$ across $N \in [69, 244]$, independent of shot count $N$, confirming the predicted $O(N/P)$ scaling: state decoupling converts sequential rendering into a bounded parallel workload invariant to narrative length. The remaining Evolution stages (dynamic planning, video rendering) share this shot-independent structure and should exhibit analogous gains.

\begin{figure}[H]  
  \centering  
  \includegraphics[width=0.88\textwidth]{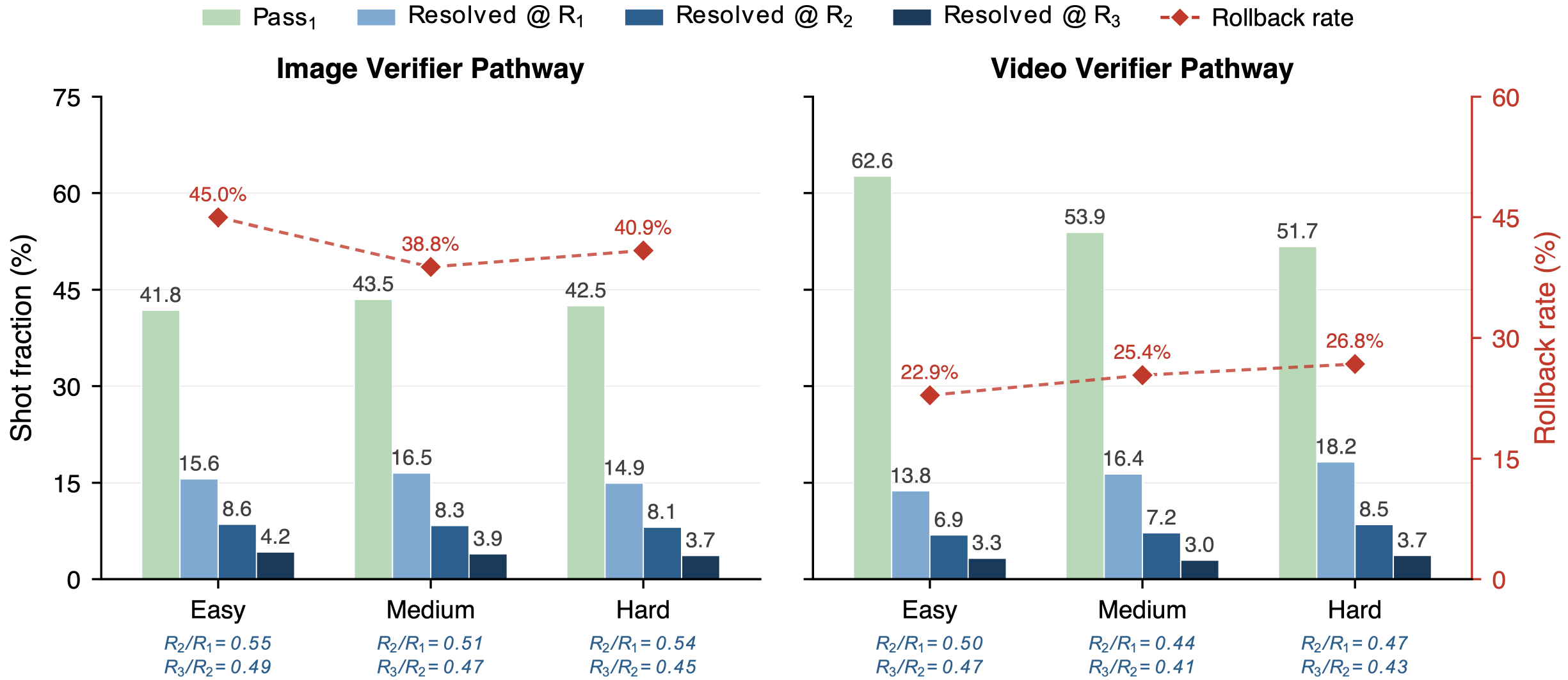}  
  \caption{\textbf{Diminishing returns of closed-loop correction rounds.} For each verifier pathway (image and video) and difficulty tier, we report the shot fraction accepted without correction ($\text{Pass}_1$) and the marginal fraction newly resolved at each subsequent round ($\text{Resolved}@R_1$–$R_3$), alongside the rollback rate (dashed line, right axis). The marginal yield decays sharply across rounds ($R_2/R_1$ and $R_3/R_2$ both below $0.55$), indicating rapidly diminishing benefit from additional correction rounds and supporting our choice of $K=3$ as a cost-effective round budget.}
\label{fig:correction_rounds}  
\end{figure}

\textbf{Diminishing Returns of Correction Rounds.} 
We further examine how much each correction round in the closed-loop verifier actually contributes, in order to justify the cost-effectiveness of our round budget. Figure~\ref{fig:correction_rounds} reports, for each verifier pathway and difficulty tier, the shot fraction resolved at each round: $\text{Pass}_1$ denotes shots accepted without correction, and $\text{Resolved}@R_k$ denotes the marginal fraction newly resolved at round $k$. 

Across both pathways and all difficulty tiers, the marginal yield decays sharply and monotonically with each successive round: $R_2/R_1$ falls in $0.44$–$0.55$ and $R_3/R_2$ further drops to $0.41$–$0.49$, meaning each additional round recovers less than half of what the previous round achieved. This steep diminishing-returns curve indicates that the first correction round already resolves the majority of recoverable defects, while later rounds are increasingly spent chasing a shrinking set of hard failures at disproportionate cost, empirically supporting our budget of $K=3$ rounds as a favorable cost-quality trade-off beyond which further iterations offer little practical benefit. We additionally observe that the video pathway is more sensitive to narrative difficulty than the image pathway, with $\text{Pass}_1$ dropping from $62.6\%$ to $51.7\%$ from Easy to Hard, and that the image pathway exhibits a higher rollback rate ($38.8$–$45.0\%$) than the video pathway ($22.9$–$26.8\%$), suggesting structural keyframe defects are comparatively harder to reliably correct than dynamic descriptions.

\begin{figure}[htb]  
  \centering  
  \includegraphics[width=\textwidth]{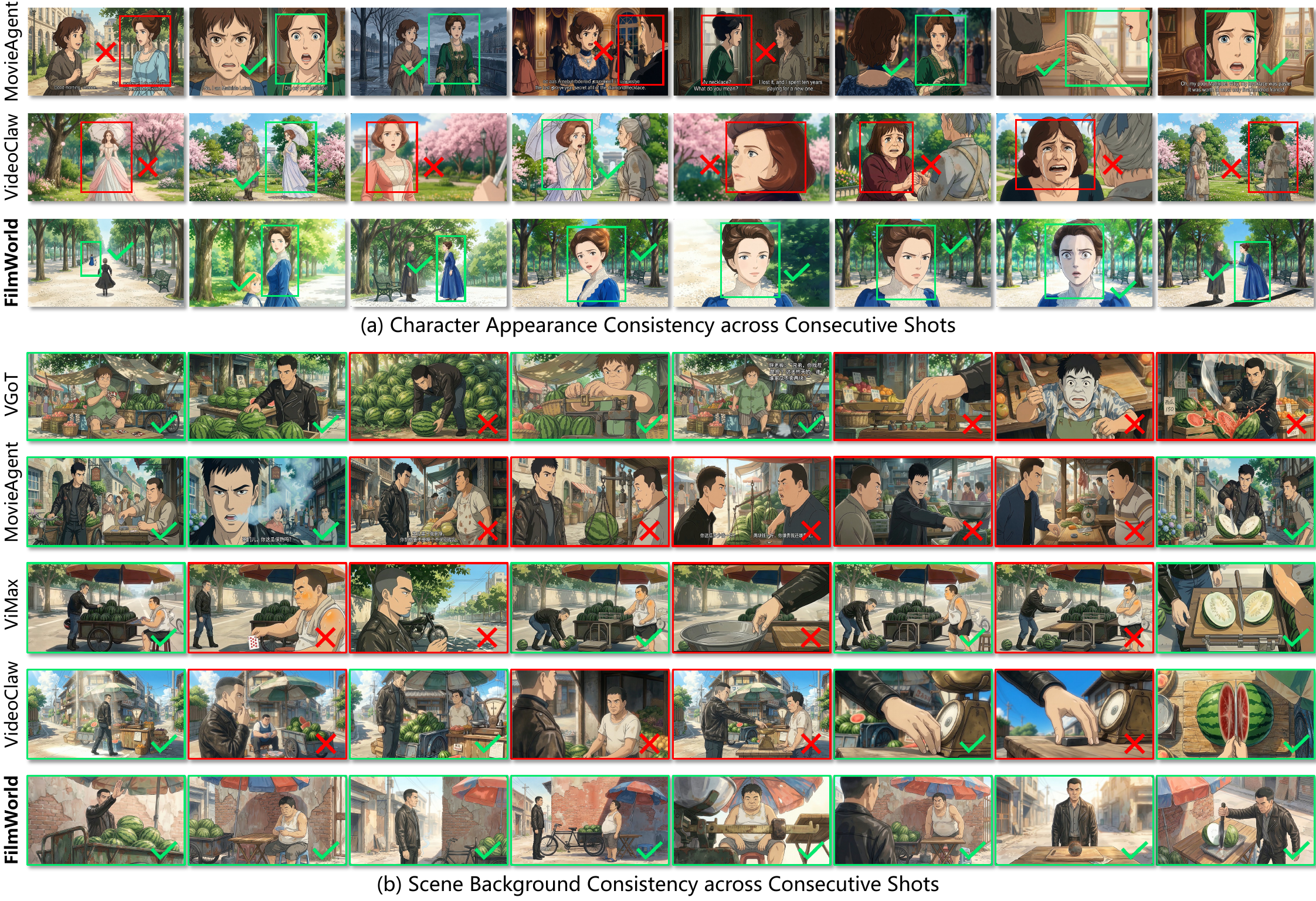}  
  \caption{\textbf{Short-range visual consistency comparison across consecutive shots.} (a) Character appearance consistency evaluated on the climactic revelation scene of \textit{The Necklace}, focusing on the visual stability of the Countess. (b) Scene background consistency evaluated on the watermelon-splitting scene of \textit{Huaqiang Buying Watermelons}, focusing on background layout and spatial arrangement. Green checkmarks and red crosses indicate whether each shot is visually consistent with the dominant appearance of the scene segment.}
  \label{fig:short_consistency}  
\end{figure}

\subsection{Qualitative Analysis}

\begin{figure}[htb]  
  \centering  
  \includegraphics[width=\textwidth]{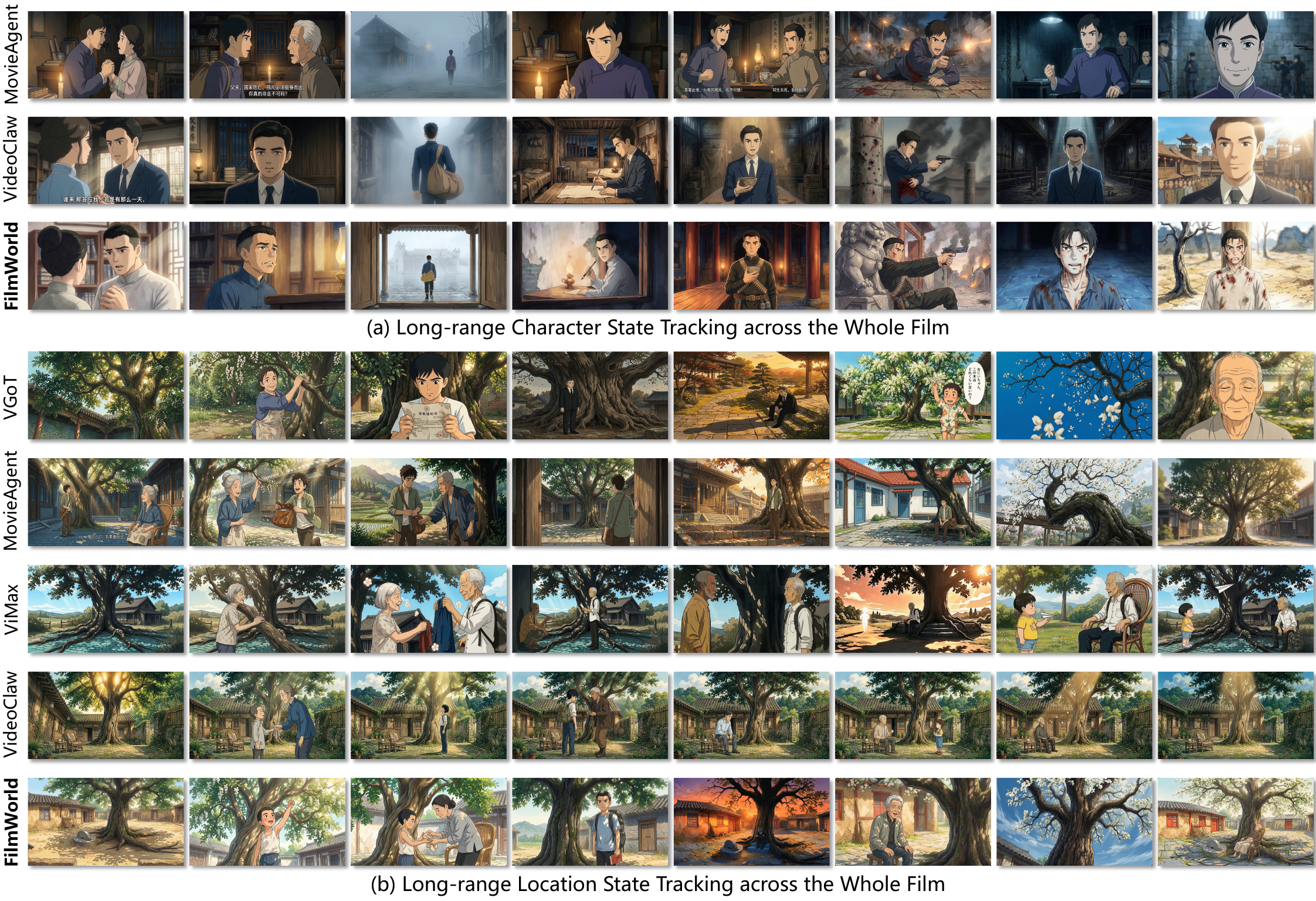}  
  \vspace{-6mm}
  \caption{\textbf{Long-range visual state tracking comparison across the whole film.} (a) Character state tracking on \textit{White Silk in a Chilly Spring}. FilmWorld correctly updates the protagonist Lin Juemin's appearance across five narrative-driven states (daily, travel, combat, interrogation, and execution attire), while baseline methods keep the character's appearance nearly unchanged regardless of plot progression. (b) Location state tracking on \textit{The Bond of Time}. FilmWorld preserves the identity of an old locust tree while faithfully reflecting its temporal evolution, whereas baseline methods either produce static, copy-paste-like appearances or incoherent scenes with little visual consistency.}
  \label{fig:long_consistency}  
\end{figure}

\textbf{Visual Consistency across Temporal Scales.}
We evaluate visual consistency at two complementary temporal scales. 1) Short-range consistency. Figure~\ref{fig:short_consistency} compares visual consistency across consecutive shots. For character appearance, evaluated on the climactic revelation scene of \textit{The Necklace}, FilmWorld maintains stable identity of the Countess while baselines exhibit noticeable appearance drift. For scene background, evaluated on \textit{Huaqiang Buying Watermelons}, FilmWorld preserves consistent spatial layout across shots whereas competing methods produce structurally inconsistent backgrounds. 2) Long-range consistency. Figure~\ref{fig:long_consistency} examines visual state tracking over the full film, targeting a more demanding capability: accurately evolving entity states in accordance with narrative progression. For character state tracking on \textit{White Silk in a Chilly Spring}, FilmWorld correctly updates the protagonist's appearance across five narrative-driven states, while baselines produce near-static character appearances throughout. For location state tracking on \textit{The Bond of Time}, FilmWorld faithfully reflects the temporal evolution of a central old locust tree, whereas baselines either generate static copy-paste-like appearances or temporally incoherent scenes. These results confirm that FilmWorld's explicit world state modeling enables coherent entity evolution at both temporal scales, a capability that existing baselines systematically fail to achieve.

\begin{figure}[H]  
  \centering  
  \includegraphics[width=\textwidth]{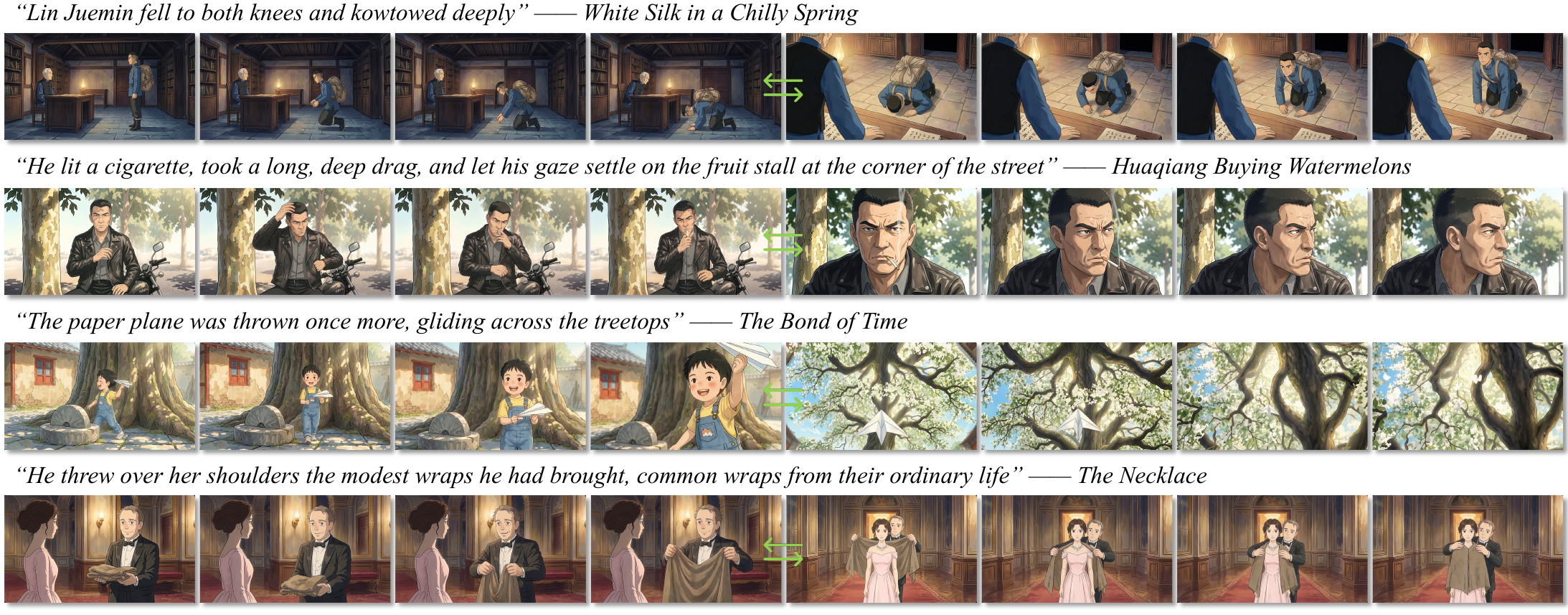}  
  \caption{\textbf{Inter-shot state continuity demonstrated by FilmWorld.} Each row shows a representative shot transition from a different source novel, with the green arrow marking the shot boundary. The left frames show the tail of the preceding shot and the right frames show the head of the succeeding shot. FilmWorld maintains seamless physical and spatial continuity at shot boundaries, with entity poses, spatial layouts, and object states naturally carried over across shots.}
  \label{fig:state_continuity}  
\end{figure}

\textbf{Inter-Shot State Continuity.} Figure~\ref{fig:state_continuity} presents representative shot transitions from four source novels, each illustrating FilmWorld's ability to maintain physical and spatial continuity across shot boundaries. In each case, the terminal state of the preceding shot, including entity poses, object placements, and spatial layout, is seamlessly carried over as the opening configuration of the succeeding shot. For instance, Lin Juemin's kneeling posture at the end of the preceding shot is precisely preserved as he kowtows at the opening of the 

next shot in \textit{White Silk in a Chilly Spring}, and the draping motion of the wrap over the woman's shoulders flows without discontinuity across the shot boundary in \textit{The Necklace}. This continuity is a direct consequence of FilmWorld's dual-channel continuity conditioning: the textual end-state constraint prescribes the precise terminal configuration that each shot must arrive at, while the visual keyframe channel ensures that the perceptual realization of the boundary is spatially and photographically coherent. Together, these two channels elevate inter-shot continuity from a best-effort heuristic into a structural guarantee, enabling FilmWorld to sustain physically plausible and narratively coherent action sequences across shot boundaries.

\begin{figure}[t]  
  \centering  
  \includegraphics[width=\textwidth]{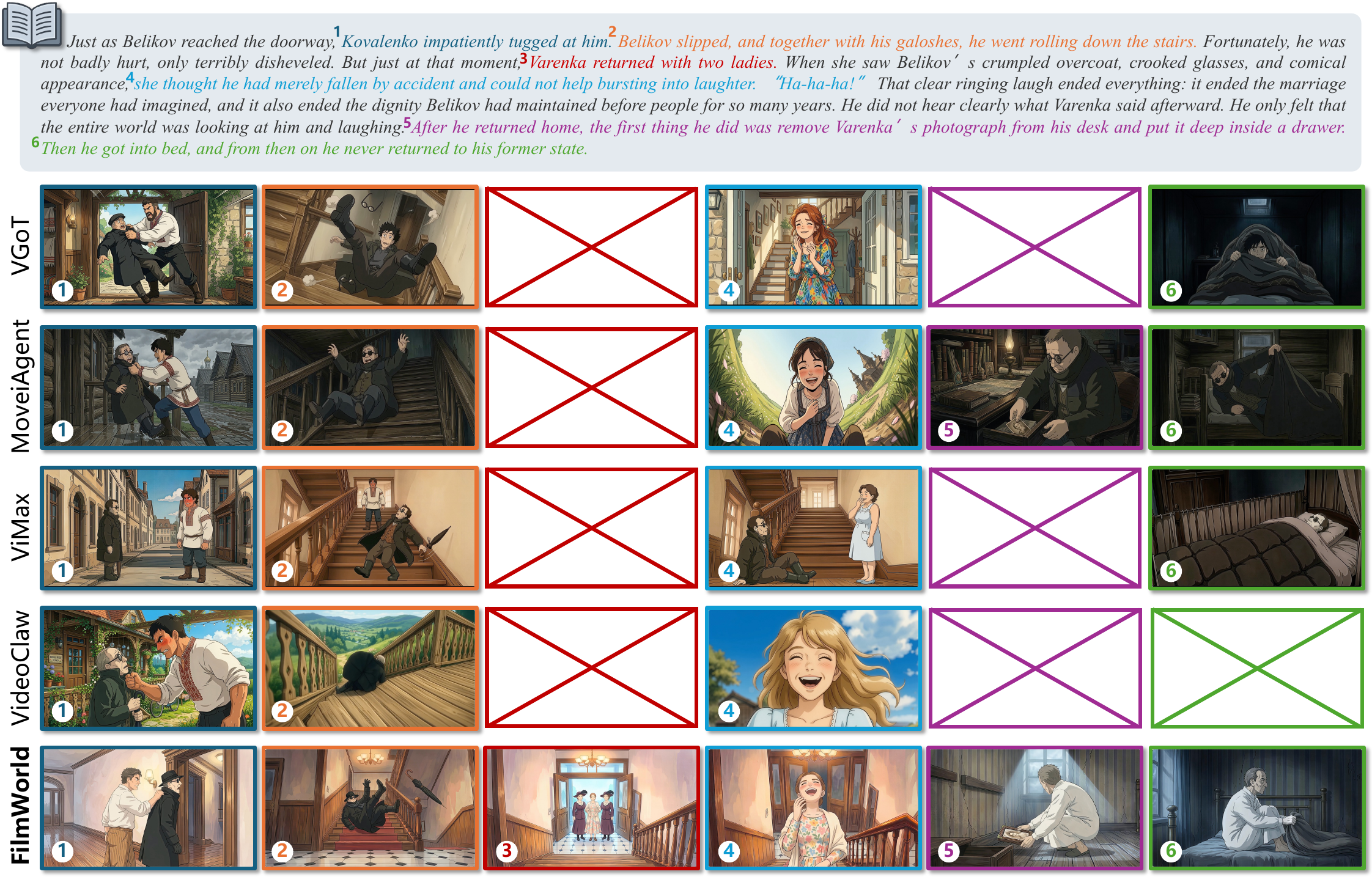}  
  \caption{\textbf{Narrative fidelity comparison on a key plot sequence from \textit{The Man in a Case}.} The source passage describes six sequential plot events, numbered 1--6: Kovalenko tugging at Belikov, Belikov tumbling down the stairs, Varenka returning with two ladies, Varenka bursting into laughter, Belikov removing her photograph, and Belikov taking to bed. Empty frames indicate plot events that a method fails to generate. All baseline methods omit at least two plot events, while FilmWorld is the only method that faithfully restores all six events in correct narrative order.}
  \label{fig:narrative_fidelity}  
\end{figure}
\textbf{Narrative Fidelity and Plot Restoration.}
Figure~\ref{fig:narrative_fidelity} presents a qualitative comparison of narrative fidelity on a key plot sequence from \textit{The Man in a Case}, comprising six causally chained plot events. All baseline methods fail to generate at least one events, with VideoClaw~\cite{videoclaw2026} omitting three events, resulting in severe narrative gaps that break the causal logic of the scene. Moreover, even for the events that baselines do generate, the visual content often deviates from the source text in character identity, spatial layout, or scene context. In contrast, FilmWorld faithfully restores all six plot events in correct narrative order, with spatially coherent scene compositions that create a convincing sense of continuous space throughout the sequence, whereas baseline methods produce visually abrupt transitions that undermine narrative immersion. These results demonstrate that FilmWorld's explicit world state modeling, grounded in a symbolic state trajectory derived from the novel, enables systematic and fine-grained plot restoration that existing baselines fail to achieve.

\begin{figure}[htb]  
  \centering  
  \includegraphics[width=\textwidth]{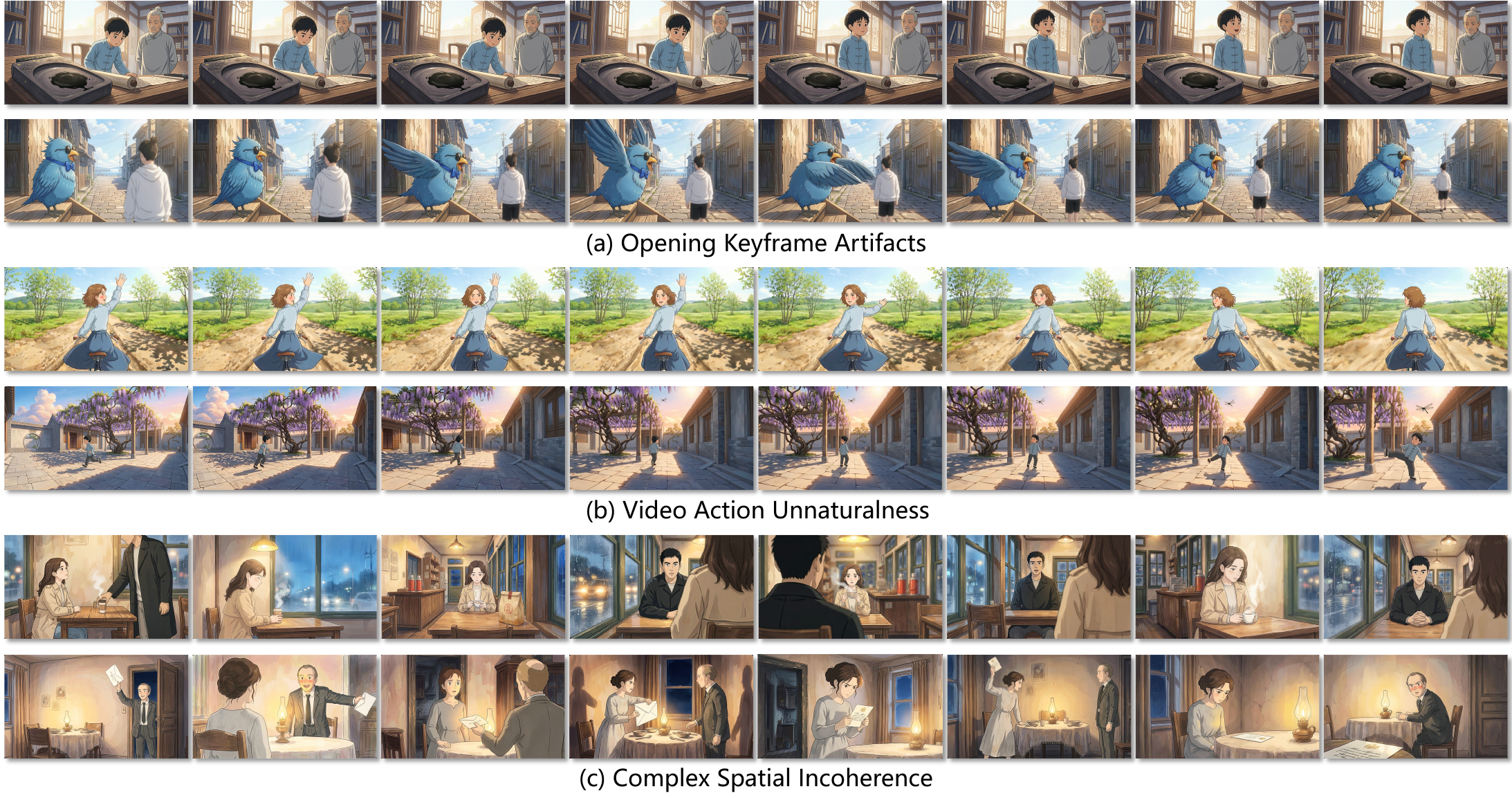}  
  \caption{\textbf{Failure cases of FilmWorld.} These failures mainly stem from the underlying generative backbones. (a) \textit{Opening Keyframe Artifacts}: keyframe synthesis occasionally produces structurally implausible entities, or unreasonable spatial relationships and scale among characters. (b) \textit{Video Action Unnaturalness}: despite accurate motion specifications, the video generator sometimes produces kinematically implausible or abrupt actions. (c) \textit{Complex Spatial Incoherence}: in scenes with intricate layouts, entities can appear spatially unanchored, drifting in position across frames.}
  \label{fig:failure_case}  
\end{figure}
\textbf{Failure Case Analysis.} While explicit world-state modeling substantially improves consistency, FilmWorld is not immune to failure. Figure~\ref{fig:failure_case} examines representative cases. Notably, the world state and shot directives supplied to the renderer remain correct in each case, indicating that these failures stem from the underlying generative backbones rather than the world-modeling formulation itself. (a) Opening keyframe artifacts arise when the image generator produces structurally implausible entities, or unreasonable spatial relationships and scale among characters, under complex conditioning bundles. (b) Video action unnaturalness manifests as kinematically implausible or abrupt actions despite accurate motion specifications, reflecting limitations in the video generator's native motion synthesis. (c) Complex spatial incoherence emerges in intricate layouts, where entities appear spatially unanchored, drifting in position or scale across frames. Such defects often persist because they are localized or subtle relative to the overall composition, making them difficult for the closed-loop verifier to reliably detect and correct. Collectively, these cases delineate the boundary of what explicit world-state modeling can guarantee: FilmWorld enforces consistency at the symbolic level, but pixel-level fidelity remains bounded by the perceptual capacity of the foundation generative and verification models it orchestrates. Since our framework is agnostic to the choice of underlying backbones, we expect these failure modes to be progressively alleviated as foundation image, video, and vision-language models continue to advance.

\section{Conclusion}
In this paper, we revisit novel-to-film generation from a representational perspective and formalize it as dynamic cinematic world modeling, decomposed into a construction phase that grounds abstract literary prose into stateful world entities and an evolution phase that updates them causally across scenes. Guided by this formalization, we propose FilmWorld, an end-to-end agentic system in which six specialized agent groups jointly instantiate and evolve the dynamic cinematic world, with state trajectories fully resolved at the symbolic level to decouple rendering from sequential dependencies and enable constant-latency parallel synthesis. To assess long-form generation, we further introduce FilmEval, a difficulty-graded benchmark of 15 novels coupled with nine automatic metrics spanning cinematic presentation, film consistency, and novel fidelity. Extensive experiments show that FilmWorld consistently surpasses state-of-the-art baselines across all metrics and difficulty tiers, with pronounced gains on narrative fidelity and cross-scene consistency, and remains stable as narrative complexity scales. 
\clearpage
\newpage
\bibliographystyle{plainnat}
\bibliography{main}
\newpage
\beginappendix
\section{FilmEval Benchmark Details}

This appendix expands the one-paragraph description of the FilmEval benchmark in Section 5.2 into a per-novel, auditable form. Appendix~\ref{sec:a1} provides the full metadata table for the 15 novels, and Appendix~\ref{sec:a2} specifies how the source texts can be accessed.

\subsection{Overview of the 15-Novel Dataset}
\label{sec:a1}

FilmEval consists of 15 source novels stratified into three difficulty tiers by character count, plot complexity, and world-building richness. Table~\ref{tab:filmeval_meta} lists the complete metadata for each novel, keyed by a canonical ID that is used for cross-reference. The dataset spans 6 English and 9 Chinese novels and draws on three distinct origins: (i) six \emph{novel adaptations}, rewritten from pre-existing literary works; (ii) three \emph{screen adaptations}, re-novelized from existing film/TV works; and (iii) six \emph{original} contemporary works created specifically for this benchmark. All adapted works, whether from novels or from screen sources, were independently rewritten rather than directly reproduced, in order to fit the length, structure, and narrative conventions, while preserving the core characters, settings, and plot of their respective originals.

\begin{table}[htb]
\centering
\caption{\textbf{Metadata for the 15 source novels comprising the FilmEval benchmark.} \#Char., \#Loc., and \#Events denote the number of narratively salient characters, key locations, and key plot events, respectively. Length is reported in English words for English sources and in Chinese characters for Chinese sources. The Origin column distinguishes three provenance types: \emph{Adapted (novel)} (rewritten from a pre-existing literary work), \emph{Adapted (screen)} (re-novelized from an existing film/TV work), and \emph{Original} (created specifically for this benchmark).}
\label{tab:filmeval_meta}
\resizebox{\textwidth}{!}{%
\begin{tabular}{lrrrrrrrrr}
\toprule
\rowcolor{headercolor}
ID & Title & Lang. & Length & \#Char. & \#Loc. & \#Events & Genre & Origin & Tier \\
\midrule
\rowcolor{easycolor}
C1 & Slot Seven & English & 1{,}015 words & 4 & 2 & 6 & Slice of life & Original & Easy \\
\rowcolor{easycolor}
C2 & The Flower Aside the Window & English & 1{,}149 words & 5 & 2 & 7 & Slice of life & Original & Easy \\
\rowcolor{easycolor}
C3 & A Lamp (\begin{CJK}{UTF8}{gkai}一盏灯\end{CJK}) & Chinese & 1{,}633 chars & 3 & 3 & 6 & Warm realism & Original & Easy \\
\rowcolor{easycolor}
C4 & Huaqiang Buying Watermelons (\begin{CJK}{UTF8}{gkai}华强买瓜\end{CJK}) & Chinese & 1{,}492 chars & 2 & 1 & 6 & Comedic satire & Adapted (screen, \emph{Conquer}) & Easy \\
\rowcolor{easycolor}
C5 & Meeting at the Station (\begin{CJK}{UTF8}{gkai}车站相遇\end{CJK}) & Chinese & 1{,}614 chars & 2 & 3 & 6 & Romance & Original & Easy \\
\rowcolor{medcolor}
C6 & A Chameleon & English & 1{,}348 words & 4 & 2 & 6 & Satire & Adapted (novel, Chekhov) & Medium \\
\rowcolor{medcolor}
C7 & The Gift of the Magi & English & 2{,}067 words & 3 & 3 & 6 & Romance / irony & Adapted (novel, O. Henry) & Medium \\
\rowcolor{medcolor}
C8 & The Bond of Time (\begin{CJK}{UTF8}{gkai}光阴的结\end{CJK}) & Chinese & 2{,}214 chars & 5 & 2 & 6 & Lyrical realism & Original & Medium \\
\rowcolor{medcolor}
C9 & Listening-to-the-Wind Courtyard (\begin{CJK}{UTF8}{gkai}听风小院\end{CJK}) & Chinese & 2{,}996 chars & 6 & 2 & 7 & Family saga & Original & Medium \\
\rowcolor{medcolor}
C10 & Su Erqi (\begin{CJK}{UTF8}{gkai}苏二七\end{CJK}) & Chinese & 2{,}907 chars & 4 & 2 & 6 & Tragic realism & Adapted (novel, Lu Xun) & Medium \\
\rowcolor{hardcolor}
C11 & The Man in a Case & English & 2{,}334 words & 5 & 4 & 8 & Satire & Adapted (novel, Chekhov) & Hard \\
\rowcolor{hardcolor}
C12 & The Necklace & English & 2{,}733 words & 5 & 5 & 8 & Realism / irony & Adapted (novel, Maupassant) & Hard \\
\rowcolor{hardcolor}
C13 & The Irreproducible Negative (\begin{CJK}{UTF8}{gkai}无法重写的底片\end{CJK}) & Chinese & 4{,}829 chars & 6 & 4 & 9 & Time-travel drama & Adapted (screen, \emph{Link Click}) & Hard \\
\rowcolor{hardcolor}
C14 & White Silk in a Chilly Spring (\begin{CJK}{UTF8}{gkai}春寒里的白方巾\end{CJK}) & Chinese & 4{,}334 chars & 6 & 4 & 9 & Historical fiction & Adapted (novel, Lin Jue-min) & Hard \\
\rowcolor{hardcolor}
C15 & I Will Take This Job (\begin{CJK}{UTF8}{gkai}这单我接了\end{CJK}) & Chinese & 4{,}663 chars & 5 & 4 & 8 & Comedic fantasy & Adapted (screen, \emph{Scissor Seven}) & Hard \\
\bottomrule
\end{tabular}
}
\end{table}

The three count columns follow a consistent convention: \#Characters counts recurring characters, excluding one-off background figures; \#Key Locations counts revisited or narratively salient settings, excluding transient pass-through locations; \#Key Plot Events counts the major plot beats of the story, not every micro-action.

Regarding provenance, six novels are \emph{novel adaptations}: four from canonical short stories by Chekhov, O. Henry, and Maupassant (C6, C7, C11, C12), and two from Chinese literary texts-\emph{Su Erqi} (C10, from Lu Xun's \emph{Kong Yiji}) and \emph{White Silk in a Chilly Spring} (C14, from Lin Jue-min's \emph{Letter to My Wife}). Three novels are \emph{screen adaptations}, re-novelized from existing film/TV works: \emph{Huaqiang Buying Watermelons} (C4, from the drama \emph{Conquer}), \emph{The Irreproducible Negative} (C13, from the animated series \emph{Link Click}) and \emph{I Will Take This Job} (C15, from the animated series \emph{Scissor Seven}). The remaining six novels (C1, C2, C3, C5, C8, C9) are wholly original, ensuring that at least half the dataset is free of prior-work bias.

\subsection{Source Text Access}
\label{sec:a2}

For space reasons, we embed the full text of one representative novel and provide access paths for the remaining fourteen. We select \textit{The Bond of Time} (\begin{CJK}{UTF8}{gkai}光阴的结\end{CJK}, C8) as the example, since it is the running example in Figure 2 of the main paper and the subject of the long-range location-state tracking case study (the old locust tree). \textbf{For the other fourteen novels, please refer to our project page.}

To let readers grasp each story without consulting external sources, we give a short summary for every novel below, followed by the full embedded text of C8.

\textbf{C1 - Slot Seven.} Librarian Ethan discovers that Slot Seven of the vending machine keeps getting stuck on milk, so he leaves a reminder note beside it. Soon strangers begin leaving their own notes there too-about exams, weather, private worries-and Slot Seven gradually becomes a quiet emotional corner. During finals week, a note reading "I failed my exam today and don't want to go back to my dorm" prompts Ethan to write back words of comfort. Later, when the machine is finally repaired, he keeps the corner anyway, so that anyone exhausted can still pause, write a line, and catch their breath.

\textbf{C2 - The Flower Aside the Window.} Maya works at a small café by the street and places a daisy by the window every day. Every morning, Mr. Green comes in for black coffee and gazes at the flower shop across the street-the place his late wife once loved most. Maya wants to keep the window seat reserved for him but is stopped by her boss. Visiting the flower shop, she learns that Mrs. Green used to buy just one, the cheapest flower, every week before she passed. So Maya leaves a note and a daisy, asking whoever sits by the window that day to look at the flower on someone else's behalf. The window seat no longer belongs to anyone in particular, yet the quiet act of remembrance passes silently between strangers.

\textbf{C3 - A Lamp (\begin{CJK}{UTF8}{gkai}一盏灯\end{CJK}).} On a rainy night, middle schooler Lin Che takes shelter at a bus stop, where an old man selling roasted sweet potatoes hands him one for free to warm his hands, asking only that he help someone else in need someday. Lin Che wants to repay him, but the old man is nowhere to be found for days. A week later, during a community volunteer activity, he recognizes the old man's home-the very person, laid up with a sore leg and unable to work his stall. Lin Che does extra chores to repay the kindness and keeps checking in on him afterward. Once recovered, the old man returns to his stall, now with a wooden sign: on cold days, take a sweet potato for free, and pass the kindness on to someone else. Goodness is like one lamp lighting another.

\textbf{C4 - Huaqiang Buying Watermelons (\begin{CJK}{UTF8}{gkai}华强买瓜\end{CJK}).} On a scorching summer afternoon, Liu Huaqiang goes to a street corner fruit stand to buy a watermelon and cool off, asking the vendor "Is it ripe?" The plump vendor is slick and evasive. As the melon is weighed, the scale tips suspiciously high; Huaqiang reaches under the scale pan and pulls out a magnet, then cuts the melon open with one stroke-revealing pale, unripe flesh. Calmly, he lectures the vendor: hiding a magnet under the scale is greed, selling unripe melons as ripe is deceit; cheating him today could mean cheating a child saving up pocket money tomorrow. He leaves the money and walks off; the vendor, ashamed, discards the magnet and replaces the scale.

\textbf{C5 - Meeting at the Station (\begin{CJK}{UTF8}{gkai}车站相遇\end{CJK}).} On a rainy evening, Lin Xia lingers alone at an old train station, holding freshly bought osmanthus cake, when she spots Zhou Yuan-the man who left without a word years ago-in the crowd. They catch up over tea at a nearby shop, where Zhou Yuan apologizes for leaving so abruptly back then without saying goodbye. He tells her that over the years, though he's traveled to many seas, he always remembered how she wanted to see a quiet winter sea. He has come back to invite her along, promising never to leave quietly again. As the train arrives, this time the two walk side by side toward the door-toward a reunion that has come late, but not too late.

\textbf{C6 - A Chameleon.} Police officer Ochumelov crosses the marketplace when goldsmith Khryukin accuses a stray dog of biting his finger. At first, the officer sternly insists the dog's owner must be punished, but the moment the crowd speculates that the dog might belong to the General, he immediately flips and accuses Khryukin of provoking it himself. As rumors about "whether it's the General's dog" keep shifting back and forth, so does the officer's attitude, swinging hot and cold like a chameleon. Finally, the General's cook confirms the dog belongs to the General's brother, and the officer breaks into a grin, releasing the dog while turning to mock and threaten Khryukin-laying bare the sycophantic ugliness of officialdom.

\textbf{C7 - The Gift of the Magi.} On Christmas Eve, Della has saved only one dollar and eighty-seven cents-not enough to buy her beloved husband Jim a gift. In anguish, she cuts off and sells her prized long hair to buy a fine platinum watch chain for Jim's heirloom gold watch. But Jim comes home looking strange-he has sold the watch to buy Della the tortoiseshell combs she'd long admired in a shop window. Both gifts instantly become useless, yet they prove the boundless love each holds for the other. The author notes that these two foolish children, who sacrificed their most treasured possessions for each other, are in fact the wisest of magi.

\textbf{C8 - The Bond of Time (\begin{CJK}{UTF8}{gkai}光阴的结\end{CJK}).} The old locust tree behind the family home binds Lin Yuan's whole life to it. As a child, he reaches for locust blossoms and has the sticky sap wiped from his hands by his mother; after the college entrance exam, his father presses pocket money into his hand as he sends him off to the city. Busy building a life there, Lin Yuan cannot make it home when his mother calls to say the old tree is dying. His father passes away quietly one frosty morning, and when Lin Yuan finally returns, he presses his hand against the tree's bark, cracked like his father's palm, in deep regret. Years later, elderly, he returns home for good, and one afternoon, sitting beneath the tree, he answers his little grandson's question-"does the tree hurt?"-before passing away peacefully in late spring, a locust blossom falling onto his lap. Like the old tree, a life eventually returns its bones to the earth.

\textbf{C9 - Listening-to-the-Wind Courtyard (\begin{CJK}{UTF8}{gkai}听风小院\end{CJK}).} Old scholar Lin, a Qing-dynasty-era graduate, is known for the wisteria vine in his "Listening-to-the-Wind Courtyard," teaching his grandson Wenbo that sitting quietly to truly hear the wind reveals the truth of the world. Half a century of upheaval later, the courtyard has become a crowded, run-down compound; middle-aged machinist Wenbo alone waters the old vine late at night, recording half a lifetime of ups and downs in a diary hidden in a brick crevice. After Wenbo passes away, his daughter Lin Yue, who lives abroad and inherits the property, nearly signs away the courtyard for a lucrative buyout from developers-until the old vine stirs memories of her father and she refuses. She finds the diary, and pours herself into faithfully restoring the courtyard, finally sitting beneath the vine and understanding the wind's message across three generations.

\textbf{C10 - Su Erqi (\begin{CJK}{UTF8}{gkai}苏二七\end{CJK}).} At the wine stall beneath the old banyan tree by the Lingdang Town ferry crossing, Su Erqi is the only patron who wears a long scholar's robe yet drinks standing up. He writes beautiful calligraphy, yet is lazy and occasionally steals books, always spouting classical phrases, his face often marked with fresh bruises from beatings, making him the target of others' mockery. He patiently teaches a young firewood-tender, Ashui, to write the character \begin{CJK}{UTF8}{gkai}"茴"\end{CJK} (fennel), and shares fennel beans with wandering children. After being caught stealing from a wealthy household past mid-autumn, his legs are broken, and he disappears for a long while; on a cold, foggy early winter day, he "walks" in on a worn-out straw mat to have one last bowl of wine, pays with muddy copper coins, and vanishes-presumably swallowed by the icy river.

\textbf{C11 - The Man in a Case.} Greek teacher Belikov is a "man in a case": he wears galoshes and carries an umbrella even on sunny days, keeps his umbrella, watch, even his penknife all tucked into cases, terrified above all of anything irregular, his catchphrase being "whatever happens, let's hope nothing comes of it." His timidity keeps the whole school, even the whole town, subdued for years. When new teacher Kovalenko and his sister arrive, matchmaking gossip has Belikov even considering marriage. But a caricature deeply upsets him, and upon seeing the siblings riding bicycles, he goes to lecture them, only to be pushed down the stairs; Varenka's laughter shatters what dignity he had left. He takes to his bed afterward and dies a month later; the town feels briefly relieved before settling back into its old, oppressive routine.

\textbf{C12 - The Necklace.} Beautiful but vain, Mathilde marries a minor clerk and constantly resents their modest life. To attend a ministry ball, her husband uses their savings to buy her a dress, and she borrows a diamond necklace from her wealthy former schoolmate, Madame Forestier. She dazzles at the ball, but on the way home discovers the necklace is lost. Too afraid to admit it, the couple takes out crushing loans to buy an identical necklace worth thirty-six thousand francs to replace it, spending the next ten years in grueling poverty to repay the debt. A decade later, she runs into her friend on the street and confesses the truth-only to learn that the original necklace was fake, worth at most five hundred francs.

\textbf{C13 - The Irreproducible Negative (\begin{CJK}{UTF8}{gkai}无法重写的底片\end{CJK}).} At Time Photo Studio, one can slip into an old photograph and briefly speak through it. Chen Xiao wants to return to the most ordinary day ten years ago to say, through his younger self, three things he's held in for a decade. Chen Xiao possesses his younger body and helps him win a basketball championship, confess to the girl he secretly loved, and apologize to his mother. Just as the three matters are settled, he sees a date on TV-May 12, 2008, the day of the Wenchuan earthquake. He goes door to door trying to warn people, but no one believes him; only his mother, without hesitation, grabs him to leave-yet the earthquake strikes just as they cross the threshold, and she shields him with her back until the end. Back in the present, he understands: the moment of death cannot be rewritten, but speaking one's regrets aloud lets the ones we love hear them, even if only once.

\textbf{C14 - White Silk in a Chilly Spring (\begin{CJK}{UTF8}{gkai}春寒里的白方巾\end{CJK}).} In the spring of 1911, peach blossoms bloom in full at Sanfang Qixiang, Fuzhou. Lin Juemin, recently returned from studying in Japan, shares deep affection with his seven-months-pregnant wife, Chen Yiying, yet his heart is troubled by the nation's turmoil. He kneels to bid farewell to his aging father before setting off for the Guangzhou uprising, secretly returning home once more to see his wife before leaving. In the dim lamplight of Guangzhou, he writes, through tears, his farewell letter to his wife on a white silk handkerchief. Wounded and captured during the uprising, he speaks with unwavering conviction at his trial and is executed with dignity, at only 24 years old. Upon hearing the news, Yiying faints and gives birth prematurely, dying of grief two years later. A century on, the ink on the white handkerchief remains as fresh as ever, transformed into the ordinary spring breeze and sound of reading in today's peaceful era.

\textbf{C15 - I Will Take This Job (\begin{CJK}{UTF8}{gkai}这单我接了\end{CJK}).} On Little Chicken Island, at a shop that doubles as a barbershop and an assassin's front, hapless assassin Ah Qi is sent by his agent to carry out a five-thousand-yuan hit, the target being a long-haired girl on Long Street. Ah Qi keeps mistaking the wrong people for his target in the busy streets, but during an incident where he helps catch a pickpocket, he meets the calm and capable girl Bai Yuan. He tags along as she buys fever medicine for her sick younger brother, secretly making up the coins she was short. Bai Yuan admits she is being hunted because she took a ledger exposing smuggled contraband drugs at the docks. Only then does Ah Qi realize this job was never a clean one. As the pursuers break down the door, he grips his scissors and steps in front of the girl, smiling as he says: this job-I'll take it.

\textbf{Full text of C8 - The Bond of Time (\begin{CJK}{UTF8}{gkai}光阴的结\end{CJK}).} 

The original Chinese source text is reproduced verbatim below.

\begin{storyquote} 
\begin{CJK}{UTF8}{gkai}
\setlength{\parindent}{2em}
\setlength{\parskip}{0.5em}
老屋后头那棵老槐树，根须深扎进黄土，也悄然系住了林远一生的命脉。它是记忆的坐标，风一过，年轮便漾开一圈圈往事。

槐花开得最盛的年头，林远才刚能勉强够到藤椅的扶手。阳光穿过层层叠叠的枝叶，碎金般洒在青砖地上，随着微风明明灭灭。他踮着脚，双手死死攥住一根被汗水浸得滑腻的长竹竿，够向最高处那串垂落的洁白。“妈，你看那串，白得发亮！”他仰着脖子喊。母亲停下针，笑着嗔怪：“急什么，槐花落不了，妈给你摇下来便是。”她起身，枯瘦的手握住低处的枝干轻轻一撼，簌簌落下一阵香雨。林远张开旧褂子的下摆去接，指尖沾上黏稠的树汁，母亲便用粗布帕子替他擦拭：“傻孩子，沾了槐胶，洗都洗不掉。”他咯咯笑着躲开，却不知那抹甜香，早已渗进骨血。那时的光阴是黏稠的，慢得能看清一只蚂蚁如何驮着花瓣翻过门槛。林远以为，这样的日子会一直亮堂堂地铺展下去，仿佛只要伸出手，就能把整个夏天妥帖地揣进兜里。

蝉鸣撕扯着闷热的空气，像一把钝锯来回拉扯神经。高考放榜后的午后，林远立在树下，指尖死死捏着那张轻飘飘的录取通知书，纸边已被汗水洇出深色的折痕。墙根下，父亲蹲在阴影里，旱烟的辛辣呛得人眼眶发酸。他终于磕了磕烟袋，将一沓叠得整整齐齐的零钱塞进林远背包侧袋。“外头不比家里，”他别过脸，喉结滚动了一下，“钱揣紧，别饿着。路是自己选的，跪着也得走完。”林远鼻尖一酸，只重重“嗯”了一声。他转身跨过高高的门槛，听见身后传来父亲低哑的补白：“……逢年过节，打个电话。”风卷起院角的落叶，他没有回头，只在心里默念：等我出息了，接你们去城里。巷口的风推着后背，故乡与老槐在视野里急速坍缩，最终凝成地平线上一个颤动的墨点。青春原是一场蓄谋已久的远行，他贪恋着远方未名的灯火，急于挣脱这方被蝉鸣与槐香腌入味的、静得令人心慌的故土。

城市拔地而起，玻璃幕墙将天空切割成规整的几何体，也把日子切成按秒计费的碎片。林远坐在二十七层的工位上，望着脚下钢铁与玻璃交织的洪流，瞳孔里映不出一点温度。手机在桌面震动，母亲的嗓音隔着千里电流传来，轻得像怕惊碎什么：“槐树……怕是不行了，叶子黄了一大半。”他压低声音，手指无意识地摩挲着咖啡杯沿：“妈，怎么了？根出问题了吗？”“老毛病，烂了半边，请人看了，说是得打个木架撑住。”母亲顿了顿，声音轻下去，“你忙你的，不用挂心。”电话那头传来窸窣的咳嗽声，他心头一紧：“我周末回去一趟……”“别！”母亲急忙打断，语气里透着不容商量的温柔，“你张总不是催方案吗？家里都好，真没事。”他张了张嘴，视线却撞上主管催促的眼神和同事疾走的皮鞋。键盘的敲击声如潮水般涌来，将他那点微弱的归乡念头拍碎。良久，他只挤出一句：“转些钱回去，找师傅挑好木料。”挂断后，隔壁工位的老李探头递来一摞文件：“林远，客户下午要终稿，今晚还得熬。”他扯出一个笑：“好，放这儿吧。”屏幕幽光映着疲惫的脸，他摸出烟盒，打火机的火苗窜起又熄灭。辛辣的烟气钻进肺腑，呛得他眼眶发烫，却终究没落下一滴泪。他在钢筋水泥的丛林里奋力泅渡，却渐渐忘了出发的码头。那些曾以为能照亮一生的星辰，早被月供、绩效与深夜的咖啡渍，研磨成地毯上拂不去的灰。

人生行至中途，悲欢都已有了重量。父亲走在一个霜降后的清晨，没有惊动任何人，只像一片熟透的秋叶，悄无声息地归根。隔壁的王伯拄着拐杖立在院墙外，叹道：“你爸走前，还天天往树下转悠，说这树跟他一样，老了，得歇歇了。”林远递上一支烟，王伯摆摆手：“戒了，医生说肺不行。”他沉默着推开虚掩的院门。老槐已至暮年，树皮皲裂翻卷，沟壑纵横，竟与父亲临终前那双枯槁的手掌如出一辙。他缓缓贴上树干，粗粝的触感顺着指尖直抵心口。“爸，”他低声说，声音散在秋风里，“我回来了。”没有回音，只有落叶擦过石阶的轻响。一股钝痛无声漫开。他忽然懂了，生命原是一场无声的交接--从泥土里挣出，在风雨中挺立，终将把筋骨还给大地。他在树下的石阶上枯坐至日暮，余晖将他的影子拉得极长，长到仿佛能触到院墙那头。恍惚间，一个赤脚的小男孩正拨开花枝朝他奔来，笑声清亮，惊飞了枝头的麻雀。

褪去一身疲惫，林远循着旧梦重返小镇。老屋重漆了门窗，换了亮瓦，唯有那棵槐树，依旧沉默地立在院后。枝干已佝偻如弓，却每逢春深，依旧撑开一树如雪的繁花。“爷爷，这树多老啦？”小孙子举着纸飞机跑过来，气喘吁吁地问。林远抿了口茶，目光抚过虬结的枝干：“比你太爷爷的年纪还大哩。”“那它疼不疼？皮都裂开了。”孩子伸出小手，小心翼翼地摸了摸树皮的沟壑。林远笑了，眼角的皱纹舒展开来：“不疼。它疼过的地方，都长出新的叶子了。”纸飞机再次掷出，划过树冠。阳光穿过新叶，在地上投下与早年别无二致的光斑。风过处，花影婆娑，时光仿佛在此打了个松软的结，将祖父的凝望与孙儿的欢笑，轻轻缝进同一片年轮里。

岁月走到最后，林远的眼界已褪成一片温润的毛玻璃，万物都裹着柔光。他仍习惯坐在槐树下，一坐便是半日。风穿过疏朗的枝桠，沙沙，簌簌，像远年的私语，又像不知名的童谣。他隐约听见藤椅的吱呀声，听见“当心脚下”的轻唤，听见“往外走”的哑嗓，又听见稚嫩的童音问：“爷爷，树疼不疼？”他不再打捞过往，也不去丈量余生，心湖早已不起微澜。某个春深似海的午后，他缓缓合上双眼，唇角漾开一抹极淡的笑意。风倏然止息，一片槐花自高处旋落，不偏不倚，停在他膝头。他嘴唇微动，似有若无地吐出一句：“……不疼了。”洁白，轻盈，一如多年前，他第一次踮起脚尖时，抬头望见的模样。

\end{CJK}
\end{storyquote}

An English translation is provided below for the convenience of readers who are not proficient in Chinese.

\begin{storyquote}
\setlength{\parindent}{0pt}
\setlength{\parskip}{0.5em}
Behind the old house stood an old locust tree, its roots sunk deep into the yellow earth, quietly binding itself to the lifeline of Lin Yuan's entire existence. It was a coordinate of memory-whenever the wind passed through, its growth rings would ripple outward with waves of bygone days.

In the year the locust blossoms bloomed most abundantly, Lin Yuan could barely reach the armrest of the wicker chair. Sunlight filtered through the layered branches and leaves, scattering like broken gold across the blue brick ground, flickering with the breeze. Standing on tiptoe, he gripped with both hands a long bamboo pole made slippery with sweat, reaching toward the highest cluster of pure white blossoms hanging above. "Ma, look at that cluster-so bright and white!" he called out, craning his neck upward. His mother paused her needlework, smiling as she scolded him gently: "What's the rush? The locust flowers won't fall away. I'll shake them down for you." She rose, and with her thin, bony hand gave a gentle shake to a low branch, sending down a shower of fragrant petals. Lin Yuan spread the hem of his old jacket to catch them, sticky tree sap clinging to his fingertips. His mother wiped it away with a coarse cloth handkerchief: "Silly child, once you're covered in locust sap, it won't wash off." He giggled and dodged away, not knowing that sweet fragrance had already seeped into his very bones. Time then was thick and slow, slow enough to watch clearly how an ant carried a petal over the threshold. Lin Yuan believed such bright days would stretch on forever, as if he only had to reach out his hand to tuck the whole summer neatly into his pocket.

Cicadas tore at the stifling air like a blunt saw dragging back and forth across nerves. On the afternoon the college entrance exam results were posted, Lin Yuan stood beneath the tree, his fingers clenched tight around that weightless admission letter, its edges already stained dark with sweat. At the foot of the wall, his father crouched in the shadows, the bitter smoke of his pipe stinging until eyes reddened. Finally, he knocked the ashes from his pipe and stuffed a neatly folded wad of cash into the side pocket of Lin Yuan's backpack. "It's not like home out there," he said, turning his face away, his throat working. "Keep your money close, don't go hungry. You chose this path yourself-you'll walk it to the end, even on your knees." Lin Yuan felt a sting at the bridge of his nose and could only manage a heavy "Mm." He turned and stepped over the high threshold, hearing his father's hoarse voice trailing behind him: "...On holidays, give us a call." The wind swept up fallen leaves at the corner of the courtyard. He did not look back, only silently vowing to himself: "Once I've made something of myself, I'll bring you both to the city." The wind at the mouth of the lane pushed against his back, and his hometown with its old locust tree collapsed rapidly in his field of vision, finally condensing into a trembling black dot on the horizon. Youth was, from the start, a long-premeditated journey away. He hungered for the unnamed lights of distant places, eager to break free from this homeland-so quiet it unsettled the heart, so steeped in the flavor of cicada song and locust fragrance.

The city rose from the ground, its glass curtain walls slicing the sky into precise geometries, cutting the days too into fragments billed by the second. Lin Yuan sat at his workstation on the twenty-seventh floor, gazing down at the torrent of steel and glass beneath him, his pupils reflecting not a shred of warmth. His phone vibrated on the desk-his mother's voice traveled a thousand miles through the wire, light as if afraid of shattering something: "The locust tree... it might not make it. Half its leaves have turned yellow." He lowered his voice, fingers absently tracing the rim of his coffee cup: "Ma, what happened? Is it the roots?" "The old trouble-half of it's rotted through. Had someone look at it, they said we need to prop it up with a wooden frame." His mother paused, her voice growing softer. "You just focus on your work, don't worry about it." A rustle of coughing came through the line, and his chest tightened: "I'll come back this weekend-" "Don't!" his mother cut in quickly, her tone carrying a tenderness that brooked no argument. "Didn't you say your boss Zhang is pushing for that proposal? Everything's fine here, really." He opened his mouth, but his gaze collided with his supervisor's urging look and a colleague's hurrying footsteps. The clatter of keyboards surged in like a tide, crushing what small longing for home had stirred in him. After a long moment, he could only squeeze out: "I'll wire some money back-find a craftsman and get good timber." After hanging up, the colleague at the next desk, Old Li, leaned over with a stack of documents: "Lin Yuan, the client needs the final draft this afternoon, we'll be pulling another late night." He forced out a smile: "Sure, just leave it here." The dim glow of the screen lit his weary face. He fished out a cigarette; the lighter's flame flared up and died. The acrid smoke burrowed into his lungs, stinging his eyes hot-but in the end, not a single tear fell. He struggled to swim through this forest of concrete and steel, gradually forgetting the harbor from which he had set out. Those stars he once believed would light his whole life had long since been ground down, by mortgage payments, performance reviews, and late-night coffee stains, into dust that no longer stirred on the carpet.

Life had reached its midpoint; both joy and sorrow now carried weight. His father passed on a frost-touched morning, disturbing no one, like a fully ripened autumn leaf returning silently to its roots. Old Wang from next door leaned on his cane outside the courtyard wall and sighed: "Before your father passed, he still went to stand under that tree every day, saying it was just like him-old now, needed its rest." Lin Yuan offered him a cigarette; Old Wang waved it off: "Gave it up. Doctor says my lungs can't take it." He silently pushed open the half-closed courtyard gate. The old locust had reached its twilight years, its bark cracked and curling, furrowed like ravines-strikingly like his father's withered hands in his final days. He slowly pressed his palm against the trunk, its rough texture traveling straight from his fingertips to his heart. "Dad," he said quietly, his voice dissolving into the autumn wind, "I'm home." There was no answer, only the faint rustle of fallen leaves brushing across the stone steps. A dull ache spread silently through him. He suddenly understood-life was, from the beginning, a silent handover: wrenched from the earth, standing tall through wind and rain, ultimately returning its bones and sinews to the ground. He sat on the stone steps beneath the tree until dusk, the sunset stretching his shadow impossibly long, long enough it seemed to touch the far wall of the courtyard. In a daze, he saw a barefoot little boy pushing through the blossoming branches, running toward him, laughter clear and bright, startling the sparrows from the treetop.

Having shed the accumulated weariness of years, Lin Yuan followed an old dream back to the small town. The old house had fresh paint on its doors and windows, new glazed tiles on the roof-only the locust tree still stood silently in the back courtyard, unchanged. Its branches had grown bent like a bow, yet every deep spring it still unfurled a crown of blossoms white as snow. "Grandpa, how old is this tree?" His little grandson ran up holding a paper airplane, breathless with the question. Lin Yuan took a sip of tea, his gaze passing over the gnarled branches: "Older than your great-grandfather was." "Does it hurt? Its bark is all cracked." The child reached out a small hand, carefully touching the furrows in the bark. Lin Yuan smiled, the wrinkles at his eyes softening: "It doesn't hurt. Wherever it once hurt, new leaves have grown." The paper airplane sailed out again, cutting across the canopy. Sunlight filtered through the new leaves, casting patches of light on the ground no different from those of years long past. Where the wind passed, the flower shadows danced, and time seemed to tie itself into a soft knot here, sewing the grandfather's gaze and the grandchild's laughter gently into the very same growth ring.

As the years drew to their close, Lin Yuan's eyesight had faded to a warm, frosted glass, all things wrapped in soft light. He still liked to sit beneath the locust tree, sometimes for half a day at a time. The wind passed through the sparse branches, rustling, whispering, like voices from distant years, or an unnamed lullaby. He seemed to hear the creak of the wicker chair, hear "watch your step" called out gently, hear a hoarse voice saying "go on out into the world," and then hear a childish voice asking: "Grandpa, does the tree hurt?" He no longer fished through the past, nor measured what remained of his life; the lake of his heart no longer rippled. On a certain afternoon deep in spring, he slowly closed his eyes, the corners of his mouth curving into the faintest smile. The wind suddenly stilled, and a single locust blossom spiraled down from above, landing precisely upon his lap. His lips moved faintly, as though breathing out the words: "...It doesn't hurt anymore." Pure white, weightless-just as it had appeared, all those years ago, the very first time he had stood on tiptoe and looked up.
\end{storyquote}

\section{Additional Quantitative and Qualitative Results}

This appendix supplements Section 6. We expand the aggregated scores into a per-novel breakdown (\ref{sec:b1}), show that FilmWorld is not tied to a single visual style (\ref{sec:b2}), examine the duration of the generated films to characterize the scale of output FilmWorld produces (\ref{sec:b3}), and conduct a cross-evaluator analysis on FilmEval to verify that its rankings are stable across different judge models (\ref{sec:b4}).

\subsection{Per-Novel Full Score Tables}
\label{sec:b1}

Table~\ref{tab:pernovel} reports the full score matrix of all 6 methods over the 15 novels and 9 sub-metrics, exposing the per-novel distribution and supplying the coordinates for the covariance-ellipse analysis. Novels use the canonical IDs (C1--C15) of Appendix~\ref{sec:a1}, ordered by difficulty tier (Easy C1--C5, Medium C6--C10, Hard C11--C15); the sub-metrics follow the main-paper grouping into Cinematic Presentation (VP, NEP, AVP), Film Consistency (CC, SC, OC), and Novel Fidelity (NHR, LR, SR).

Two main-paper findings are visible at per-novel granularity. First, FilmWorld's overall scores stay tightly concentrated (roughly 87--91) across all 15 novels, whereas every baseline spreads far wider with pronounced low outliers on the harder novels. Second, FilmWorld's largest gains lie in Novel Fidelity (NHR, LR, SR) and do not collapse on the Hard tier, confirming that grounding generation in an explicit symbolic state trajectory suppresses narrative omission and hallucination as complexity grows.

\subsection{Style Generalization Study}
\label{sec:b2}

The main experiments uniformly adopt a Studio Ghibli anime style across all methods, purely to eliminate stylistic discrepancy as a confounding factor in evaluation. This choice does not imply any dependence of FilmWorld on a particular visual style. Because style enters the pipeline only through a short natural-language descriptor injected into the image generator, and never interacts with the symbolic world-state trajectory, FilmWorld is agnostic to the rendered look. To make this concrete, this section lists five style prompts spanning animation and photorealistic looks, and shows the same shot rendered under each.

Each style is defined by a compact descriptor appended to the tail of the image generator's system prompt. The descriptor conditions only the appearance of the rendered pixels; the entity identities, spatial composition, and end-state metadata carried by the shot directive are left untouched. The five style prompts are:

\textbf{Ghibli style.}

\begin{lstlisting}
Studio Ghibli animation style, soft watercolor textures, hand-drawn cel-shading, warm pastel color palette, gentle natural lighting, painterly backgrounds.
\end{lstlisting}

\textbf{Shinkai style.}

\begin{lstlisting}
Makoto Shinkai anime style, Japanese anime aesthetics, crystal clear and transparent colors, soft golden sunlight, fresh and airy atmosphere, luminous sky and clouds, delicate light and shadow details.
\end{lstlisting}

\textbf{Inuyasha style.}

\begin{lstlisting}
Inuyasha (anime series) style, in the art style of Rumiko Takahashi, late 90s and early 2000s retro anime aesthetic, traditional cel-shaded animation, bold and clean outlines, simple, hard-edged shadows, a color palette of earthy tones mixed with bold primary colors, painterly backgrounds depicting feudal Japan, mystical forests, and ancient temples, a hint of film grain and analog texture for a nostalgic feel.
\end{lstlisting}

\textbf{Disney 3D style.}

\begin{lstlisting}
Disney 3D animation style, Pixar-style rendering, vibrant and saturated colors, smooth and glossy surfaces, expressive big eyes, soft global illumination, warm and whimsical atmosphere, rounded character design, cinematic depth of field, high-quality subsurface scattering on skin.
\end{lstlisting}

\textbf{Photorealistic style.}

\begin{lstlisting}
Cinematic photorealistic style, authentic film-like quality, natural film grain, soft natural lighting, true-to-life color without oversaturation, matte surface texture, avoiding plastic sheen and waxy look, natural organic depth of field, retaining authentic imperfections and detail layers, cinema-grade color grading (Kodak/Fuji film color science aesthetic).
\end{lstlisting}

\begin{figure}[!t]  
  \centering  
  \includegraphics[width=\textwidth]{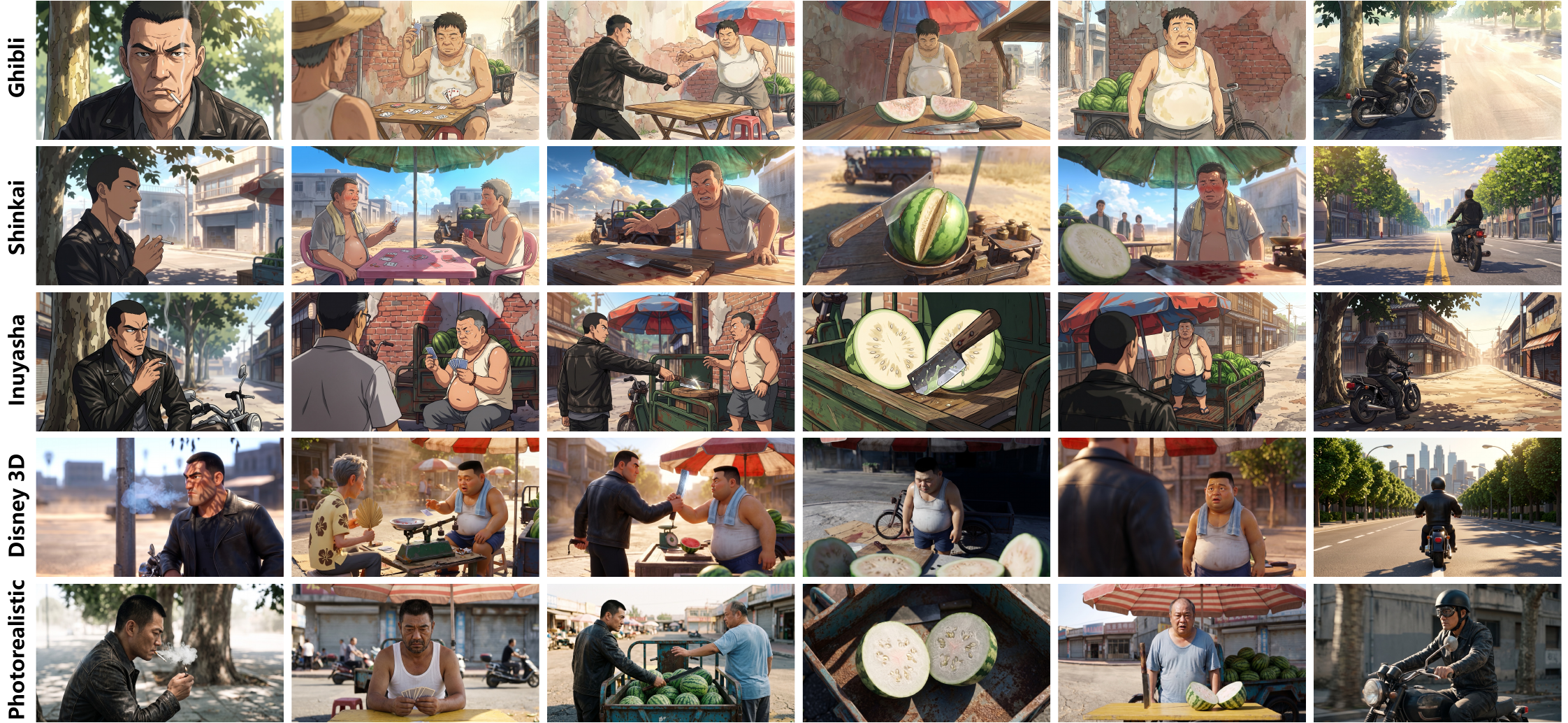}  
  \caption{\textbf{Style generalization across five style prompts.} Each column is a representative keyframe from \textit{Huaqiang Buying Watermelons} (C4), rendered independently under each style prompt (rows). Each style descriptor faithfully imposes its intended visual look, while entity identity and narrative content remain consistent across styles, confirming that style is effectively controlled and decoupled from the underlying world-state trajectory.}
  \vspace{-3mm}
  \label{fig:style}  
\end{figure}

Figure~\ref{fig:style} shows the effect of varying only the style descriptor while keeping the input narrative fixed. Rather than restricting to a single shot, we select several representative keyframe positions drawn from across the fully rendered film of \textit{Huaqiang Buying Watermelons} (C4), spanning different scenes and character configurations. Because the shot directive is itself automatically derived by the system rather than manually specified, minor stochastic variation in the parsed entities, composition, or pose may arise across independent runs; nonetheless, the overall narrative content, participating entities, and spatial arrangement remain highly consistent. This section is illustrative only: it demonstrates cross-style transferability and makes no claim about the relative quality of the styles, which would confound the controlled evaluation of the main paper.

\begin{table}[H]
\centering
\scriptsize
\setlength{\tabcolsep}{4pt}
\renewcommand{\arraystretch}{0.9}
\caption{\textbf{Per-novel scores of all 6 methods across the 15 novels and 9 sub-metrics.} The results are grouped into Cinematic Presentation, Film Consistency, and Novel Fidelity. Within each subtable the shaded rows report the per-tier means (Easy/Medium/Hard) and the method-level overall; the Overall column is the mean of the nine sub-metrics.}
\label{tab:pernovel}

\begin{subtable}{\textwidth}
\centering
\caption{MM-StoryAgent}
\label{tab:pernovel_mmstory}
\begin{tabular}{lrrrrrrrrrr}
\pernovelhead
C1 & 83.00 & 82.00 & 80.00 & 81.80 & 86.22 & 96.73 & 96.71 & 93.00 & 85.00 & 87.16 \\
C2 & 84.00 & 84.00 & 84.00 & 81.77 & 87.24 & 97.82 & 90.88 & 88.00 & 86.00 & 87.08 \\
C3 & 82.00 & 82.00 & 81.00 & 65.37 & 83.72 & 97.56 & 65.54 & 78.00 & 57.00 & 76.91 \\
C4 & 82.00 & 76.00 & 80.00 & 51.86 & 80.54 & 93.67 & 68.35 & 77.00 & 55.00 & 73.82 \\
C5 & 82.00 & 81.00 & 80.00 & 77.96 & 87.79 & 95.99 & 93.83 & 86.00 & 57.00 & 82.40 \\
\rowcolor{tieravg}\textbf{Easy avg} & \textbf{82.60} & \textbf{81.00} & \textbf{81.00} & \textbf{71.75} & \textbf{85.10} & \textbf{96.35} & \textbf{83.06} & \textbf{84.40} & \textbf{68.00} & \textbf{81.47} \\
\midrule
C6 & 73.00 & 74.00 & 77.00 & 43.36 & 87.12 & 95.47 & 97.22 & 93.00 & 85.00 & 80.57 \\
C7 & 78.00 & 78.00 & 84.00 & 77.27 & 89.55 & 93.72 & 98.19 & 91.00 & 79.00 & 85.41 \\
C8 & 79.00 & 79.00 & 77.00 & 76.15 & 86.74 & 96.61 & 91.04 & 80.00 & 55.00 & 80.06 \\
C9 & 80.00 & 75.00 & 81.00 & 47.36 & 83.49 & 87.82 & 29.83 & 71.00 & 19.00 & 63.83 \\
C10 & 76.00 & 75.00 & 76.00 & 72.21 & 85.38 & 96.08 & 87.27 & 82.00 & 58.00 & 78.66 \\
\rowcolor{tieravg}\textbf{Medium avg} & \textbf{77.20} & \textbf{76.20} & \textbf{79.00} & \textbf{63.27} & \textbf{86.46} & \textbf{93.94} & \textbf{80.71} & \textbf{83.40} & \textbf{59.20} & \textbf{77.71} \\
\midrule
C11 & 83.00 & 81.00 & 82.00 & 76.48 & 92.20 & 98.87 & 99.29 & 89.00 & 82.00 & 87.09 \\
C12 & 82.00 & 83.00 & 85.00 & 79.08 & 87.23 & 98.07 & 99.37 & 83.00 & 69.00 & 85.08 \\
C13 & 83.00 & 81.00 & 86.00 & 79.56 & 88.87 & 96.50 & 98.60 & 77.00 & 48.00 & 82.06 \\
C14 & 81.00 & 80.00 & 75.00 & 75.02 & 91.15 & 100.00 & 83.28 & 81.00 & 62.00 & 80.94 \\
C15 & 74.00 & 76.00 & 75.00 & 69.52 & 88.03 & 94.95 & 91.71 & 79.00 & 41.00 & 76.58 \\
\rowcolor{tieravg}\textbf{Hard avg} & \textbf{80.60} & \textbf{80.20} & \textbf{80.60} & \textbf{75.93} & \textbf{89.50} & \textbf{97.68} & \textbf{94.45} & \textbf{81.80} & \textbf{60.40} & \textbf{82.35} \\
\midrule
\rowcolor{grandavg}\textbf{Overall} & \textbf{80.13} & \textbf{79.13} & \textbf{80.20} & \textbf{70.32} & \textbf{87.02} & \textbf{95.99} & \textbf{86.07} & \textbf{83.20} & \textbf{62.53} & \textbf{80.51} \\
\bottomrule
\end{tabular}
\end{subtable}

\vspace{1ex}

\begin{subtable}{\textwidth}
\centering
\caption{VGoT}
\label{tab:pernovel_vgot}
\begin{tabular}{lrrrrrrrrrr}
\pernovelhead
C1 & 86.00 & 82.00 & 79.00 & 85.37 & 80.51 & 94.05 & 78.39 & 79.00 & 64.00 & 80.92 \\
C2 & 88.00 & 83.00 & 77.00 & 79.86 & 87.65 & 93.64 & 80.33 & 85.00 & 65.00 & 82.16 \\
C3 & 84.00 & 85.00 & 83.00 & 80.81 & 91.06 & 96.43 & 89.27 & 89.00 & 70.00 & 85.40 \\
C4 & 85.00 & 86.00 & 81.00 & 75.00 & 88.91 & 90.30 & 93.31 & 81.00 & 68.00 & 83.17 \\
C5 & 80.00 & 84.00 & 86.00 & 80.44 & 86.05 & 93.94 & 59.93 & 90.00 & 60.00 & 80.04 \\
\rowcolor{tieravg}\textbf{Easy avg} & \textbf{84.60} & \textbf{84.00} & \textbf{81.20} & \textbf{80.30} & \textbf{86.84} & \textbf{93.67} & \textbf{80.25} & \textbf{84.80} & \textbf{65.40} & \textbf{82.34} \\
\midrule
C6 & 74.00 & 77.00 & 74.00 & 72.44 & 81.27 & 91.56 & 94.78 & 87.00 & 75.00 & 80.78 \\
C7 & 78.00 & 82.00 & 78.00 & 78.73 & 94.36 & 94.23 & 94.83 & 89.00 & 63.00 & 83.57 \\
C8 & 83.00 & 87.00 & 85.00 & 81.00 & 90.27 & 93.39 & 95.97 & 87.00 & 69.00 & 85.74 \\
C9 & 85.00 & 86.00 & 72.00 & 83.49 & 88.07 & 98.09 & 93.23 & 86.00 & 68.00 & 84.43 \\
C10 & 79.00 & 82.00 & 83.00 & 79.81 & 89.09 & 100.00 & 77.68 & 85.00 & 70.00 & 82.84 \\
\rowcolor{tieravg}\textbf{Medium avg} & \textbf{79.80} & \textbf{82.80} & \textbf{78.40} & \textbf{79.09} & \textbf{88.61} & \textbf{95.45} & \textbf{91.30} & \textbf{86.80} & \textbf{69.00} & \textbf{83.47} \\
\midrule
C11 & 74.00 & 75.00 & 76.00 & 75.90 & 87.00 & 98.28 & 82.74 & 80.00 & 57.00 & 78.44 \\
C12 & 74.00 & 80.00 & 78.00 & 78.92 & 92.72 & 97.12 & 98.43 & 75.00 & 52.00 & 80.69 \\
C13 & 83.00 & 84.00 & 84.00 & 80.27 & 87.86 & 98.19 & 93.40 & 83.00 & 47.00 & 82.30 \\
C14 & 81.00 & 82.00 & 81.00 & 84.32 & 88.89 & 98.95 & 84.29 & 77.00 & 51.00 & 80.94 \\
C15 & 82.00 & 82.00 & 84.00 & 74.43 & 84.50 & 92.53 & 87.27 & 79.00 & 44.00 & 78.86 \\
\rowcolor{tieravg}\textbf{Hard avg} & \textbf{78.80} & \textbf{80.60} & \textbf{80.60} & \textbf{78.77} & \textbf{88.19} & \textbf{97.01} & \textbf{89.23} & \textbf{78.80} & \textbf{50.20} & \textbf{80.25} \\
\midrule
\rowcolor{grandavg}\textbf{Overall} & \textbf{81.07} & \textbf{82.47} & \textbf{80.07} & \textbf{79.39} & \textbf{87.88} & \textbf{95.38} & \textbf{86.92} & \textbf{83.47} & \textbf{61.53} & \textbf{82.02} \\
\bottomrule
\end{tabular}
\end{subtable}

\vspace{1ex}

\begin{subtable}{\textwidth}
\centering
\caption{MovieAgent}
\label{tab:pernovel_movieagent}
\begin{tabular}{lrrrrrrrrrr}
\pernovelhead
C1 & 82.00 & 82.00 & 77.00 & 79.62 & 85.00 & 96.90 & 87.14 & 97.00 & 75.00 & 84.63 \\
C2 & 82.00 & 86.00 & 75.00 & 83.23 & 92.44 & 96.16 & 84.30 & 92.00 & 71.00 & 84.68 \\
C3 & 72.00 & 79.00 & 81.00 & 76.24 & 86.74 & 95.21 & 80.26 & 95.00 & 85.00 & 83.38 \\
C4 & 67.00 & 64.00 & 75.00 & 43.33 & 77.23 & 92.20 & 87.51 & 92.00 & 76.00 & 74.92 \\
C5 & 77.00 & 81.00 & 79.00 & 79.26 & 86.37 & 94.20 & 89.96 & 93.00 & 83.00 & 84.75 \\
\rowcolor{tieravg}\textbf{Easy avg} & \textbf{76.00} & \textbf{78.40} & \textbf{77.40} & \textbf{72.34} & \textbf{85.56} & \textbf{94.93} & \textbf{85.83} & \textbf{93.80} & \textbf{78.00} & \textbf{82.47} \\
\midrule
C6 & 75.00 & 80.00 & 74.00 & 84.81 & 89.21 & 94.88 & 91.70 & 97.00 & 85.00 & 85.73 \\
C7 & 80.00 & 78.00 & 74.00 & 74.90 & 86.72 & 98.27 & 95.09 & 97.00 & 79.00 & 84.78 \\
C8 & 82.00 & 85.00 & 79.00 & 72.46 & 88.79 & 97.91 & 86.14 & 90.00 & 77.00 & 84.26 \\
C9 & 83.00 & 82.00 & 77.00 & 81.11 & 91.56 & 98.24 & 91.20 & 93.00 & 79.00 & 86.23 \\
C10 & 80.00 & 79.00 & 76.00 & 69.22 & 81.40 & 96.50 & 87.15 & 81.00 & 70.00 & 80.03 \\
\rowcolor{tieravg}\textbf{Medium avg} & \textbf{80.00} & \textbf{80.80} & \textbf{76.00} & \textbf{76.50} & \textbf{87.54} & \textbf{97.16} & \textbf{90.26} & \textbf{91.60} & \textbf{78.00} & \textbf{84.21} \\
\midrule
C11 & 82.00 & 85.00 & 78.00 & 81.69 & 92.15 & 98.81 & 91.04 & 97.00 & 76.00 & 86.85 \\
C12 & 82.00 & 85.00 & 76.00 & 80.68 & 91.22 & 98.42 & 95.33 & 97.00 & 75.00 & 86.74 \\
C13 & 80.00 & 84.00 & 79.00 & 71.51 & 90.08 & 99.45 & 87.67 & 93.00 & 62.00 & 82.97 \\
C14 & 84.00 & 85.00 & 80.00 & 83.72 & 92.32 & 97.38 & 89.71 & 95.00 & 78.00 & 87.24 \\
C15 & 79.00 & 82.00 & 85.00 & 78.51 & 86.83 & 98.75 & 86.64 & 91.00 & 55.00 & 82.53 \\
\rowcolor{tieravg}\textbf{Hard avg} & \textbf{81.40} & \textbf{84.20} & \textbf{79.60} & \textbf{79.22} & \textbf{90.52} & \textbf{98.56} & \textbf{90.08} & \textbf{94.60} & \textbf{69.20} & \textbf{85.27} \\
\midrule
\rowcolor{grandavg}\textbf{Overall} & \textbf{79.13} & \textbf{81.13} & \textbf{77.67} & \textbf{76.02} & \textbf{87.87} & \textbf{96.89} & \textbf{88.72} & \textbf{93.33} & \textbf{75.07} & \textbf{83.98} \\
\bottomrule
\end{tabular}
\end{subtable}
\end{table}

\begin{table}[H]\ContinuedFloat
\centering
\scriptsize
\setlength{\tabcolsep}{4pt}
\renewcommand{\arraystretch}{0.9}
\caption{(continued) Per-novel scores of all 6 methods across the 15 novels and 9 sub-metrics.}

\begin{subtable}{\textwidth}
\centering
\caption{ViMax}
\label{tab:pernovel_vimax}
\begin{tabular}{lrrrrrrrrrr}
\pernovelhead
C1 & 82.00 & 78.00 & 85.00 & 88.14 & 81.97 & 97.96 & 75.74 & 72.00 & 56.00 & 79.65 \\
C2 & 83.00 & 79.00 & 84.00 & 83.70 & 86.28 & 86.56 & 91.95 & 78.00 & 64.00 & 81.83 \\
C3 & 86.00 & 87.00 & 87.00 & 84.00 & 89.33 & 91.33 & 97.75 & 80.00 & 59.00 & 84.60 \\
C4 & 72.00 & 80.00 & 82.00 & 78.68 & 91.54 & 97.96 & 77.55 & 88.00 & 72.00 & 82.19 \\
C5 & 86.00 & 88.00 & 88.00 & 75.59 & 85.98 & 95.95 & 95.27 & 86.00 & 84.00 & 87.20 \\
\rowcolor{tieravg}\textbf{Easy avg} & \textbf{81.80} & \textbf{82.40} & \textbf{85.20} & \textbf{82.02} & \textbf{87.02} & \textbf{93.95} & \textbf{87.65} & \textbf{80.80} & \textbf{67.00} & \textbf{83.09} \\
\midrule
C6 & 66.00 & 74.00 & 65.00 & 84.51 & 83.62 & 94.50 & 93.32 & 81.00 & 68.00 & 78.88 \\
C7 & 79.00 & 83.00 & 83.00 & 77.20 & 89.43 & 87.50 & 98.98 & 71.00 & 56.00 & 80.57 \\
C8 & 73.00 & 80.00 & 80.00 & 70.47 & 83.57 & 94.91 & 91.21 & 76.00 & 69.00 & 79.80 \\
C9 & 84.00 & 86.00 & 85.00 & 83.50 & 86.73 & 94.79 & 98.15 & 81.00 & 69.00 & 85.35 \\
C10 & 67.00 & 69.00 & 70.00 & 75.59 & 87.06 & 95.78 & 92.50 & 78.00 & 56.00 & 76.77 \\
\rowcolor{tieravg}\textbf{Medium avg} & \textbf{73.80} & \textbf{78.40} & \textbf{76.60} & \textbf{78.25} & \textbf{86.08} & \textbf{93.50} & \textbf{94.83} & \textbf{77.40} & \textbf{63.60} & \textbf{80.27} \\
\midrule
C11 & 86.00 & 87.00 & 87.00 & 80.32 & 87.73 & 95.00 & 98.62 & 71.00 & 55.00 & 83.07 \\
C12 & 67.00 & 77.00 & 77.00 & 76.00 & 92.60 & 86.27 & 97.83 & 68.00 & 46.00 & 76.41 \\
C13 & 66.00 & 78.00 & 84.00 & 79.40 & 87.77 & 94.47 & 77.78 & 83.00 & 56.00 & 78.49 \\
C14 & 82.00 & 81.00 & 79.00 & 84.35 & 86.42 & 96.96 & 90.15 & 78.00 & 63.00 & 82.32 \\
C15 & 73.00 & 78.00 & 79.00 & 79.48 & 89.51 & 97.72 & 86.92 & 83.00 & 49.00 & 79.51 \\
\rowcolor{tieravg}\textbf{Hard avg} & \textbf{74.80} & \textbf{80.20} & \textbf{81.20} & \textbf{79.91} & \textbf{88.81} & \textbf{94.08} & \textbf{90.26} & \textbf{76.60} & \textbf{53.80} & \textbf{79.96} \\
\midrule
\rowcolor{grandavg}\textbf{Overall} & \textbf{76.80} & \textbf{80.33} & \textbf{81.00} & \textbf{80.06} & \textbf{87.30} & \textbf{93.84} & \textbf{90.91} & \textbf{78.27} & \textbf{61.47} & \textbf{81.11} \\
\bottomrule
\end{tabular}
\end{subtable}

\vspace{1ex}

\begin{subtable}{\textwidth}
\centering
\caption{VideoClaw}
\label{tab:pernovel_videoclaw}
\begin{tabular}{lrrrrrrrrrr}
\pernovelhead
C1 & 77.00 & 79.00 & 78.00 & 83.53 & 91.11 & 91.40 & 84.75 & 81.00 & 64.00 & 81.09 \\
C2 & 87.00 & 88.00 & 87.00 & 51.68 & 99.23 & 95.57 & 78.45 & 78.00 & 53.00 & 79.77 \\
C3 & 82.00 & 83.00 & 83.00 & 80.18 & 94.82 & 98.17 & 92.33 & 99.00 & 90.00 & 89.17 \\
C4 & 77.00 & 81.00 & 79.00 & 76.15 & 95.95 & 94.74 & 92.94 & 94.00 & 79.00 & 85.53 \\
C5 & 82.00 & 85.00 & 86.00 & 49.86 & 92.80 & 97.01 & 91.90 & 97.00 & 88.00 & 85.51 \\
\rowcolor{tieravg}\textbf{Easy avg} & \textbf{81.00} & \textbf{83.20} & \textbf{82.60} & \textbf{68.28} & \textbf{94.78} & \textbf{95.38} & \textbf{88.07} & \textbf{89.80} & \textbf{74.80} & \textbf{84.21} \\
\midrule
C6 & 75.00 & 69.00 & 70.00 & 71.33 & 78.71 & 88.42 & 81.67 & 68.00 & 52.00 & 72.68 \\
C7 & 83.00 & 82.00 & 86.00 & 74.21 & 97.95 & 96.58 & 89.18 & 86.00 & 75.00 & 85.55 \\
C8 & 85.00 & 88.00 & 85.00 & 80.05 & 94.98 & 98.47 & 94.28 & 95.00 & 82.00 & 89.20 \\
C9 & 85.00 & 86.00 & 87.00 & 82.07 & 93.73 & 94.35 & 93.00 & 79.00 & 56.00 & 84.02 \\
C10 & 79.00 & 82.00 & 83.00 & 78.97 & 96.80 & 98.53 & 95.09 & 97.00 & 89.00 & 88.82 \\
\rowcolor{tieravg}\textbf{Medium avg} & \textbf{81.40} & \textbf{81.40} & \textbf{82.20} & \textbf{77.33} & \textbf{92.43} & \textbf{95.27} & \textbf{90.64} & \textbf{85.00} & \textbf{70.80} & \textbf{84.05} \\
\midrule
C11 & 64.00 & 72.00 & 81.00 & 79.60 & 92.80 & 97.23 & 97.76 & 80.00 & 56.00 & 80.04 \\
C12 & 76.00 & 84.00 & 86.00 & 73.90 & 90.86 & 83.43 & 98.56 & 76.00 & 40.00 & 78.75 \\
C13 & 83.00 & 88.00 & 88.00 & 85.38 & 98.81 & 98.96 & 96.90 & 99.00 & 81.00 & 91.01 \\
C14 & 81.00 & 83.00 & 81.00 & 82.44 & 95.72 & 99.88 & 88.83 & 99.00 & 85.00 & 88.43 \\
C15 & 80.00 & 80.00 & 81.00 & 49.53 & 94.39 & 98.88 & 87.17 & 96.00 & 83.00 & 83.33 \\
\rowcolor{tieravg}\textbf{Hard avg} & \textbf{76.80} & \textbf{81.40} & \textbf{83.40} & \textbf{74.17} & \textbf{94.52} & \textbf{95.68} & \textbf{93.84} & \textbf{90.00} & \textbf{69.00} & \textbf{84.31} \\
\midrule
\rowcolor{grandavg}\textbf{Overall} & \textbf{79.73} & \textbf{82.00} & \textbf{82.73} & \textbf{73.26} & \textbf{93.91} & \textbf{95.44} & \textbf{90.85} & \textbf{88.27} & \textbf{71.53} & \textbf{84.19} \\
\bottomrule
\end{tabular}
\end{subtable}

\vspace{1ex}

\begin{subtable}{\textwidth}
\centering
\caption{FilmWorld (Ours)}
\label{tab:pernovel_ours}
\begin{tabular}{lrrrrrrrrrr}
\pernovelhead
C1 & 80.00 & 86.00 & 85.00 & 85.28 & 90.53 & 94.15 & 93.07 & 94.00 & 75.00 & 87.00 \\
C2 & 82.00 & 87.00 & 85.00 & 84.47 & 94.38 & 98.06 & 97.69 & 98.00 & 86.00 & 90.29 \\
C3 & 84.00 & 86.00 & 82.00 & 83.25 & 92.95 & 98.65 & 99.48 & 97.00 & 89.00 & 90.26 \\
C4 & 78.00 & 82.00 & 84.00 & 78.33 & 91.63 & 100.00 & 98.06 & 99.00 & 87.00 & 88.67 \\
C5 & 85.00 & 88.00 & 86.00 & 83.19 & 93.73 & 95.26 & 96.53 & 97.00 & 83.00 & 89.75 \\
\rowcolor{tieravg}\textbf{Easy avg} & \textbf{81.80} & \textbf{85.80} & \textbf{84.40} & \textbf{82.90} & \textbf{92.64} & \textbf{97.22} & \textbf{96.97} & \textbf{97.00} & \textbf{84.00} & \textbf{89.19} \\
\midrule
C6 & 73.00 & 79.00 & 77.00 & 81.47 & 93.93 & 95.61 & 91.49 & 100.00 & 91.00 & 86.94 \\
C7 & 80.00 & 84.00 & 85.00 & 80.78 & 94.76 & 95.55 & 95.75 & 100.00 & 87.00 & 89.20 \\
C8 & 85.00 & 88.00 & 86.00 & 82.84 & 94.29 & 99.44 & 97.55 & 100.00 & 88.00 & 91.24 \\
C9 & 84.00 & 88.00 & 85.00 & 84.72 & 94.94 & 97.04 & 92.39 & 97.00 & 83.00 & 89.57 \\
C10 & 79.00 & 81.00 & 83.00 & 80.92 & 98.31 & 98.81 & 90.48 & 99.00 & 81.00 & 87.95 \\
\rowcolor{tieravg}\textbf{Medium avg} & \textbf{80.20} & \textbf{84.00} & \textbf{83.20} & \textbf{82.15} & \textbf{95.25} & \textbf{97.29} & \textbf{93.53} & \textbf{99.20} & \textbf{86.00} & \textbf{88.98} \\
\midrule
C11 & 78.00 & 80.00 & 80.00 & 84.69 & 95.65 & 98.53 & 98.61 & 99.00 & 85.00 & 88.83 \\
C12 & 81.00 & 86.00 & 84.00 & 81.74 & 96.10 & 98.08 & 98.85 & 99.00 & 91.00 & 90.64 \\
C13 & 80.00 & 83.00 & 83.00 & 85.03 & 95.94 & 99.29 & 97.92 & 98.00 & 83.00 & 89.46 \\
C14 & 83.00 & 86.00 & 85.00 & 85.19 & 96.79 & 98.06 & 95.85 & 100.00 & 85.00 & 90.54 \\
C15 & 83.00 & 85.00 & 86.00 & 83.88 & 95.74 & 99.55 & 95.11 & 98.00 & 87.00 & 90.36 \\
\rowcolor{tieravg}\textbf{Hard avg} & \textbf{81.00} & \textbf{84.00} & \textbf{83.60} & \textbf{84.11} & \textbf{96.04} & \textbf{98.70} & \textbf{97.27} & \textbf{98.80} & \textbf{86.20} & \textbf{89.97} \\
\midrule
\rowcolor{grandavg}\textbf{Overall} & \textbf{81.00} & \textbf{84.60} & \textbf{83.73} & \textbf{83.05} & \textbf{94.64} & \textbf{97.74} & \textbf{95.92} & \textbf{98.33} & \textbf{85.40} & \textbf{89.38} \\
\bottomrule
\end{tabular}
\end{subtable}

\end{table}

\begin{figure}[htb]  
  \centering  
  \includegraphics[width=0.64\textwidth]{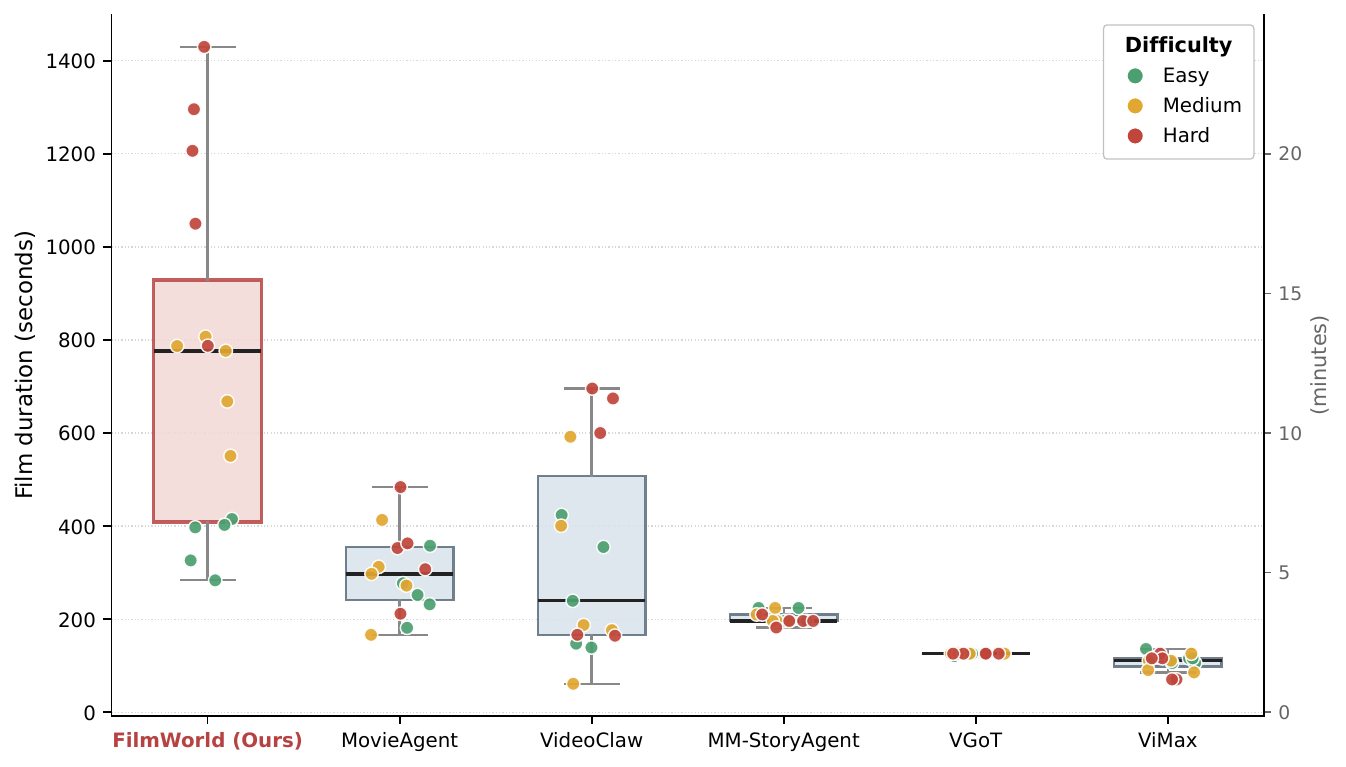} 
  \vspace{-2mm}
  \caption{\textbf{Distribution of generated video durations across the six systems.} Each box summarizes the durations of the 15 films produced by a system (one per FilmEval novel), with individual points colored by difficulty tier (Easy/Medium/Hard); systems are ordered by median duration. FilmWorld yields both the longest films (median $12.9$\,min) and the widest spread, with duration scaling markedly from Easy to Hard.}
\label{fig:duration_boxplot}
\end{figure}

\begin{table}[htb]
\centering
\caption{\textbf{Generated video duration (\texttt{mm:ss}) across the six systems on the 15 FilmEval novels.} Each cell is the actual duration of the film produced by a given system for a given novel; rows are grouped by difficulty tier. The rightmost column reports the source length (English words for C1--C2, C6--C7, C11--C12; Chinese characters otherwise). \textbf{FilmWorld} scales its output length with narrative complexity, whereas fixed-pipeline baselines emit near-constant durations.}
\label{tab:filmeval_duration}
\renewcommand{\arraystretch}{1.15}
\setlength{\tabcolsep}{5.2pt}
\resizebox{0.58\textwidth}{!}{%
\begin{tabular}{lcccccc c}
\toprule
\rowcolor{headercolor}
ID & \textbf{FilmWorld} & VideoClaw & MM-StoryAgent & MovieAgent & VGoT & ViMax & Length \\
\midrule
\rowcolor{easycolor}
C1  & \textbf{4:43}  & 2:19  & 3:44 & 4:37 & 2:06 & 1:56 & 1{,}015 \\
\rowcolor{easycolor}
C2  & \textbf{6:55}  & 2:27  & 3:44 & 5:58 & 2:06 & 1:46 & 1{,}149 \\
\rowcolor{easycolor}
C3  & \textbf{6:43}  & 7:04  & 3:16 & 4:12 & 2:06 & 1:56 & 1{,}633 \\
\rowcolor{easycolor}
C4  & \textbf{5:26}  & 3:59  & 3:16 & 3:01 & 2:01 & 1:46 & 1{,}492 \\
\rowcolor{easycolor}
C5  & \textbf{6:37}  & 5:55  & 3:16 & 3:52 & 2:06 & 2:16 & 1{,}614 \\

\rowcolor{medcolor}
C6  & \textbf{9:11}  & 1:01  & 3:44 & 6:53 & 2:06 & 1:51 & 1{,}348 \\
\rowcolor{medcolor}
C7  & \textbf{13:07} & 3:07  & 3:16 & 5:12 & 2:06 & 1:26 & 2{,}067 \\
\rowcolor{medcolor}
C8  & \textbf{11:08} & 6:41  & 3:30 & 4:32 & 2:06 & 1:51 & 2{,}214 \\
\rowcolor{medcolor}
C9  & \textbf{12:57} & 2:56  & 3:16 & 4:57 & 2:06 & 1:31 & 2{,}996 \\
\rowcolor{medcolor}
C10 & \textbf{13:27} & 9:52  & 3:16 & 2:46 & 2:06 & 2:06 & 2{,}907 \\

\rowcolor{hardcolor}
C11 & \textbf{17:30} & 2:46  & 3:30 & 8:04 & 2:06 & 1:10 & 2{,}334 \\
\rowcolor{hardcolor}
C12 & \textbf{21:36} & 2:44  & 3:16 & 5:53 & 2:06 & 1:10 & 2{,}733 \\
\rowcolor{hardcolor}
C13 & \textbf{20:06} & 11:36 & 3:02 & 5:07 & 2:06 & 2:06 & 4{,}829 \\
\rowcolor{hardcolor}
C14 & \textbf{23:50} & 11:14 & 3:16 & 6:03 & 2:06 & 1:56 & 4{,}334 \\
\rowcolor{hardcolor}
C15 & \textbf{13:07} & 10:00 & 3:16 & 3:32 & 2:06 & 1:56 & 4{,}663 \\
\bottomrule
\end{tabular}
}
\end{table}

\subsection{On the Duration of Generated Films}
\label{sec:b3}
Duration serves as a diagnostic signal, not a metric. Faithfully adapting a novel requires enough screen time to stage its scenes, characters, and events; a system that compresses a 4{,}800-character story into two minutes has necessarily elided most of them. We thus read generated duration as a physical trace of narrative coverage, that is, a lower bound on how much of the story could have been rendered, rather than as a quality metric, since a longer film is not intrinsically better. Interpreted alongside the fidelity and consistency scores, it diagnoses whether a system engages with the full narrative or merely samples a fragment. Figure~\ref{fig:duration_boxplot} and Table~\ref{tab:filmeval_duration} report per-film durations for all six systems across the 15 novels.

FilmWorld scales duration with narrative complexity. The six systems fall into two regimes. Fixed-pipeline baselines emit near-constant durations that are decoupled from the source: VGoT collapses to $\sim 2.1$ minutes and MM-StoryAgent to $\sim 3.3$ minutes on every novel, regardless of whether the input is a 1{,}015-word vignette or a 4{,}829-character novella. FilmWorld instead exhibits both the longest films (median $12.9$ minutes) and the widest spread, growing from $4$--$7$ minutes on Easy sources to as much as $17$--$24$ minutes on most Hard ones. Importantly, this trend reflects content volume rather than difficulty per se: our difficulty tiers are defined by narrative-structural complexity, which need not translate into longer screen time. The one exception is C15, a Hard-tier story whose FilmWorld rendition runs only $13.1$ minutes, shorter than its Hard peers, because its complexity stems largely from dense, fast-paced dialogue, which is verbally rich yet visually compact and thus occupies little screen time. This length-to-content scaling is precisely the behavior expected of genuine narrative coverage: because our construction phase materializes the full set of world entities and plot events before rendering, downstream shot planning expands to accommodate the content a story actually requires rather than truncating it to a fixed budget. The baselines' flat profiles, by contrast, expose a structural ceiling on how much narrative they can physically represent.

\begin{figure}[htb]
\centering
\includegraphics[width=0.98\textwidth]{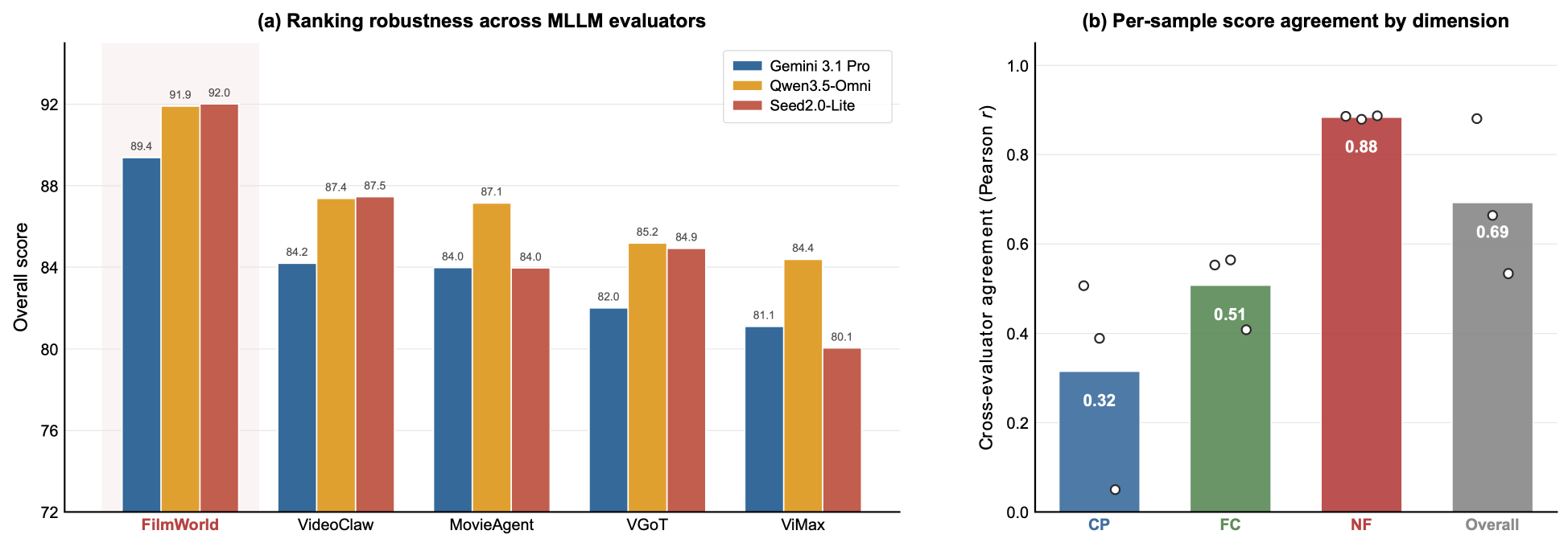}
\vspace{-3mm}
\caption{\textbf{Cross-evaluator analysis on FilmEval.} We re-run the automatic protocol with three distinct MLLM judges (Gemini~3.1~Pro, Qwen3.5-Omni, and Seed2.0-Lite) on the five systems. (a)~Overall score per method under each judge: FilmWorld ranks first under all three, and the induced rankings are near-identical (Spearman $\rho\!\geq\!0.9$). (b)~Mean pairwise per-sample agreement by dimension, with the three pairwise values overlaid as dots: correlation increases from Cinematic Presentation ($r\!=\!0.32$) to Film Consistency ($r\!=\!0.51$) to Novel Fidelity ($r\!=\!0.88$).}
\vspace{-3mm}
\label{fig:cross_evaluator}
\end{figure}

\begin{figure}[htb]
\centering
\includegraphics[width=0.98\textwidth]{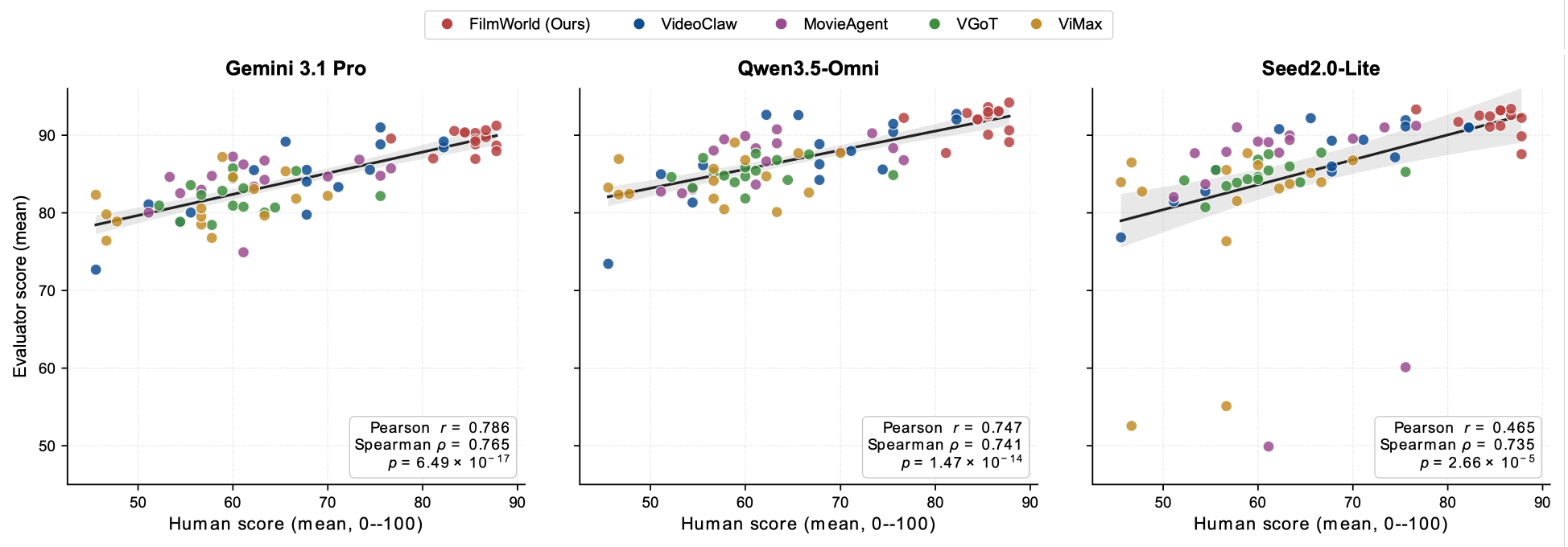}
\vspace{-3mm}
\caption{\textbf{Alignment between each MLLM evaluator and human ratings.} For each of the three judges, we plot the per-film overall score (Y) against the mean human rating (X, $0$--$5$ scaled to $0$--$100$) over the 75 films, colored by system. The solid line is a linear fit with its $95\%$ confidence band; Pearson $r$, Spearman $\rho$, and the two-sided $p$-value are annotated. All three evaluators correlate significantly and positively with human judgment.}
\vspace{-3mm}
\label{fig:evaluator_human}
\end{figure}

\subsection{Cross-Evaluator Analysis on FilmEval}
\label{sec:b4}

A natural concern for any MLLM-based protocol is whether its verdicts hinge on the specific judge model. To test this, we re-run the entire FilmEval protocol with two additional multimodal models, Qwen3.5-Omni and Seed2.0-Lite, and compare their scores against those of our default Gemini~3.1~Pro judge.

\textbf{System-level rankings are model-invariant.} As shown in Figure~\ref{fig:cross_evaluator}(a), FilmWorld is ranked first by every judge, and by a consistent margin. The induced method orderings are near-identical, with a Spearman rank correlation of $\rho=1.0$ between Gemini and Qwen and $\rho=0.9$ against Seed, the sole difference being a single adjacent swap between the closely scored MovieAgent and VGoT. The conclusions drawn from FilmEval, and in particular the superiority of FilmWorld, therefore do not depend on the choice of evaluator.

\textbf{Agreement follows an objective-to-subjective gradient.} At the finer per-sample level (Figure~\ref{fig:cross_evaluator}(b)), the mean pairwise correlation across the three judges rises monotonically from Cinematic Presentation ($r=0.32$) to Film Consistency ($r=0.51$) to Novel Fidelity ($r=0.88$). This pattern mirrors the automatic-human alignment: the checklist-grounded, novel-anchored Novel Fidelity and Film Consistency metrics assess objectively verifiable properties and transfer almost perfectly across heterogeneous judges, whereas Cinematic Presentation targets aesthetic quality, which is inherently judge-dependent.

\textbf{Every evaluator aligns with human judgment.} Beyond agreeing with one another, all three judges track human perception. Figure~\ref{fig:evaluator_human} plots each evaluator's per-film overall score against the mean human rating over the 75 films. Every judge is significantly and positively correlated with human ratings (Pearson $r=0.79$, $0.75$, and $0.47$ for Gemini, Qwen, and Seed; all $p<10^{-4}$), and their Spearman coefficients are consistently high ($\rho\approx0.74$ for all three). The lower Pearson value of Seed2.0-Lite reflects a few compressed low-score outliers rather than a weaker monotonic trend, so the human-perceived quality ordering of the generated films is recovered regardless of which MLLM model instantiates FilmEval.

\section{Baseline Adaptation Details}

We detail the five baselines of the main paper. Each entry follows a uniform four-part structure: the native pipeline, our adaptation, the preserved versus modified orchestration, and the source. To isolate framework-level differences, all baselines share the same foundation stack as FilmWorld: text planning uses Gemini~3.1~Pro, image generation uses Nano~Banana~2, and video generation uses Wan~2.7. Our full reproduction code is publicly available at the repository below.

\noindent
  \raisebox{-0.2em}{\includegraphics[width=0.025\linewidth]{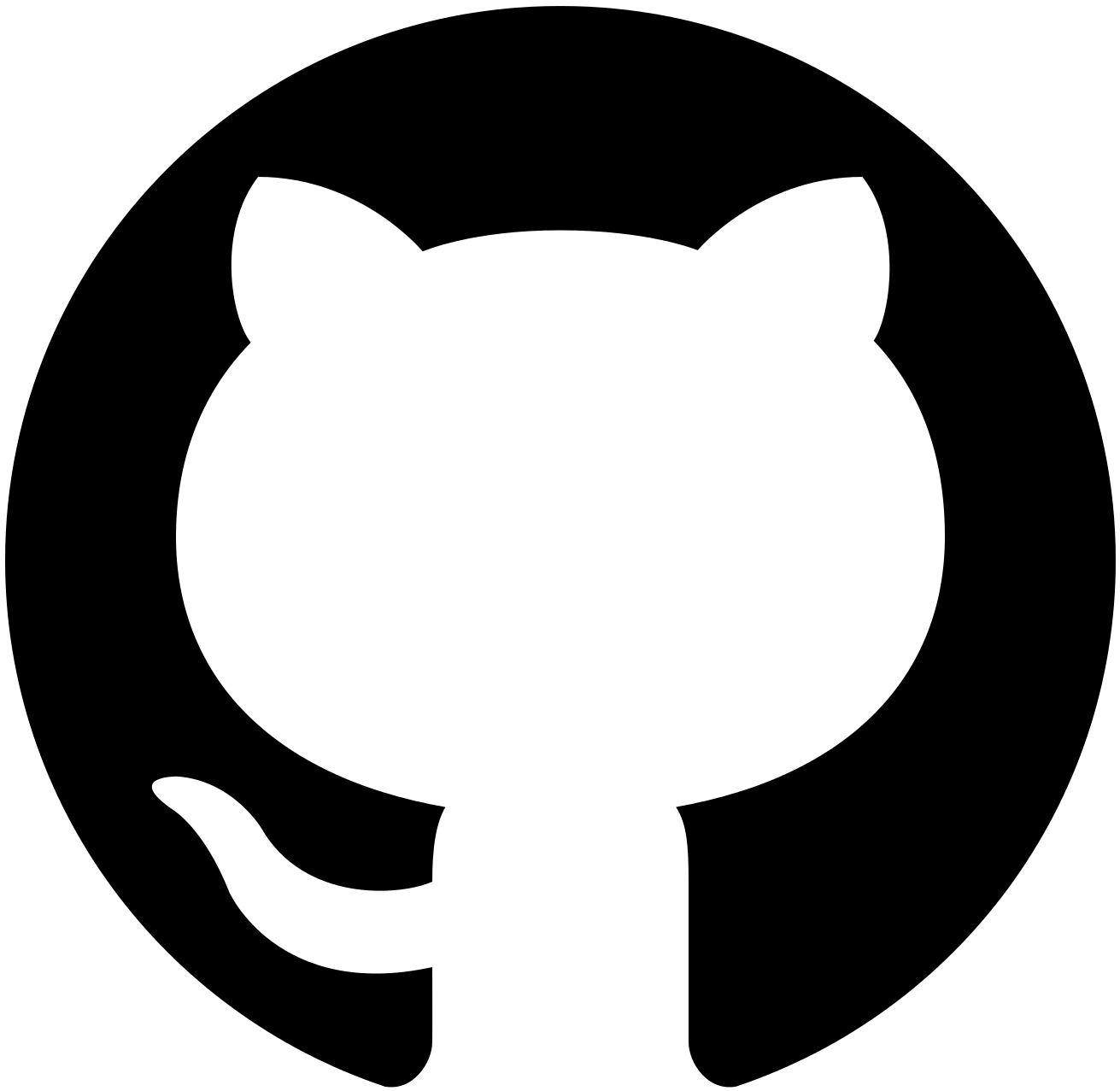}}
  \enspace\textbf{GitHub:}\enspace
  \href{https://github.com/HaoTone-monster/N2FBaseline}{
    \texttt{github.com/HaoTone-monster/N2FBaseline}
  }

\textbf{MM-StoryAgent.}
\textit{Native pipeline.} MM-StoryAgent converts a story specification into an outline and a page sequence, and then generates page-level image, speech, sound, and music assets that are composed into a video.
\textit{Adaptation.} We feed the complete novel and prompt the text model to produce a source-faithful outline and an ordered set of pages. The image model renders stylistically consistent illustrations, and MusicGen supplies the background music.
\textit{Preserved / modified orchestration.} We retain the outline-to-page hierarchy and the slideshow composition, but keep only the image and music branches; the speech and sound-effect branches are disabled.
\textit{Source.} Official public implementation: \href{https://github.com/X-PLUG/MM_StoryAgent}{\texttt{github.com/X-PLUG/MM\_StoryAgent}}.

\textbf{VGoT (VideoGen-of-Thought).}
\textit{Native pipeline.} VGoT expands a short prompt into a multi-shot storyline, generates character-stage avatars and shot keyframes, and enforces cross-shot consistency through identity conditioning and latent propagation.
\textit{Adaptation.} We convert each novel into a 25-shot storyline, generate the avatars and avatar-conditioned keyframes, and condition each shot video on its corresponding keyframe.
\textit{Preserved / modified orchestration.} We retain the dynamic storyline planning, the five-dimensional shot prompts, and the avatar assignment. Cross-shot identity is instead propagated through explicit image-space avatar references, replacing the original IP-Adapter and adjacent-latent propagation.
\textit{Source.} Official public implementation: \href{https://github.com/DuNGEOnmassster/VideoGen-of-Thought}{\texttt{github.com/DuNGEOnmassster/VideoGen-of-Thought}}.

\textbf{MovieAgent.}
\textit{Native pipeline.} MovieAgent hierarchically decomposes a screenplay into sub-scripts, scenes, and shots, and then renders character-conditioned shot images and videos for final assembly.
\textit{Adaptation.} We automatically convert the novel into a MovieScript with structured character profiles, generate canonical character portraits, and add source-faithful animation and dialogue-performance constraints.
\textit{Preserved / modified orchestration.} We retain the Screenwriter, Scene Planning, and Shot Plot Create agents. The original manual input preparation and rendering backends are replaced by our automatic novel conversion and the shared foundation stack.
\textit{Source.} Official public implementation: \href{https://github.com/showlab/MovieAgent}{\texttt{github.com/showlab/MovieAgent}}.

\textbf{ViMax.}
\textit{Native pipeline.} ViMax performs character extraction, multi-view portrait generation, storyboard design, camera-tree construction, reference selection, frame generation, and shot-video composition.
\textit{Adaptation.} We use the complete novel as the script input and remove the built-in shot-count limit. The revised prompts emphasize source faithfulness, unrestricted shot planning, stable character identities, and a consistent Ghibli style.
\textit{Preserved / modified orchestration.} We retain the camera tree, reference management, transition generation, and first/last-frame conditioning, while redirecting the text, image, and video generation to the shared backends.
\textit{Source.} Official public implementation: \href{https://github.com/HKUDS/ViMax}{\texttt{github.com/HKUDS/ViMax}}.

\textbf{VideoClaw.}
\textit{Native pipeline.} VideoClaw processes a creative concept through screenplay, character, storyboard, reference-frame, video-generation, and post-editing stages.
\textit{Adaptation.} We replace the short creative input with the complete novel and revise the prompts from short-drama expansion to faithful single-episode adaptation under consistent character and animation constraints.
\textit{Preserved / modified orchestration.} We retain the six-stage agent workflow and the project-session asset passing, while adapting the input semantics, the prompts, and the generative backends.
\textit{Source.} Official public implementation: \href{https://github.com/HITsz-TMG/AIGC-Claw}{\texttt{github.com/HITsz-TMG/AIGC-Claw}}.

\begin{figure}[htb]
\centering
\includegraphics[width=0.72\textwidth]{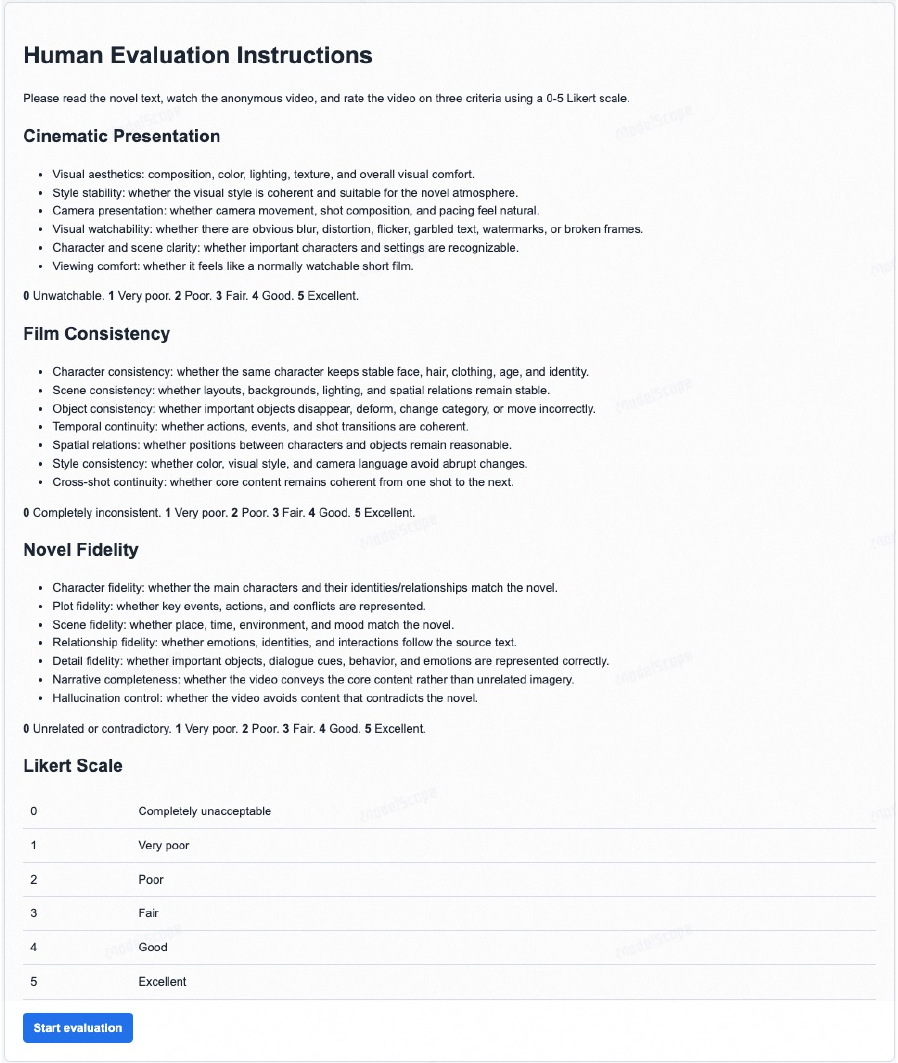}
\caption{\textbf{Human-evaluation rating rubric presented to annotators.} Each of the three dimensions is scored on a $0$--$5$ Likert scale, accompanied by a list of concrete sub-criteria and level-wise anchor descriptions.}
\label{fig:human_eval_instruction}
\end{figure}

\section{Human Evaluation Protocol}

This section documents how the human-evaluation scores reported in the main paper are collected. Human evaluation complements the automatic metrics by capturing holistic, perceptual judgments that are difficult to fully operationalize in an automated evaluator.

\subsection{Rating Rubric}
\label{app:rating_rubric}

Each annotator first reads the source novel and then watches the anonymized video before rating it on three dimensions, namely Cinematic Presentation, Film Consistency, and Novel Fidelity, each on a $0$--$5$ Likert scale ($0$ = completely unacceptable, $5$ = excellent). The three dimensions are judged independently, since strong visual quality does not imply faithful adaptation, and a plot-accurate video is not necessarily internally consistent. Cinematic Presentation targets the intrinsic perceptual quality of the video, covering visual aesthetics, style stability, camera presentation, and overall watchability. Film Consistency targets temporal stability across shots, covering the coherence of characters, scenes, objects, and their spatial relations. Novel Fidelity targets faithfulness to the source, covering character, plot, scene, and relationship restoration as well as hallucination control. The complete instructions and per-level anchors provided to annotators are shown in Figure~\ref{fig:human_eval_instruction}.

\begin{figure}[!t]
\centering
\includegraphics[width=0.8\textwidth]{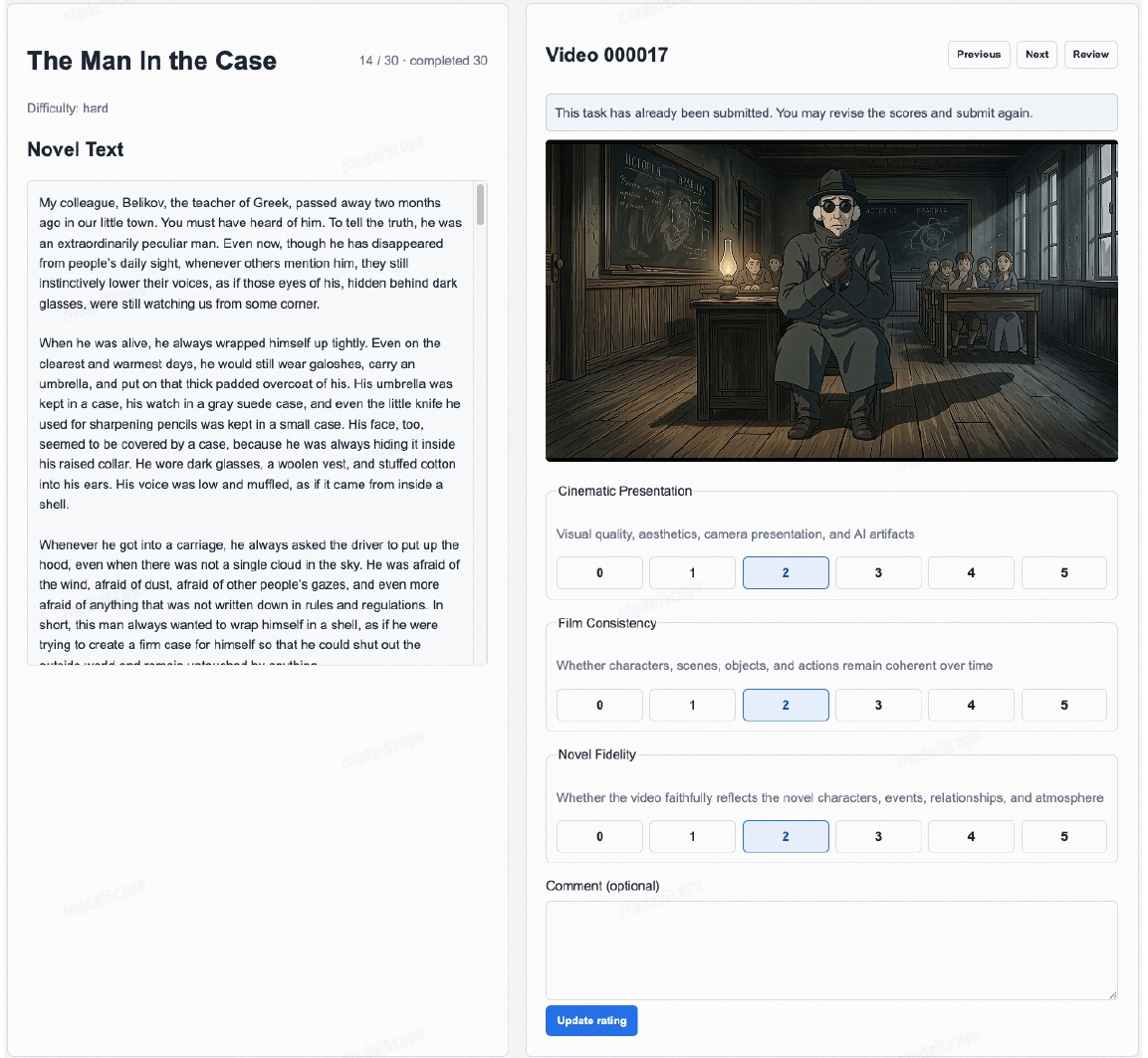}
\caption{\textbf{Web-based annotation interface for human evaluation.} The left panel shows the source novel with its title and difficulty tier; the right panel presents the anonymized video and collects independent $0$--$5$ ratings for Cinematic Presentation, Film Consistency, and Novel Fidelity, together with an optional comment.}
\label{fig:human_eval_interface}
\end{figure}

\subsection{Annotation Interface}
\label{app:annotation_interface}

To standardize the evaluation process and minimize cognitive load, we develop a dedicated web-based annotation interface, shown in Figure~\ref{fig:human_eval_interface}. The interface adopts a two-panel layout that places the source material and the video side by side, so that annotators can cross-reference the story while watching. The left panel displays the novel's title, difficulty tier, and full scrollable text, while the right panel embeds the anonymized video in a standard player with playback controls. Below the player, the three evaluation dimensions, namely Cinematic Presentation, Film Consistency, and Novel Fidelity, are each rated through a row of $0$--$5$ buttons accompanied by a one-line reminder of the criterion, followed by an optional free-text comment box. All videos are presented under anonymized identifiers (e.g., \texttt{Video 000017}) to prevent method-identity bias, and their display order is randomized across annotators. Submitted ratings can be revised and resubmitted at any time, allowing annotators to recalibrate as they gain familiarity with the pool.

\vspace{-3mm}
\section{Discussion}
\vspace{-1mm}
\subsection{Limitations and Future Work}
\label{app:limitations}

Despite its strong empirical performance, FilmWorld has several limitations that delineate its current boundary and point to directions for future work.

\textbf{Input length and hierarchical world memory.} Our experiments focus on narratives of roughly 1,000–5,000 Chinese characters or 1,000–3,000 English words. Scaling to full-length novels remains unverified and would require a hierarchical world memory that summarizes chapter-level state and reinstantiates only locally relevant entities per scene, to keep both context and computation tractable.

\textbf{Upstream world-construction errors.} FilmWorld treats the parsed symbolic trajectory as the shared source of truth for downstream agents. This design amplifies consistency when the trajectory is correct, but the same mechanism can propagate errors from entity resolution, implicit-event recovery, temporal ordering, or state assignment, yielding films that are internally consistent yet subtly unfaithful to the source. Uncertainty-aware state construction and provenance links from each state assignment back to the text are promising safeguards.

\textbf{Discrete world-state representation.} Encoding entities through discrete attributes such as age stage, costume, season, and weather is what makes consistency explicitly enforceable, but it also quantizes properties that are inherently continuous, such as gradual aging, evolving lighting, and subtle emotional shifts. Hybrid representations that combine discrete events with continuous attributes and states are a natural next step.

\textbf{Absence of an explicit 3D spatial prior.} Spatial relations in FilmWorld are anchored only at the textual and visual-reference level, without an underlying geometric scene, which can leave entities spatially unanchored in visually complex layouts. Introducing a coarse 3D scene proxy would strengthen cross-shot spatial coherence and camera-consistent rendering.

\textbf{Cinematic artistry beyond narrative fidelity.} FilmWorld privileges faithful restoration of narrative content, yet cinematic adaptation also relies on omission, pacing, montage, symbolic camera language, and thematic reinterpretation. Rendering prose event by event can therefore penalize legitimate directorial re-creation, and our automatic metrics, validated only on 15 moderate-length narratives across two languages under a single visual style, do not assess aesthetic qualities such as tone, mood, or auteur voice. Broader benchmarking and explicit modeling of directorial intent are left to future work.

\textbf{Rendering fidelity bounded by foundation models.} As analyzed in Appendix~\ref{sec:b3}, FilmWorld enforces consistency at the symbolic level, but pixel-level fidelity remains bounded by the underlying image and video generators. Since our framework is modular with respect to these backbones, residual rendering artifacts are expected to diminish as foundation models advance.

\vspace{-1mm}
\subsection{Misuse Risk and Responsible Release}
\label{app:misuse}

As a novel-to-film system, FilmWorld inherits the dual-use risks common to generative media. We outline the principal concerns below, together with the measures we take to mitigate them.

\textbf{Intellectual property.} A film generated from a novel is a derivative work, and applying FilmWorld to copyrighted texts without authorization may raise intellectual-property concerns. FilmEval is therefore built exclusively from original, public-domain, or substantially rewritten sources, and users deploying the system elsewhere are responsible for securing the rights to their inputs.

\textbf{Synthetic media misuse.} Like any video generation system, FilmWorld could in principle be misused to fabricate misleading footage. Two properties of our current setting partially mitigate this risk: outputs are rendered in an overtly stylized, non-photorealistic aesthetic, and character identities are drawn from fictional entities rather than real people. These are mitigations rather than guarantees, and continued caution will be warranted as foundation generators become more photorealistic.

\textbf{Responsible release.} We release our resources for reproducibility and downstream research. The release is intended for research use, is distributed under a license that prohibits harmful applications such as disinformation and non-consensual depiction of real persons, and encourages clear disclosure that outputs are AI-generated. FilmWorld serves as a tool to assist human creators rather than to replace editorial judgment.

\end{document}